\documentclass[manuscript,screen]{acmart}

\usepackage{pdflscape}
\usepackage[absolute]{textpos}
\setcopyright{acmcopyright}
\copyrightyear{2025}
\acmYear{2025}
\acmDOI{10.1145/3777455}

\begin{document}

\begin{textblock*}{15cm}(2.5cm, 1.0cm)
"This paper is an early version of work that has been published in an enhanced/improved version in the Transactions on Human-Robot Interaction (under a slightly changed title). We keep this version for consistency, but please rather read/use/cite the final Paper named \textit{Design Implications for Robots that Facilitate Groups–A Scoping Review on Improving Group Interactions through Directed Robot Action}. The definitive Version can be found at http://dx.doi.org/10.1145/3777455."
\end{textblock*}

\title[Social Mediation through Robots]{Social Mediation through Robots - A Scoping Review on Improving Group Interactions through Directed Robot Action using an Extended Group Process Model}

\author{Thomas H. Weisswange}
\authornote{Both authors contributed equally to this research.}
\email{thomas.weisswange@honda-ri.de}
\orcid{0000-0003-2119-6965}
\affiliation{
  \institution{Honda Research Institute Europe GmbH}
  \streetaddress{Carl-Legien-Str. 30}
  \city{Offenbach}
  \country{Germany}
  \postcode{63073}
}

\author{Hifza Javed}
\authornotemark[1]
\email{hifza_javed@honda-ri.com}
\orcid{0000-0002-5414-6318}
\affiliation{
  \institution{Honda Research Institute USA, Inc.}
  \streetaddress{70 Rio Robles}
  \city{San Jose}
  \state{California}
  \country{USA}
  \postcode{95134}
}

\author{Manuel Dietrich}
\email{manuel.dietrich@honda-ri.de}
\orcid{0000-0001-6819-8656}
\affiliation{
  \institution{Honda Research Institute Europe GmbH}
  \streetaddress{Carl-Legien-Str. 30}
  \city{Offenbach}
  \country{Germany}
  \postcode{63073}
}

\author{Malte F. Jung}
\affiliation{
  \institution{Cornell University}
  \streetaddress{343 Campus Road}
  \city{Ithaca}
  \state{NY}
  \country{USA}
}
\orcid{0000-0001-9359-7122}
\email{mjung@cornell.edu}

\author{Nawid Jamali}
\affiliation{
  \institution{Honda Research Institute USA, Inc.}
  \streetaddress{70 Rio Robles}
  \city{San Jose}
  \state{California}
  \country{USA}
  \postcode{95134}
}
\orcid{0000-0003-0660-000X}
\email{njamali@honda-ri.com}

\renewcommand{\shortauthors}{Weisswange and Javed, et al.}

\begin{abstract}
Group processes refer to the dynamics that occur within a group and are critical for understanding how groups function. With robots being increasingly placed within small groups, improving these processes has emerged as an important application of social robotics. \textit{Social Mediation Robots} elicit behavioral change within groups by deliberately influencing the processes of groups. While research in this field has demonstrated that robots can effectively affect interpersonal dynamics, there is a notable gap in integrating these insights to develop coherent understanding and theory.
We present a scoping review of literature targeting changes in social interactions between multiple humans through intentional action from robotic agents. To guide our review, we adapt the classical Input-Process-Output (I-P-O) models that we call "Mediation I-P-O model". 
We evaluated 1633 publications, which yielded 89 distinct social mediation concepts. We construct 11 mediation approaches robots can use to shape processes in small groups and teams. This work strives to produce generalizable insights and evaluate the extent to which the potential of social mediation through robots has been realized thus far. We hope that the proposed framework encourages a holistic approach to the study of social mediation and provides a foundation to standardize future reporting in the domain.
\end{abstract}


\received{September 2024}

\maketitle

\section{Introduction}
Group processes refer to the various dynamics and interactions that occur within a group setting, influencing the behaviors, attitudes, and performances of its members \cite{Zander1979TheProcesses, Forsyth2018GroupDynamics}. 
These processes encompass a wide range of phenomena, including communication patterns, decision-making mechanisms, conflict resolution strategies, leadership roles, group cohesion, and the establishment of norms and roles within the group \cite{Brown2019GroupGroups}. 
Understanding group processes is critical for analyzing how groups function, how they achieve their goals, and how individual members are affected by their group membership.

As robots are increasingly placed within small groups and teams, more and more research has focused on building understanding of how robots shape the processes and dynamics of small groups and teams \cite{Sebo2020}.
This work encompasses studies across the lab (e.g., \cite{Erel2021, Fraune2020OurTeams, Tennent2019}) and field (e.g., \cite{Cheatle2019SensingRoom, Kanda2007AInteraction, Sabanovic2013PARODementia} and initial efforts to develop theory \cite{Abrams2020, Chen2023RobotsConnection}). More recently, a line of research has been developing that seeks to understand how robots can be used deliberately to shape the processes of small groups and act as moderators and mediators (e.g.,\cite{Jung2015UsingViolations, Shen2018}), social catalysts (e.g., \cite{Joshi2017ACatalysts, Chen2023RobotsConnection}), or socially assistive robots for groups (e.g., \cite{Short2017}). For recent reviews on a robot's influence on interactions and interpersonal dynamics in groups, see ~\cite{Chita-Tegmark2020ElderlyMediationReview, Dahiya2023, Gillet2024Interaction-ShapingAgents, Sebo2020, Schneiders2022Review}.

While this work has demonstrated that robots can effectively influence social and interpersonal dynamics in small groups, there is a notable gap in integrating these insights to develop coherent understanding and theory about deliberate robotic influence on small group processes. This lack of comprehensive understanding leaves critical questions unanswered:
What specific roles can robots play in shaping group processes? 
What are the specific affordances offered by robots that enable them to shape group processes? 
How do specific design characteristics and intervention strategies impact a robot’s effectiveness in shaping group processes? 
Addressing these questions is essential for advancing our understanding about the design of robots as mediators of small group processes.

The goal of this paper is to provide answers to these questions through an integrative review of the literature on HRI research that targets changes in social interactions between multiple humans through directed and intentional action from an embodied robotic agent. We adopt the term \textit{Social Mediation Robots} to describe robots that elicit behavioral change within groups by deliberately influencing the processes of small groups and teams. To guide our review, we use an extended version of classical Input-Process-Output (I-P-O) models ~\cite{Forsyth2018GroupDynamics, Ilgen2005TeamsModels, Marks2001AProcesses} that we call "Mediation I-P-O model". The IPO model was originally proposed by Hackman and Morris ~\cite{Hackmann1975} as a framework to guide research on work groups and teams. We extend this model by distinguishing mediation processes from group processes and by highlighting specific "mediation factors" that shape those mediation processes. Our review identifies 11 mediation concept categories for which we have found evidence that robotic mediation is possible and highlights specific mediation approaches robots can use to shape processes in small groups and teams. We also identify unique affordances offered by robots that make them particularly suitable for social mediation.

\section{Background}
We aim to embed our overview into existing research on groups from social, psychological and organizational sciences. In this section, we highlight relevant taxonomies and models and identify trends and open research areas in these domains.

\subsection{Group Processes}		
We define a group using Forsyth's definition \cite{Forsyth2018GroupDynamics} as “two or more individuals who are connected to one another by social relationships” (pp 2-3). Groups can be distinguished into different types based on their perceived entitativity. Lickel et al. \cite{Lickel2000} distinguished between \textit{intimacy groups} such as friends or families,  \textit{task groups} such as surgical teams or a student campus committee, \textit{social categories} such as women or children, and \textit{loose associations} such as people in line at a bank.

In social psychology, the term group process is used as an umbrella term that refers to a broad range of group related phenomena that influence the behavior, attitudes, and performance of its members \cite{Zander1979TheProcesses, Forsyth2018GroupDynamics}. This often includes group cognition, social influence, cohesiveness, social categorization and identity, group structure, conflict, and leadership\cite{Hogg2002BlackwellProcesses, Zander1979TheProcesses}. Research in management and organizational behavior with its focus on work groups and teams often defines group processes more narrowly as ''interactions such as communication and conflict that occur among group
members and external others'' \cite{Cohen1997WhatSuite}, or more broadly as ''patterned relations among team members'' \cite{McGrath1984Groups:Performance}. Marks, Mathieu and Zaccaro famously distinguished group processes from emergent states as ''members' interdependent acts that convert inputs to outcomes through cognitive, verbal, and behavioral activities directed toward organizing taskwork to achieve collective goals'' \cite{Marks2001AProcesses}. 

A highly influential model to guide research into group processes has been the Input-Process-Outcome (I-P-O) model \cite{McGrath1984Groups:Performance}. The model views group processes "as mediating
mechanisms linking such variables as member, team, and organizational characteristics with
such criteria as performance quality and quantity, as well as members' reactions'' \cite{Marks2001AProcesses}. The model has later been extended to include not only task outcomes, but also individual and group factors~\cite{Hackman2002LeadingPerformances}. While the I-P-O model has mostly focused on work groups and teams and has been critiqued for its rather static treatment of groups \cite{Ilgen2005TeamsModels}, it still provides a useful framework to guide our review of a robot's influence on group processes.

\subsection{Mediation}
Mediation is most commonly associated with conflict resolution through the help of an independent party that is responsible for creating and repairing social bonds by organizing exchanges between the involved parties \cite{Jokinen2012SocialExclusion}. According to Fischer et al.~\cite{Fisher1980GettingIn}, the role of a mediator is to enable effective negotiation by managing and containing “negative” emotions such as anger, hostility, envy, and distrust. In practice, mediation is applied across a variety of settings. Social mediation in workplace environments usually involves resolving disputes or facilitating meetings to achieve team goals~\cite{Viller1991}. In multicultural learning environments, mediation is also applied to cater to the social, cultural, and linguistic heterogeneity of the learners by providing diverse and more inclusive teaching practices and tools \cite{DeAbreu2005TheIntroduction, Cesar2005TheSchool}. While team coaching typically entails improving team performance through task-focused interventions\cite{Hackman2005ACoaching}, feedback from practitioners and experimental studies often includes improvements in interpersonal dynamics~\cite{CarrPeters2012, Hastings2019TeamCoaching}. Other applications of mediation include couples or family counseling~\cite{Shadish2003MaritalTherapy} or moderated support groups~\cite{Jacob21998Counseling}.

Notably, process models for group mediation have not yet been established~\cite{CarrPeters2012, Rico2016TeamEffectiveness}. Kressel \cite{Kressel2006MediationRevisited} defined three types of interventions: reﬂexive (establishing the groundwork), contextual (producing a climate conducive to constructive dialogue and problem solving), and substantive interventions (dealing directly with the issues in dispute). However, most mediation theories remain mainly descriptive in nature rather than predictive and are missing the general framework that can be empirically tested \cite{Zariski2010}. For example, Moore \cite{Moore2014Mediation} defined 12 stages of the negotiation and conflict mediation process but did not explicitly relate them to the underlying group dynamics. However, some research is available on the types of interventions that can improve group processes and outcomes in given settings~\cite{Bostrom1993GroupSystems}. It was shown that process facilitation can have a positive impact on meeting processes, and meeting processes have a strong positive impact on satisfaction \cite{Miranda1999MeetingInterventions}. Additionally, an active intervention style during a community mediation led to higher participant satisfaction, even in cases where the settlement was not in a person’s favor \cite{Alberts2005DisputantProgram}.

We are interested in evaluating the potential of social robots to act as mediators, as they might allow a more ubiquitous use to also benefit groups that do not have a human mediator available or to act in spontaneously occurring group situations.

\subsection{Social Robotics}

In contrast to other types of robots, social robots are not meant to serve as tools of human purpose but rather as mediating artefacts between humans or humans and their environments \cite{Turkle2017AloneOther}. 
They have been shown to be effective in several human-robot interaction applications. They serve as companions in older adult care settings, particularly for patients of dementia or Alzheimer's disease \cite{Moyle2013SocialEnvironment, Cifuentes2020SocialCare, Anghel2020SmartServices}. In addition to functional roles such as medication reminders and physical assistance \cite{Huang2021AttitudesLiving, ukasik2018CouldUsers} companion robots serve in social roles that require interpersonal dialog, establishing social bonds, and providing entertainment with activities such as board games, singing, dancing, and storytelling \cite{Gasteiger2022OlderStudy, Chiu2021NeedsStudy}. They are also shown to be effective in autism interventions for children, facilitating sensory experiences \cite{Javed2019ADisorder,Cabibihan2018SocialTest}, development of social skills \cite{Javed2022PromotingCare, Chung2021Robot-MediatedStudy, Rakhymbayeva2021ATherapy}, and understanding of emotional expressions \cite{Pinto-Bernal2023DoASD, Xiao2020DeepDisorders}. Social robots have also shown promise as tutors in educational settings, helping students achieve their learning goals effectively \cite{Donnermann2020IntegratingStudy, Donnermann2022SocialEducation, Kanero2022AreLearning, Shi2022TowardAutism, Nasir2022WhatApplications}. In addition, they have been used in mental health applications \cite{Li2023TellHealth}, specifically offering cognitive interventions for social anxiety \cite{Rasouli2022PotentialAnxiety, Rossi2020EmotionalVaccination} and depression \cite{Lee2017StepsDepression}. However, these applications typically involve one-on-one interactions between a human user and a robot, where influence on social interactions with other humans may be a byproduct of the human-robot interaction but not the primary target. 

Embodied social robots come in various forms: bio-inspired designs imitate humans or animals, artifact-shaped ones resemble everyday objects like lamps or cars, and functional robots have mechanical appearances \cite{Baraka2020AnRobots}. These designs offer distinct affordances for socio-emotional interactions. For instance, a furry animal robot can be petted to alleviate social isolation in older adults \cite{Moyle2013SocialEnvironment}, while a bio-inspired robot with a screen and multi-modal expressivity can enhance children's engagement in play activities \cite{Kim2021YoungRobot}.

Notably, robot-assisted mediation of human group interactions remains relatively under-explored. 
Key challenges in this domain include perception of complex interpersonal dynamics in group interactions \cite{Javed2023GroupMediation} and designing contextually-relevant and effective robot behaviors to influence these dynamics \cite{Pham2024EmbodiedConsensus-Building, Rifinski2021Human-human-robotInteraction}.
This research may benefit from a stronger foundation in the principles from social psychology and group research, for example, by utilizing a standardized framework to identify input factors, group processes, and outcomes relevant to group mediation applications. The Mediation I-P-O model presented in this work offers such a framework, aimed at enhancing group outcomes beyond task-related and especially social ones, by leveraging robots to mediate the underlying social dynamics in a group.

\subsection{Mediation for Group Interactions: Perspectives from Current Frameworks}
One objective of a scoping review is to identify key approaches, theories, and gaps in a research theme~\cite{Munn2022WhatSynthesis}. To frame these discussions, we utilize established models from group research to structure our review. Firstly, we examine a group process model and its application beyond the traditional restriction to teams. Subsequently, we describe an extension to this framework that builds upon previous models to integrate a focus on mediation-related aspects. To broaden its relevance beyond merely task-oriented interactions, we highlight the social factors that influence group dynamics, integrating these into the model to better understand the social phenomena that occur during group interactions. Finally, we also highlight the key mediation-related factors in the extended model that are important for examining the function of Social Mediation Robots in this review.

\subsubsection{Group Process Model to examine group interactions}
\begin{figure}
     \centering
     \includegraphics{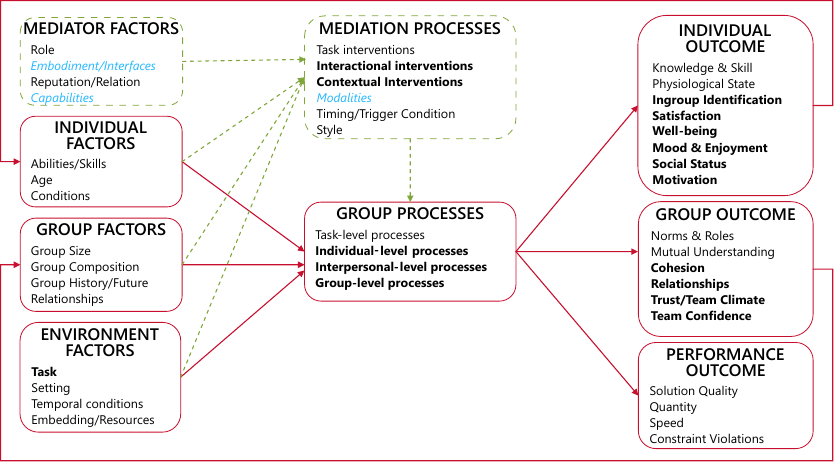}
     \caption{Mediation I-P-O model. Red: Original model as proposed in \cite{Forsyth2018GroupDynamics} with feedback of outcomes as next inputs as discussed in \cite{Ilgen2005TeamsModels, Marks2001AProcesses}. Green: Extension to cover group mediation. Specific aspects and process types that are specifically relevant for social group mediation are highlighted in bold. Mediation aspects of high relevance if the mediator is a robot are shown in blue.}
     \Description{Figure 1. The figure depicts the structure of the mediation I-P-O model. On the left side are the different input factors - individual, group, and environment input factors from the original model, and additionally a box for mediator factors. The former are connected to the center box containing group processes, while the latter leads to another new box containing mediator factors. Mediator factors are influencing group processes. The group processes box is connected to three different types of outcomes - individual, group, and performance outcomes. From individual and group outcomes there is another connection, feeding back into the respective input factors to influence the next iteration of a group activity. All boxes contain more details about the respective aspects they cover, which are described further in the text.}
     \label{fig:RoboIPO}
\end{figure}

This section provides an overview of the classical I-P-O framework~\cite{Forsyth2018GroupDynamics, Hackmann1975} (shown as boxes and arrows in red in Figure \ref{fig:RoboIPO}) and includes a commentary to motivate the need for an extended model. 
The structure includes the initial conditions before any interaction begins (inputs), the interactions and activities that occur among group members (processes), and the outcomes of the group's activities (outputs).
We are aware that such models have been critizised for not capturing the complexity of causal interactions through a group's life \cite{Ilgen2005TeamsModels, Marks2001AProcesses}, but we believe it is still useful to structure the analysis of instances of group activities as they are present in work relevant for this review.
Although we agree with follow-up proposals that the framework should also include a feedback of at least some of the outputs into inputs of a next step in a groups lifetime~\cite{Ilgen2005TeamsModels, Marks2001AProcesses}, we refrain from using the adapted terminology of "I-M-O-I" (input-mediator-output-input) to emphasize group processes as being dynamic \footnote{and to prevent confusion between their statistical mediator understanding and our understanding as agent role}. 

\textbf{Input factors} can be grouped into individual, group, and environment factors. Examples of \textit{individual factors} include age, skills and knowledge, or physical and health conditions of each group member.
\textit{Group factors} include the number and composition of its members, the social and professional relationships and existing cohesion between them, and the reason and timeline for the group's existence. 
Finally, examples of \textit{environment factors} include the setting in which the interaction takes place, available resources and tools, or other implications from the general within which the interaction is embedded, such as time pressure or general impact.
A very important environment factor is the task, that is, the explicit assignment or target of the group interaction.
There have been several proposals to structure task categories for groups (e.g., \cite{Davis1976TheInteraction, McGrath1984Groups:Performance, Wildman2011TaskAttributes}). 
Although these typically originate from performance outcomes, there may also be task categories that revolve around individual and group outcomes \cite{McGrath1991TIP}. 
For our analysis, we will adopt the categories proposed by McGrath \cite{McGrath1984Groups:Performance} 
that describe tasks based on performance processes (\textit{Generate, Choose, Negotiate, and Execute}), 
the required type of interaction between the members (\textit{Cooperation, Conflict})
, and if the task requires behavioral or conceptual actions.
Additional information about task types can be found in appendix \ref{Appendix:TaskList}.

\textbf{Group processes} refer to the interactions, mechanisms, and dynamics that occur within a group and describe the variables that causally influence a mapping from the inputs to the outputs~\cite{Marks2001AProcesses}. In our understanding, group processes describe what is happening during group actions, often expressed through the dynamics of events. 
Based on this understanding, we view factors such as ‘cohesion’, which can transfer across group activities, as target outcomes instead of emergent states unlike in previous proposals \cite{Marks2001AProcesses}. 
Although many models (e.g. \cite{Aldag1993BeyondProcesses, Marks2001AProcesses} characterize group processes with a strong focus on working teams and their tasks, Forsyth emphasizes social factors, with process categories comprising formative, influence, performance, conflict, and contextual processes~\cite{Forsyth2018GroupDynamics}. 
We consider the following four types of group processes in this work.
\textit{Task-level processes} encompass both transition and action processes, including task assignment and collaborative task execution processes \cite{Marks2001AProcesses}. 
\textit{Individual-level processes} describe processes within each group member, including changes in engagement, motivation, or affect in response to other processes.
\textit{Interpersonal-level processes} contain social interactional dynamics between two or more group members.
Finally, \textit{group-level processes} encompass the dynamics of the group as a whole, including membership changes, the group’s progression through developmental stages, and role distribution among members.
In addition, this includes group-specific psychological phenomena such as evaluation apprehension, social loafing, groupthink (e.g. \cite{Maier2020IDNMediation}), or group flow \cite{Pels2018GroupFindings}. 
In this paper, we focus on the interpersonal and group-level processes for social mediation applications, although others may be affected as well.
It is important to note that a number of studies utilize measurable constructs in place of group processes, such as participation balance to estimate group performance~\cite{Woolley2010EvidenceGroups} and interaction counts to estimate cooperation~\cite{Wollstadt2023QuantifyingTheory}. 
As these do not describe an ongoing dynamics, but are also only correlates of final group outcomes, we refer to these as \textit{proxy factors}.

Finally, \textbf{Outcome factors} include potential targets that can be used to describe and quantify states resulting from a group activity. 
They are typically split into three components related to performance, individual and group factors \cite{Forsyth2018GroupDynamics, McGrath1991TIP}. 
Current literature focuses mainly on \textit{performance outcomes}, which describe if and how well a task has been solved with respect to quality, quantity, and speed of solutions, and potential constraint violations.
\textit{Individual outcomes} measure changes in group members’ states over time, such as gains in skills or knowledge. 
Finally, \textit{group outcomes} describe states of the group as a whole, such as established norms and roles, shared and distributed knowledge like transactive memory\cite{Wegner1987TransactiveMind}, and shared mental models~\cite{Klimoski2016}.
A detailed list of relevant outcome types can be found in appendix \ref{Appendix:OutcomeList}.
In addition, both group and individual outputs may influence the input factors of future group activities, forming the feedback loops shown in Figure \ref{fig:RoboIPO}. This is supported by prior work~\cite{Ilgen2005TeamsModels, Marks2001AProcesses} that proposes the inclusion of a feedback loop to indicate that the effects of group interactions can inform and reshape initial conditions.

\subsubsection{Extending Group Process Model for mediation}
Since the I-P-O model has not yet been applied to mediation scenarios, we propose an extended model called the Mediation I-P-O. The green boxes in Figure \ref{fig:RoboIPO} represent our extensions, highlighting that a mediator changes the outcomes by influencing group processes. Firstly, we added \textbf{Mediator factors}, which represent an additional input category that describes factors related to the mediator. These include the role and position of the mediating agent, that is, whether they are an external party with a designated mediation role or an internal member. Additionally, these include the embodiment of the mediating agent and the interface it offers for group members to interact with it. Other important mediator factors include the relation of the mediator to other group members and their capabilities and skills.

Secondly, we added \textbf{Mediation processes}, which encompass all aspects of group-directed behavior performed by the mediator. These include both verbal and non-verbal actions and are strategically timed based on the existing group dynamics. Mediators typically engage in contextual interventions to set up a conductive group environment and reflexive interventions to better understand the task and group context~\cite{Kressel2006MediationRevisited}. Our model primarily considers interventions occurring during group activities, thus excluding reflexive interventions. We also establish contextual interventions to include actions directed at improving the context \emph{during} a group activity. Other mediation behaviors that directly target group processes include task interventions and interactional interventions~\cite{Dickson1996FacilitatingEnvironment}. Task interventions focus on optimizing how the group tackles tasks, for example, by structuring activities, assigning roles or by providing additional information. Mediators may also offer feedback or intermediate assessments during this process. Interactional interventions involve managing the dynamics between two or more group members, such as encouraging equal participation or facilitating information exchange. The mediator will also have to decide when to execute an intervention, for example evaluate how much each group member contributed and eventually encourage those with low activity.
Some of these interventions can also be performed without explicit awareness of the group or without directly addressing an ongoing process ('Mediation Style').
For comprehensive details on types of mediator interventions, please refer to the appendix \ref{Appendix:InterventionList}.

While mediation processes do not directly determine outcomes, they aim to support the group's goals. 
Specific measures, such as the perceived neutrality \cite{Szejda2019NeutralityPerspectives} or fairness \cite{Cropanzano2001MoralJustice} of the mediator by the group members, offer insights into the mediator's influence. These subjective evaluations, although usually made upon the completion of the group process, are only correlated with the outcome factors and cannot be considered outcomes themselves. In line with our previous choice of terminology, we refer to these as \textit{mediation proxy factors}.

\subsubsection{Emphasizing the social factors}
As mentioned, a significant portion of group research has concentrated on task-focused processes and outcomes. However, we argue that the I-P-O framework inherently does not impose such restrictions. The effectiveness of the framework is contingent upon the chosen factors and the evaluation of their interactions. For this reason, we will highlight and discuss some factors that we consider particularly relevant when researching socio-affective group interactions (see parameters highlighted in bold in Fig. \ref{fig:RoboIPO}). Among the input factors, the nature of the task carries considerable importance. However, so far, the study of task types has primarily been explored within a business context. It must be noted that although we refer to McGrath’s task taxonomy \cite{McGrath1984Groups:Performance}, this does not encompass tasks that groups engage in outside work environments. Comprehensive discussions on such tasks are scarce as these have solely social objectives. Since tasks remain the main driving force behind a group's interaction and directly influence the desired outcome, we suggest expanding the taxonomy to include six new tasks, along with the prospect of an absence of a task. The new task categories encompass ‘Leisure’ activities, such as conversations or free play, aimed at enhancing affective states through social interaction, and ‘Information Exchange’ tasks where groups seek to understand each other’s personal experiences and histories. ‘Networking’ tasks involve groups coming together to improve relationships and cohesion through interaction and shared experiences. ‘Learning/Teaching’ and ‘Care’ tasks focus on groups that work on developing each other’s cognitive or physical abilities or improving and overcoming cognitive or physical challenges. Lastly, ‘Co-existing’ describes a collection of individual tasks and goals which that necessitate interaction through the communal use of space or resources. In such scenarios, the social objective is to minimize negative impact on others and not violate social norms \cite{Rother2023Coexistence}. A comprehensive exposition of both the new and existing task types is available in appendix \ref{Appendix:TaskList}.

When considering outcomes, we would like to highlight some socio-affective factors that have been discussed previously, at times even within the classical model. 
Among individual outcomes, affective states are important, which individuals develop in response to the activities within the group. These may include satisfaction \cite{Hecht1978TowardSatisfaction, Keyton1991EvaluatingVariable, Wanous1972MeasurementSatisfaction}, well-being \cite{Cooke2016Wellbeing}, or mood states \cite{Desmet2016Mood}.
Another example includes the perception towards the group (in-group identification \cite{Abrams2020, Leach2008Group-LevelIdentification}).
A frequently employed construct, albeit inconsistently, is cohesion, which describes the strength of member unity within a group \cite{Carron1985TheQuestionnaire, Dion2000GroupConstruct, Salas2015Cohesion}.
Related factors include group trust, belief in the dependability of group members and a member’s willingness to be vulnerable to group actions \cite{Mayer1995Trust}, and psychological safety, the perception of the consequences of taking interpersonal risks \cite{Edmondson1999PsychologicalTeams}.
Interpersonal relationships also emerge as an outcome of group interactions, enriching future group dynamics.

Among mediation processes, we consider interventions that either directly address social interactions within a group activity or take a more indirect approach to support the group to achieve socio-affective outcomes.
Examples of direct interventions may include facilitating conflict resolution or encouraging positive interaction norms.
Examples of social outcome-focused interventions may include the strengthening of a group's identity or the relationships between individual members, as well as taking action to include outsiders in the group or equalize participation opportunities.
Additional details regarding the types of interventions can be found in appendix \ref{Appendix:InterventionList}.

\subsubsection{Key factors for mediation through robots}
In contrast to the framework used by Sebo et al.~\cite{Sebo2020} and You et al.~\cite{You2017TeamingTeamwork}, our framework asserts that all mediator factors influence the group processes exclusively through mediation processes. We highlight some key mediation-related input factors and processes that apply specifically to mediation robots, as highlighted in blue in Fig. \ref{fig:RoboIPO}. Firstly, robot embodiment is a critical input factor since it has been shown to influence user expectations and acceptance (see e.g. \cite{Mori2012TheValley, Shibata2001MentalChildren}). It can also impact the roles a robot is able to perform--for instance, a small form factor may be better suited for peripheral roles \cite{Hoffman2015}. Additionally, embodiment is also related to the interfaces available to a user while interacting with the robot and dictates the robot's capabilities in performing mediation. Finally, within the mediation processes, the modalities used by the robot to interact with its users, such as speech, gestures, gaze focus, etc., are crucial in enabling robot actions that are interpretable, meaningful, and impactful, and allow for unique affordances that may help navigate limitations compared to human mediators.

\section{Review Methods}
Our methods follow the PRISMA extension for scoping reviews \cite{Tricco2018PRISMAExplanation}.
The review aims to examine the extent, variety, and nature of the work on the topic of social mediation with robotic technologies and identify gaps in the literature to aid the planning and commissioning of future research.
We considered scientific publications published until September 2023 that are available online and are written in English. We excluded books and theses, but included conference proceedings, extended abstracts and preprints (e.g. Arxiv.org) and journal papers.
An exhaustive search was conducted using the ACM Digital Library as well as Google Scholar, which allows access to cross-disciplinary sources and pre-print archives \cite{Halevi2017SuitabilityLiterature}.
We defined 4 inclusion criteria to identify applications and concepts relevant to our definition of social group mediation through robots. All considered literature must include:
\begin{enumerate}
    \item A scenario with a group of humans that interact with each other, where a group is comprised of 2 or more individuals,
    \item An embodied technical entity that can react to humans (including robot designs and simulated robots),
    \item An entity whose behavior is designed specifically for influencing a group or interpersonal interactions (as opposed to interacting multiple humans on an individual level), and
    \item A robot with a distinct role within the interaction rather than acting as an interactant in an equal capacity as the humans.
\end{enumerate}
We target to discuss any work that discusses new \emph{concepts} for mediating groups with robots. To gather insights about roles and framings and potential applications, we mainly want to look at how the authors intend the robot to work, rather than what they demonstrated in an experiment. 
We therefore also included design and technical demonstration work without experimental studies, as well as Wizard-of-Oz studies if they meet our criteria. Given their potential to identify out-of-the-box ideas, without the constraints of implementation, these studies can be crucial for identifying gaps and shaping future research directions.

To enable an intuitive filtering of relevant literature, we use the following exclusion criteria.
We do not include work where the robot addresses only a single human without including a relation to a concrete group. Examples are robots that provide training for social skills to an individual which is later affecting general social interactions (e.g. \cite{Kim2013SocialAutism, Wainer2010Kaspar}), or robots interacting in a way with humans that biases their future social behavior \cite{Erel2021ExcludedOstracism, Kothgassner2017Real-lifeAgents}. 
We also do not include work were the pure presence of a device impacts a group (e.g. \cite{Carros2020Elderly, Joshi2019RobotsSettings, Kidd2006AElderly, Riether2012SocialRobots, Tan2018InducingMechanisms}), including robots used as a therapy tool \cite{Shibata2011_RoboTherapy}). 
Work from human-robot teaming, which focuses on task collaboration within a mixed group of humans and robots \cite{Carlson2015Team-buildingRobots, ONeill2022HumanAutonomyLiterature, You2017TeamingTeamwork} will also not be included, unless the robot additionally acts on the social interactions between the team members. 
This also involves cases where robots play games with one or multiple humans \cite{Correia2023RobotGames} or provides task assistance to a group of humans \cite{Tanneberg2024ToInteractions}, and the research field of social navigation or evacuation, which is considering the interaction between robot and crowd behavior \cite{Mavrogiannis2023CoreSurvey, Jiang2020PedestrianInteraction, Zhou2019GuidedChallenges}.
We also want to exclude work in which a robot is designed for a 1-to-1 interaction with a human but is robust to perform this also when multiple humans approach (e.g. \cite{Gehle2014SignalingRobot., Kanda2009AnMall, Tasaki2004DynamicDistance}). 
Similarly, we do not include approaches where a robot performs behaviors in front of a group, as for exampling in teaching or training applications (e.g. \cite{Belpaeme2018SocialReview, Cabibihan2013WhyAutism, Rosanda2019RobotTeacher, Sabanovic2013PARODementia, Shiomi2015CanClassrooms, Woo2021TeacherRobots}), if it is not addressing interpersonal processes.
Meeting support tools \cite{Bergstrom2007Clock, Kim2008MeetingFeedback, Waibel2003SMaRT} that do not feature an explicit embodiment are also not considered relevant.
Similarly, we do not include devices, like smart tables, that people can use to enhance group activities \cite{Morris2006MediatingDesign, Buisine2012HowCollaboration, zumHoff2022InteractiveHome}, as long as the device does not make own decisions on interventions. 
We do not include non-actuated embodiments of video-mediated communication tools (e.g. \cite{Cesta2016Long-TermStudy, Judge2010a, Tang2013, Weisswange2023TelepresenceAdults}) or smart voice assistance (e.g. \cite{Lee2020HeyHarmony, Maier2020IDNMediation}) even if they discuss effects on group qualities.
However, we do consider work where a group and the interaction with it is distributed in space (or time), and if a system is composed of multiple devices as long as there is a common control or design concept.

Our search consisted of articles with titles containing the following combinations of keywords, informed by prior work in HRI and group research:
\begin{itemize}
    \item (robot OR embodiment OR embodied OR agent) AND human AND (group OR team OR crowd)
    \item (robot OR embodiment OR agent) AND (interpersonal OR "multi-human" OR "social dynamics" OR "human-human-interaction" OR “multi-party”)
    \item (robot OR embodiment OR embodied OR agent) AND group AND (influence OR change OR impact OR improve OR intervention OR dominance OR participation OR balance OR conflict OR engagement)
    \item (robot OR embodiment OR agent) AND (mediation OR mediator OR mediate OR coach OR counselling OR counselling OR "conflict resolution" OR facilitation OR facilitator OR moderator OR moderation OR liaison OR inclusive OR conciliation OR "group therapy" OR "social networking" OR arbitration) 
    \item (robot OR embodiment OR agent) AND (cohesion OR entitativity OR "group dynamics" OR "group affect/climate/satisfaction" OR "social dynamics" OR prosocial OR "group affect" OR "group balance")
    \item "human-human-robot interaction" 
    \item "human-robot interaction" AND (multiple OR multi-agent)
    \item group AND ("human-computer interaction" OR "human-machine interaction" OR "human-agent interaction")
    \item social AND catalysts AND (robot OR agent)
\end{itemize}
We added the full set of references of prior reviews related to robots and groups (see \cite{Chita-Tegmark2020ElderlyMediationReview, Dahiya2023, Javed2023GroupDynamics, Oliveira2021TowardsBehaviour, Schneiders2022Review, Sebo2020, Weisswange2023, Wisowaty2019GroupReview}).
In addition, we considered papers from the proceedings of the RO-MAN 2023 conference, which took place in August 2023 but whose proceedings were not yet accessible online at the time of search. We included all papers matching the following keywords: 

human-human OR group OR team OR mediat* Or moderat* OR facilit* OR multi OR prosocial OR coach OR counsel.

Our search was performed by two of the authors and the papers were collected through the Mendeley reference management software \footnote{https://www.mendeley.com/}. The software was also used for automatic removal of double entries after merging results from the different search processes.
Three authors initially evaluated search results based on title and abstract, applying the inclusion and exclusion criteria. Papers accepted by at least two authors proceeded to the next stage, while those accepted by only one author were reevaluated. The authors then reassessed the inclusion criteria based on full text, ensuring each paper was reviewed by at least two authors. Any papers accepted by only one author were reevaluated. Moreover, they searched the references of accepted papers for additional work, applying the same process as above. From the final pool, papers reporting on the same underlying mediation concepts were grouped together.

\section{Results} \label{sec:results}

\begin{figure}
     \centering
     \includegraphics{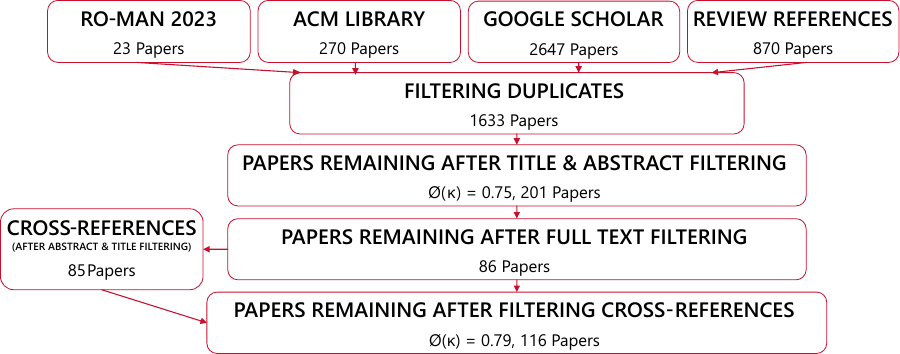}
     \caption{Search statistics}
     \label{fig:Numbers}
     \Description{Figure 2. Overview about the number of papers found through the stages of the search process - fully described in the text.}
\end{figure}

Figure \ref{fig:Numbers} shows the statistics along our filtering process.
Our original search produced 2647 hits on Google Scholar and 270 on the ACM Library, together with 870 references from the relevant review publications and 23 additional papers published at RO-MAN 2023. 
After filtering duplicates our database contained 1633 unique entries.
Filtering using our inclusion criteria on title and abstract of the papers left us with 201 publications. 
The average agreement on inclusion between each pair of authors is described by a Cohen’s kappa of $\kappa_{\varnothing} = 0.75$. Of these, 178 papers received at least 2 votes and 47 papers received only a single vote. 
Upon reevaluation, the authors agreed to accept 23 of these papers for the next stage and rejected 24. These 201 papers proceeded to the next stage, where we evaluated the full text of these papers for if the inclusion criteria still hold. This was true for 100 papers, with 27 receiving only a single vote. After discussion of those papers, we agreed to keep 13 and reject the remaining 14, leaving us with 86 papers.
To make sure to include all relevant publications, we compiled the list of references of all selected papers and performed another duplicate check and title and abstract evaluation.
This created a second selection of 85 papers for which two researchers again evaluated inclusion based on full text. We found another 30 unanimous includes. 
Three papers were put up for a discussion, of which one was kept, and the others rejected.
The average agreement on inclusion between each pair of raters for the second stage of filtering is described by a Cohen’s kappa of $\kappa_{\varnothing} = 0.79$.
Rejections of full text papers were classified according to the dominant inclusion criterion which was not fulfilled: number of humans >=2: 38; reacting embodiment: 24; influencing a group: 81; dedicated role: 18; other\footnote{review or opinion paper, duplicate, full text unavailable}: 9. 

Final analysis was performed on the combined selection of 116 papers, in which we found 89 unique robotic social mediation concepts, where the remaining papers mostly represent early reports of concepts that appears again in later publications\footnote{in contrast, \cite{Dorrenbacher2023} introduces three concepts that we consider within a single paper}.

\begin{figure}
    \centering
    \includegraphics{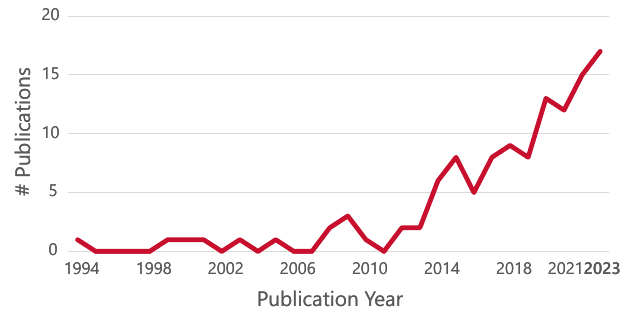}
    \caption{Development of number of scientific publications in the area of group mediation (116 papers overall).}
    \label{fig:PubYears}
    \Description{Figure 3. Plot of the publication dates of the 116 publications covered in this paper. The plot shows a few individual papers published between 1994 and 2010. Beginning in 2011, the number keeps rising almost linearly with every year.}
\end{figure}

When looking at when these 116 papers have been published (Fig. \ref{fig:PubYears}, we see a substantial increase in work on robotic group mediation. This follows a general trend of increased work on robotic interaction with groups that has been reported by previous work \cite{Schneiders2022Review}.

\subsection{Input Factors}
\begin{figure}
    \centering
    \includegraphics{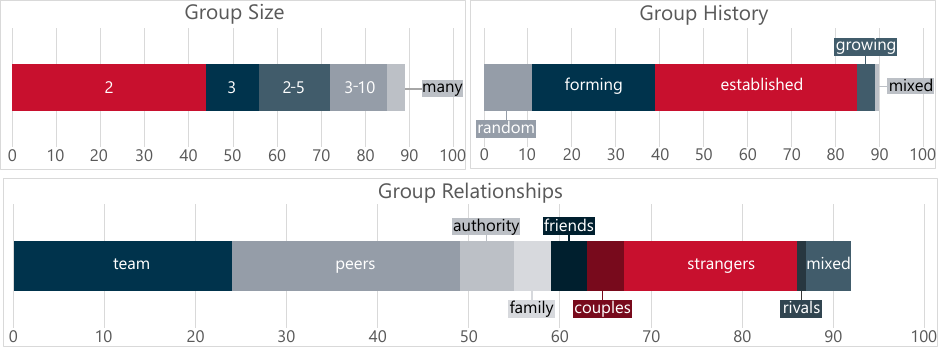}
    \caption{Statistics of target group sizes (top left), group histories (top right) and member relationships (bottom) among the social mediation concepts}
    \Description{Figure 4. Histograms of the use of three input factors -- group size, history and relationships -- among the reviewed mediation concepts. For group sizes, the fractions are: 49\% dyads, 13\% triads, 18\% small groups of 2 to 5 members, 15\% medium-size groups of 3 to 10 members, and 4\% of bigger groups. 
    For group histories, the fractions are: 51\% are established groups, in 31\% the group is newly forming	, 4\% use growing groups, 12\% are random, and 1\% have members with mixed history.
    For group relationships, the fractions are: 27\% peers, 26\% team, 19\% strangers, 5\% mixed, 7\% authority, 4\% each for family, couples, and friends, and 1\% rivals.}
    \label{fig:Stats_SizeHistory}
\end{figure}

\textit{\textbf{Individual factors}}:
\begin{itemize}
    \item \textit{Age}: About two thirds of the concepts are designed for adult participants or did not specify an explicit target age group, 21\% consider children and teenagers and the remaining ones target older adults, young children (<6y) or inter-generational groups.
\end{itemize}
\textit{\textbf{Group factors}}:
\begin{itemize} 
    \item \textit{Group size}: Figure \ref{fig:Stats_SizeHistory} left shows the distribution of target group sizes. Half of the concepts are designed for the special group size of two persons. Groups of exactly three members (12) can be found almost as often as concepts for small groups (16) and medium sized groups (13). The remaining 4 papers consider potentially larger crowds. Only a single concept explicitly addresses dynamically changing group sizes \cite{Zheng2005DesigningGuide}.
    \item \textit{Group composition}: Seven papers conceptualized mediators for groups including members with special health conditions \cite{Birmingham2020, Moharana2019RobotsCaregivers, Neto2023TheChildren, Pliasa2019CanDevelopment, Scassellati2018ImprovingRobot, Shim2017AnEvaluation, Shimoda2022ApplicationRehabilitation}. Other group compositions that are addressed  involve specific skills and capabilities, such as diverse language proficiency \cite{Cumbal2022ShapingBackchannels, Li2021InfluencingTeams} or low social skills \cite{Hayamizu2013AnControl}. Furthermore, we find groups with mixed ages \cite{Chen2022DesigningInterviews, Scassellati2018ImprovingRobot, Shim2017AnEvaluation, Sinnema2019TheInteraction} or cultural background \cite{Aylett2023, Isbister2000HelperSpace}.
    \item \textit{Relationships}: Looking at relations between group members (Fig. \ref{fig:Stats_SizeHistory} center), robotic mediation concepts follow classical small group research in that they focus on work relations as in peers or teams. The only other strong cluster considers people unknown to each other before entering the mediation situation. 
    \item \textit{Group structure}: Along the same lines, most of the group structures considered in the evaluated publications are assumed to have been established already at the time of mediation (Fig. \ref{fig:Stats_SizeHistory} right). Other significant categories are groups that are newly forming, and short-lived groups that neither existed before nor are supposed to exist after the considered interaction (“random” in Fig. \ref{fig:Stats_SizeHistory} right).
\end{itemize}

\begin{figure}
    \centering
    \includegraphics{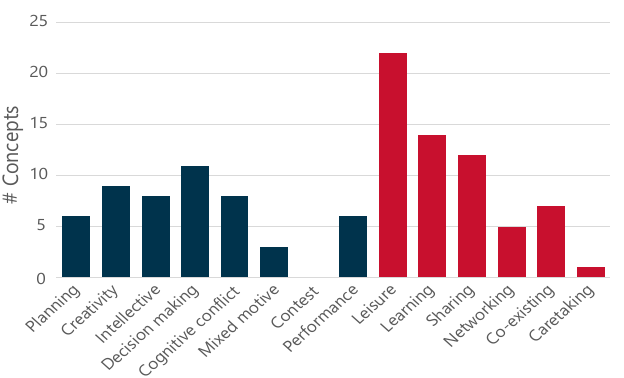}
    \caption{Statistics of task types among the social mediation concepts. Blue bars are categories based on \cite{ McGrath1984Groups:Performance}, gray bars are social task types we have defined in this paper.}
    \Description{Figure 5. Histogram of the use of task types among the reviewed mediation concepts. 
    Among classical task, there are 6 Planning Tasks, 9 Creativity, 8 Intellective, 11 Decision making, 8 Cognitive conflict, 3 Mixed motive, 6 Performances, and none in Contests. The fractions among the new task categories from this paper are: 22 Conversation, 14 Learning/Teaching, 12 Sharing and Information Exchange, 5 Networking and Bonding, 7 Co-existing, one Care-taking, and no paper that does not use any task.}
    \label{fig:Stats_Tasks}
\end{figure}

\textit{\textbf{Environmental factors}}:
\begin{itemize}
    \item \textit{Task}: It is not always very clear, if the task that was selected in a study does necessarily constrain the mediation concept and many cases involved multiple tasks. 
    Looking at the distribution of tasks (Fig. \ref{fig:Stats_Tasks} left), however, we find an almost equal balance between classical tasks \cite{McGrath1984Groups:Performance} and what we call social tasks. 
    The dominating task is one of the latter, 20\% of the concepts work with leisure and conversation settings, where the goal of the group members is the interaction itself. 
    Support of groups performing care-taking \cite{Moharana2019RobotsCaregivers} was only investigated in a single study and we did not find any research on mediating groups engaged in a competition.
    \item \textit{Setting}: In terms of environment setting, the constrained setup with people sitting or standing in a room, typically around a table or in front of a screen or stage, is dominating (57\%). This is not too surprising as tabletop robots (32\%) are frequently used, which impose certain limitations on the setup. The most complex embodiments, humanoid robots, were used in 27\% of the papers. However, in most papers it was not clear, if the embodiment is part of the core mediation concept, or just part of the specific implementation used in the studies.
    \item \textit{Temporal conditions}: Time pressure is part of some of the experimental tasks, for example to increase the chance for errors and conflicts \cite{Jung2015UsingViolations}, or simply used to limit experiment duration. However, implications for the concepts have not been explicitly mentioned. A regularly re-occurring group situation was part of the longer-term mediation concept in \cite{Chen2022DesigningInterviews}.
    \item \textit{Embedding}: Almost all reviewed papers neglect the social and cultural embedding when describing their concepts. An exception is the work from Kang et al. \cite{Kang2023TheGroup} who explicitly mediate adherence to (Korean) cultural norms.
    \item \textit{Resources}: The concept in \cite{Jung2020Robot-assistedGroups} selects interactions to influence group processes that rely on resource sparsity. Otherwise, specific resource conditions mostly play a role for concepts involving resource conflict \cite{Druckman2021WhoHumans, Dorrenbacher2023,Kanda2012ChildrenRobot, Kang2023TheGroup, Shen2018} and resources about the individual members for the use of a mediator to work on shared interests \cite{Fu2021UsingConversations, Kochigami2018DoesCommunication, Ono1999, Shin2021, Uchida2020ImprovingExperience}. Many experiments additionally use hypothetical scenarios for decision-making tasks (e.g. the "desert survival" task) that involve resource limitations. 
\end{itemize}

\subsection{Group Processes}
\begin{figure}
    \centering
    \includegraphics{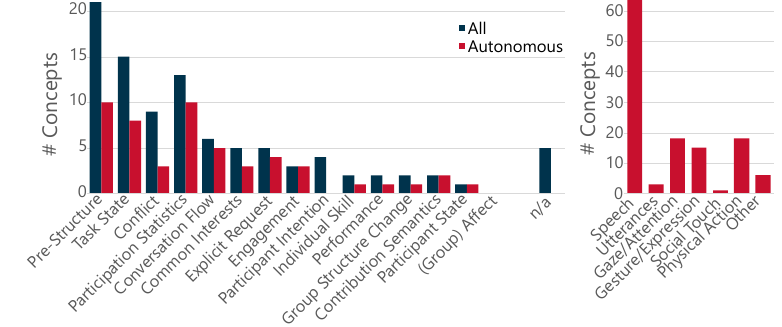}
    \caption{Left: Intervention triggers used among all concepts (blue) and limited to those demonstrated on an autonomous agent. Right: Statistics of mediator modalities among the social mediation concepts.}
    \Description{Figure 6. Histograms of the intervention triggers and the mediator modalities used in the reviewed mediation concepts. 
    For intervention triggers, the figure additionally shows the number of those concepts that actually implemented the triggers for the experiments. The counts are respectively for all and only implemented triggers: 21 and 11 using Pre-determined Structure, 15 and 7 using Task State, 9 and 2 using Conflict Detection, 13 and 3 based on Participation Statistics, 6 and 5 measuring Conversation Flow, 5 and 5 using Common Interests, 5 and 3 detecting an Explicit Request, 3 and 3 evaluating Engagement, 4 and 1 using Participant Intentions, 2 and 2 each using Individual Skill or Task Performance, 2 and 1 based on Group Structure Changes, 2 and 0 Contribution Semantics, 1 and 1 Participant State, none using (Group) Affect, and 5 and 3 not reporting any trigger condition.}
    \label{fig:Stats_Mediator}
\end{figure}

Few papers explicitly evaluated specific group processes, but we can assess relevant dynamics through the description of what triggers an intervention within the mediation concepts (Fig. \ref{fig:Stats_Mediator} right). 36 of the concepts rely on either pre-determined timing or the progression of a specific task to determine when to initiate a mediation behavior. \begin{itemize}
    \item \textit{Interpersonal processes}: The most common measures were the detection of conflicts, for example through speech volume \cite{Hoffman2015, Shim2017AnEvaluation}, or a person interfering with another person’s task \cite{Kanda2012ChildrenRobot} (~10\%), and the distribution of participation among the group members (14\%). 
    Other related trigger conditions revolved around quality measures of conversation flow and the detection of common topics of interest between multiple members of the group. 
    \item \textit{Individual-level processes}: With respect to individual-level processes, we find examples using the evaluation of a team member’s skill, intentions, or engagement. 
    Surprisingly, we did not find any mediation concept explicitly reacting to individual or group affective states.
    \item \textit{Group-level processes}: Two papers \cite{Kochigami2018DoesCommunication, Zheng2005DesigningGuide} incorporated group-level process, specifically, changes to the group structure through an additional member joining, in their concept.
\end{itemize}  

Figure \ref{fig:Stats_Mediator} right shows the statistics of studies automatically detecting the behavior triggers in a way that the robot could operate autonomously in red. 
Many more concepts instead described a coarse behavior pattern of a human experimenter controlling the robot in wizard-of-oz type studies or hypothetical trigger conditions formulated as part of a concept design. 

\subsection{Mediator Input Factors \& Mediation Processes}
\textbf{\textit{Mediation input factors}}:
\begin{itemize}
    \item \textit{Role}: Three quarters of all concepts (74\%) describe the robot in a facilitator role, as compared to a group member or bystander. This is certainly also a result of our inclusion criteria, as we required the robot to have a specific role different from the human group members.
    \item \textit{Reputation/Relation}: We restrict our analysis to concepts that use explicit interventions, whose success might be influenced by the group's perception of the mediators background and intention. We evaluate how either the experimenter or the robot itself framed its background and role. Only a quarter of the relevant concepts included a framing of the mediator's targets in relation to the group, in most cases as either neutral (5 concepts) or aligned with the group (6). 31\% of the considered concepts introduced some form of reputation of the robot before interacting with the group, either as having expertise with mediation (9 concepts) or with the task at hand (7). One paper explicitly framed the robot as being a novice \cite{Kim2023ChildrobotBehaviors}.
    The mediator being a robot might already imply certain expectations by the group members which can mimic a reputation \cite{Dou2020UserStudy}. However, studies that analyzed the perception of the robot by the participants only did so after the interaction.
\end{itemize}

\begin{figure}
    \centering
    \includegraphics{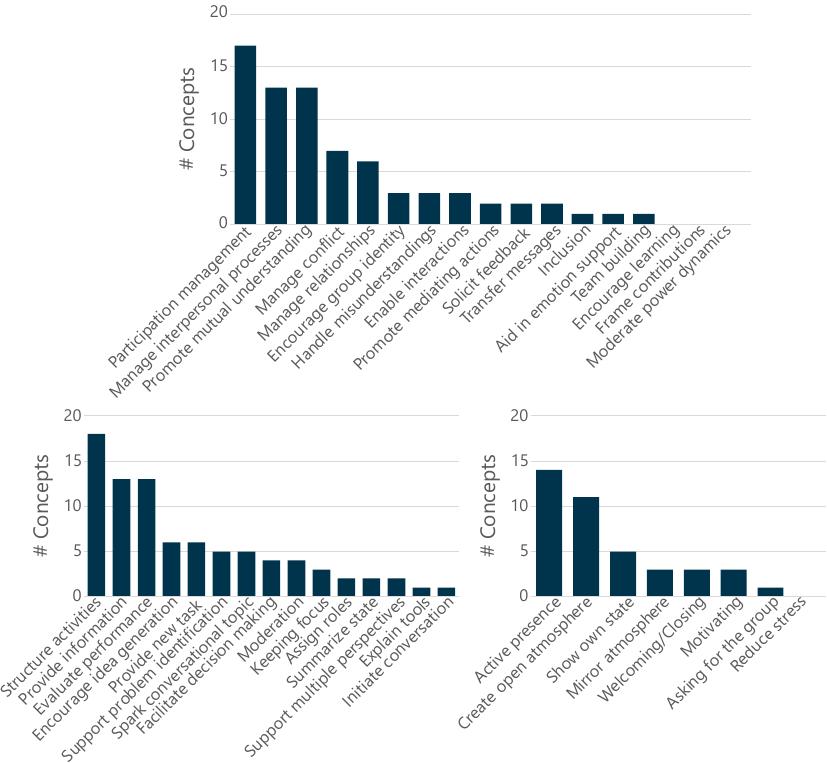}
    \caption{Statistics of mediator interventions among the social mediation concepts. Interactional (top), contextual (bottom right), and task (bottom left) interventions.}
    \Description{Figure 7. Histograms of the mediator intervention types proposed in the reviewed concepts. The figure separately reports numbers for interactional, contextual, and task interventions. For interactional, these are: Participation management 17, Manage interpersonal processes13, Promote mutual understanding 13, Manage conflict 7, Manage relationships 6, 3 each for Encourage group identity, Handle misunderstandings, and Enable interactions, 2 each for Promote mediating actions, Solicit feedback, and Transfer messages, one each for Inclusion, Aid in emotion support, and Team building, and none for Encourage learning, Frame contributions, and Moderate power dynamics. For task interventions the numbers are: Structure activities 18, Provide information or Task support 13, Evaluate performance 13, 6 for both Encourage idea generation and Provide new task, 5 for both Support problem identification and Spark conversational topic, 4 for both Facilitate decision making and Moderation, 3 times Keeping focus, 2 each for Assign roles, Summarize state/progress, and Support multiple perspectives, and one each for Provide and explain tools and Initiate conversation. Contextual interventions are distributed as: Active presence 14, Create open atmosphere 11, Show own state 5, 3 each for Mirror atmosphere, Welcoming/Closing, and Motivating, one for Asking for the group, and none for Reduce stress.}
    \label{fig:Stats_Interventions}
\end{figure}

\textbf{\textit{Mediation processes}}:
\begin{itemize}
    \item \textit{Mediation Style}: In 52 concepts, mediation is framed and performed explicitly – the robot's goals and actions for influencing group processes are transparent to the group members. 
    However, we also find a larger number of concepts (25) with more subtle, implicit mediation actions, for example by adjusting a task or showing active listening behavior as a group member. 
    In the remaining cases the robot performs very general interventions to change the group state without looking at what the current state and processes of the group is, or concepts where the mediation was rather a side effect \cite{Fu2023TheCollaboration, Jung2020Robot-assistedGroups}. 
    \item \textit{Task interventions}: For task-directed mediation behaviors (Fig. \ref{fig:Stats_Interventions}, bottom left), many concepts included the structuring of task activities, help with the task (through action or information), or the evaluation of the group’s performance.
    \item \textit{Interactional intervention}: More than 70\% of the concepts included at least one interactional intervention, that is, explicitly selected behaviors that target processes between team members. About half of those additionally integrated task-focused interventions.
    Frequent intervention types working on interaction processes include participation management, addressing ongoing interpersonal activities, and promoting mutual understanding (see Fig. \ref{fig:Stats_Interventions}, top).
    \item \textit{Contextual interventions}: We also found a number of behaviors targeted at changing the context for the group, for example through contributing to a positive atmosphere. As many as 14 concepts proposed interventions with which the robot actively made its presence known, for example by continuous breathing motions (e.g. \cite{Takano2009PsychologicalNeeds}) or by performing active listening behaviors such as confirmatory utterances or explicit eye contact with the speaker \cite{Erel2021, Kobuki2023RoboticStudy}, although this is often mainly used to establish the robot as a social agent.
    \item \textit{Modalities}: In more than two thirds (74\%) of the concepts, the robot mediator uses speech as one of its modalities for intervention, in fact in 47\%, speech is the only modality that is explicitly mentioned. Figure \ref{fig:Stats_Mediator} right shows the details about mediation modalities. 

\end{itemize}

\subsection{Outcome Factors}
\begin{figure}
    \centering
    \includegraphics{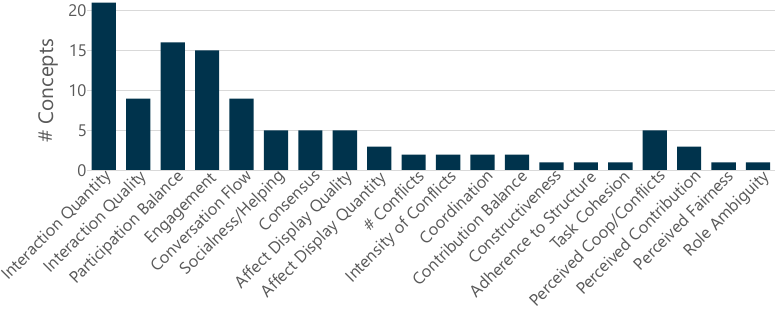}
    \caption{Statistics of the proxy factors used as part of the target outcomes across the social mediation concepts.}
    \Description{Figure 8. Histogram of the proxy factors used in the reviewed concepts. The number of uses are: Quantity of Interactions 21, Quality of Interactions 9, Participation Balance 16, Engagement 15, Conversation Flow 9, each 5 times for Socialness of behavior/Helping, Consensus, and Quality of Affect Displays, 3 times Quantity of Affect Displays, 2 each of Number of Conflicts, Intensity of Conflicts, Cooperation/ Synchronization, and Contribution Balance, one each of Constructiveness of Behavior, Adherence to Structure, and Task Cohesion, as well as Perceived Cooperation/Conflicts 5, Perceived Contribution 3, Perceived Fairness 1, and Role Ambiguity1.}
    \label{fig:Stats_Proxy}
\end{figure}

We also analyze which outcomes the concepts target through their mediation processes. 
As discussed above, we discriminate between individual, group and performance outcomes which can be considered results of group processes, and proxy factors which likely have a correlation with such outcomes but are rather quantities that can be measured on the group processes.
We find 17 concepts that are solely defined through such proxy factors. Overall, 80\% of the concepts use these measures as part of their goals. 
The most prominent factor is the number of interactions between group members, followed by individual engagement in the task and participation balance (Fig. \ref{fig:Stats_Proxy}).  

\begin{figure}
    \centering
    \includegraphics{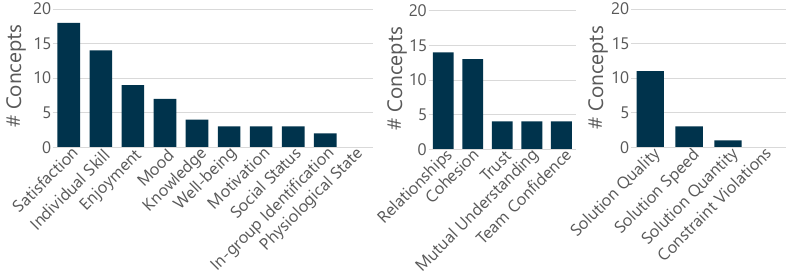}
    \caption{Statistics of occurrence of target outcomes across the social mediation concepts. Individual (left), group (center) and task-related (right) outcomes.}
    \Description{Figure 9. Histogram of target outcomes mentioned in the reviewed concepts. For individual outcomes, the numbers are: Satisfaction 18, Individual Skill 14, Enjoyment 9, Mood 7, Knowledge 4, Well-being 3, Motivation 3, Social Status 3, In-group Identification 2, and Physiological State 0. For group outcomes, we report: Interpersonal Relationships 14, Cohesion 13, Trust/Team Climate 4, Mutual Understanding 4, Team Confidence 4. For performance outcomes, the numbers are: Solution Quality 11, Solution Speed 3, Solution Quantity 1, Constraint Violations 0.}
    \label{fig:Stats_Outcomes}
\end{figure}

Among the outcome measures, concepts focusing on individual states and changes play the biggest role (Fig. \ref{fig:Stats_Outcomes}). 
Around 15\% of the concepts include a performance target, but almost all work (80/85) includes at least one of the individual and group measures.
Only three concepts explicitly considered all three types of outcomes \cite{Fucinato2023CharismaticCreativity, Jung2020Robot-assistedGroups, Shamekhi2019}, while two more are using a proxy for one of the outcome types \cite{Jung2015UsingViolations, Neto2023TheChildren}. 
Twelve papers use only individual outcomes, even if considering potential proxy factors are considered. 
Common individual outcomes that are targeted through the mediation interventions are satisfaction, skill development and enjoyment. 
Group outcomes are mainly represented through formulations of cohesion or interpersonal relationships.

\subsection{Implementation}
As discussed before, this review is not primarily reporting experiment setup or results of the discussed work. 
However, we did analyze the “maturity” of the concepts based on if and how they were implemented for possible experiments. We also looked at differences between input factors in the original concept versus those actually tested in the experiments. Details about our categorizations can be found in appendix \ref{Appendix:ExperimentList}.
16 papers only describe design concepts; 14 groups report a first proof of concept implementation. 
The majority (49) did report user experiments in a lab setting, but 10 groups even tested their mediation concepts in an environment close to its target setting. 
Almost half of the lab studies (20) and a third of the field studies (3) worked with the wizard-of-Oz method \cite{Steinfeld2009TheResearch, Kelley1984WoZ}, where a human controls the robot behaviors.
20\% of the experimental settings (of the 68 approaches that performed a study with participants) had input conditions not matching the original concept definition (and another 17\% only partially matched). 
In most cases, the concept has been designed to work with established teams or groups with existing relationships, while it was applied to groups of non-related people (often students). 
18\% of the groups testing concepts with users did not run any type of control group.
Concerning measures, we found that almost every paper uses an individual way to assess target outcomes. 
Among the work that reported using, or at least adapting, existing scales or behavior annotation schemes, we find 39 different sources, only 4 of which have been used in more than one paper (for details, see Tables \ref{tab:allpapers_experiments}-\ref{tab:allpapers_experiments12} in the appendix). The only scale that has been used more than twice is the Inclusion of Other in the Self scale \cite{Aron1992InclusionCloseness} (6 times).
Results are slightly more comparable when looking only at reported proxy factors, such as participation statistics or more objective ones such as task performance.
24 authors report finding significant effects of their mediation concepts, albeit with respect to different types of baselines. 17 of those used objective measures or constructs that have been validated or at least demonstrated internal consistency.

\subsection{Mediator Concepts}
This section introduces some details about the concepts found in our selection of papers and highlights common themes in the types of social mediator concepts that we reviewed.
A table with all the details of all papers can be found in the appendix \ref{Appendix:AllPapers}.

Many approaches provide (sub-)tasks to a group or contribute to performing the task in a way to foster working together \cite{Alves-Oliveira2019EmpathicStudy, Fan2021FieldAdults, Gillet2020, Jung2020Robot-assistedGroups, Kim2023ChildrobotBehaviors, Short2017}. For example, the robot in \cite{Alves-Oliveira2019EmpathicStudy} selects task actions that should trigger discussion among the group members.
One concept proposed a robot that learns when to encourage and reward interpersonal interactions to increase participation \cite{BaghaeiRavari2021EffectsLearning}.
In two papers \cite{Charisi2021TheDynamics, Shen2018} a robot proposed ways to play together with children, while the concept in \cite{Scassellati2018ImprovingRobot} provides guidance for mutual gaze to enable cooperation. 
Stoican et al \cite{Stoican2022LearningCollaboration} propose a concept, where the impression of and trust between one group member and another one is improved through robot interventions with the second person.

An important intervention for improving collaboration within a group is the informed assignment and communication of roles. The transactional leader robot presented in \cite{Lopes2021SocialTeams} uses role assignments to structure a group's task activities.
Similarly, this is also an approach used for the concept of learning by teaching, which is supported with the mediator robots presented in \cite{Chandra2016ChildrensRobot, Edwards2018ALearning, Mitnik2008AnMediator}. 

The learning and teaching scenario is also addressed in \cite{Gordon2022InvestigationAgent} but includes many additional interventions to encourage participation and discussion and the convergence to a joint answer.
Whereas in many of the above groups, cooperation was part of approaching a task, this paper and many other concepts work in particular on groups that need to come to a joint decision to progress in the task.
The robot usually tries to structure the decision-making process and encourages the discussion of options \cite{AlMoubayed2013TutoringTutor, Basu2001TowardsSettings, Buchem2023Scaling-UpNAO, Alves-Oliveira2019EmpathicStudy, Gillet2022Ice-BreakersTeenagers, Rosenberg-Kima2020Robot-supportedFacilitation, Shamekhi2018}.
Sometimes the focus is only on the moderation of these discussion, trying to ensure balanced turn-taking \cite{Fu2017TurnTurn-Taking, Hitron2022AIRobots, Li2023ImprovingGroups, Shamekhi2019, Tennent2019}.
Shen and colleagues \cite{Shen2017RobotConformity} discuss interventions to bias decision making towards certain members using behavioral mimicry to enhance conformity of a group. 
Combining the need of a joint decision with a first phase of generating ideas drives the concept presented in \cite{Ohshima2017Neut:Conversations} using convergence and divergence promoting interventions.

Multiple concepts focus explicitly on this latter part of mediating open discussions and the generation of ideas (e.g. \cite{Ikari2020Multiple-RobotDiscussion, Chew2023WhoInteractions}).
In some cases, robots try to establish a positive and motivating atmosphere to facilitate creativity \cite{deRooij2023Co-DesigningDynamics, Fucinato2023CharismaticCreativity}.
The robot in \cite{Cumbal2022ShapingBackchannels} mainly aims at a balance of participation in a creative game setting.

Another cluster contains concepts proposing mediation of conflict situations. 
Seven concepts work with cognitive and/or verbal conflicts \cite{Hoffman2015, Sadka2023AllConflict, Stoll2018KeepingFormat, Jung2015UsingViolations, Wang2023ExploringConflict, Dorrenbacher2023, Ablett2007BuildBot:Teams}.
Interventions vary greatly, including raising awareness of the conflict \cite{Hoffman2015, Jung2015UsingViolations}, trying to promote empathetic behaviors \cite{Sadka2023AllConflict, Wang2023ExploringConflict}, or attenuating conflict strength \cite{Stoll2018KeepingFormat, Ablett2007BuildBot:Teams}.

Three papers focus on a conflict of resources.
One such scenario is that of common negotiation mediation tasks, as in a company de-merger \cite{Druckman2021WhoHumans} using the structure provided by professional tools.
The other two papers work on social resolutions of object possession conflicts in children either by scolding \cite{Kanda2012ChildrenRobot} or using integrative pedagogical approaches \cite{Shen2018}.

We also find concepts in which the mediator aims to motivate a group to perform a joint task while keeping up high engagement \cite{Lopes2021SocialTeams, Leite2016AutonomousInteraction, Fu2023TheCollaboration}.
A good example of the first type is presented in Lopes et al. \cite{Lopes2021SocialTeams}, who discussed leaderships style of robots. 
The main role was to introduce and motivate team members through a collaborative task using transactional or transformational leadership styles \cite{Bass1985Leadership}. 
While the former type of interventions resulted in better task performance in the experiments with real work teams, the latter improved engagement, in line with previous results from group research.

A related type of mediation concepts focuses on interventions to increase enjoyment or satisfaction of a group during a joint activity.
The robot can be the lead of such an activity, for example as a tour guide in a museum, encouraging interaction between group members \cite{Yamazaki2012AVisitor, Zheng2005DesigningGuide, Sinnema2019TheInteraction}.
In other cases, it acts as a special participant of a game, trying to activate participation from "inside" \cite{Matsuyama2010PsychologicalGame, Traeger2020}.
In two concepts, the robot acts as an intermediate between a leading human and the rest of the group \cite{Karatas2020UtilizationArt, Shimoda2022ApplicationRehabilitation}.

There are also a few papers on robots supporting the affective quality of human-human conversations.
Erel and colleagues \cite{Erel2021} use a peripheral robotic object (same as in \cite{Hoffman2015}) which supports empathetic sharing of problems between two persons through supportive gesturing. The robot’s behaviors were built based on a pilot study in which a professional therapist controlled the activity in such a situation. The robot itself was introduced as a side-participant, but a third of the participants of their study reported no influence of its presence. Nevertheless, it was found that the mediating robot had a positive influence on multiple group measures such as perceived quality of emotional support and interpersonal coordination.

Another concept involves a robot mediating human-human communication in a bystander role \cite{Takano2009PsychologicalNeeds}. The authors take inspiration by the positive influence of the “Chameleon Effect”, mimicking another person’s expression, posture, etc. during interactions, increases interpersonal evaluation of communication partners \cite{Chartrand1999Chameleon}. 
A common theme across all these papers are robot behaviors that actively show its presence and listening. It seems that a positive effect can be reached, even if the mediator role is not made explicit.

Some mediation concepts encourage sharing of own states and problems with others to allow empathetic and supportive responses from the group \cite{Aylett2023, Dietrich2022, Kobuki2023RoboticStudy, Noguchi2023HowRecipients}.
Only a single paper is working explicitly on mediation in support groups -- \cite{Birmingham2020} proposed a robotic system interacting with such a group to increase interpersonal trust. 

The robots presented in \cite{Hasegawa2014FacilitationExaggeration, Nagao1994SocialAgents} all help group members to become aware of each other’s intentions to ease interactions.

Five concepts discuss artificial agents that try to help a group of people to follow societal norms.
They either consider conversational norms and providing feedback when behaviors of one group member should be adjusted towards it \cite{Shim2017AnEvaluation, Tahir2020AConversations}, or societal norms, such as offering public transport seats to those in need or adjusting music volume to acknowledge the presence of others (two concepts, SEYNO and SEATY, presented in \cite{Dorrenbacher2023}).
Kang and colleagues present a robotic table concept that tries to ease people's burden of estimating other's expectations to follow social interaction norms \cite{Kang2023TheGroup}.
Robots can also help training and improving interactions skills within a group \cite{Chen2022DesigningInterviews, Utami2017, Utami2019CollaborativeRobot}.
Three concepts enhance group identity and offer opportunities for shared experiences and ideas \cite{Fu2021UsingConversations, Jeong2018, Moharana2019RobotsCaregivers}.
Fribo \cite{Jeong2018} is a distributed embodied mediator for groups of friends that live remote of each other. 
It tries to prevent a weakening of interpersonal relationships and cohesion in such groups through providing opportunities for shared experiences of daily activities. 
There is also a number of concepts where the robot more generally influences group dynamics to positively influence group cohesion. 
This is approached by robots acting as team members but providing interventions like showing vulnerability and trust \cite{Traeger2020}, or asking members to reflect on their relationships \cite{Strohkorb2016ImprovingRobot}, or by helping to balance the feeling of contributing to a task \cite{Short2017}.

Facilitation by social robots for establishing a new group is a widely prevalent theme in the body of literature. 
Such formation robots are tasked with conducting behaviors to foster connections among individuals by highlighting commonalities and providing an environment that encourages conversations and interactions \cite{Birmingham2020, Hayamizu2013AnControl, Kim2023ChildrobotBehaviors, Nakanishi2003CanCommunities, Zhang2023}. 
Isbister and colleagues additionally consider cultural background when selecting common conversation topics \cite{Isbister2000HelperSpace}.

In related concepts, the robot only mediates a first introduction between people, while the group context exists for only a limited time (e.g. one conversation) \cite{Sadka2022ByOpening-encounters, Takeuchi2014Whirlstools:Affordance, Uchida2020ImprovingExperience, Xu2014}.
A first challenge to be mediated is the detection \cite{Xu2014} or arrangement \cite{Ono1999} of people in spatial proximity necessary to enables social interaction.

Two teams of researchers developed communication robots that try to influence ongoing conversations in a way to integrate bystanders \cite{Inoue2021AParticipants, Matsuyama2015Four-participantParticipant}. In both concepts, the robot participates in the conversation and tries to create an opportunity to join in for a person currently not participating.
In \cite{Sebo2020TheBehavior}, the related target is to integrate an outgroup member during a decision-making task of an established group.
Three concepts explicitly addressed inclusion of members with different backgrounds into groups by encouraging and balancing participation across all group members \cite{Gillet2020, Neto2023TheChildren, Pliasa2019CanDevelopment}.

\section{Insights}
In light of the findings from Section \ref{sec:results}, we now address the key questions we previously identified. Given the available literature, not all these questions can be answered to the same extent. Nonetheless, we try to provide insights that we think are crucial for advancing our understanding of Social Mediation Robots and provide a valuable guide to researchers looking to design studies targeting social mediation through robots in group settings.

\subsection{What specific roles can robots play in shaping group processes?} \label{subsec:concepts}
To understand the specific roles played by Social Mediation Robots in shaping group processes, we sought to extract the important categories of mediation found in the reviewed literature. To this end, we leveraged the Mediation I-P-O model to highlight key dimensions of the identified mediation categories. In addition, all authors collected different role descriptions for humans working with groups and drew from existing taxonomies for teams, groups, and their relevant dynamics and processes~\cite{Hollenbeck2012BeyondDescription, Lickel2000, Marks2001AProcesses, Aldag1993BeyondProcesses,Tuckman1965DevelopmentalGroups, Tuckman1977StagesRevisited, Bostrom1993GroupSystems, Clawson1993TheMeetings, Dickson1996FacilitatingEnvironment, Jones2001FacilitatingField, McFadzean2002DevelopingCompetencies, Jones2001FacilitatingField, Yukl2002AResearch, Kressel2006MediationRevisited, Moore2014Mediation}. This produced 11 distinct robot mediation categories that researchers may target when conducting studies on social mediation through robots. Some categories have been explored more extensively than others, yielding detailed insights and revealing understudied areas that may motivate promising directions for future work.

\subsubsection{Foster Collaboration}
Concepts in this category feature mediator interventions that help a group improve interaction processes while performing a task with a shared goal, where the goal cannot be accomplished by an individual alone. This may either be due to (physical) task requirements (e.g. lifting a heavy object) or the need for a set of skills that only emerges through the combined abilities of group members. Such tasks are often performance-oriented and assigned to newly forming or already established teams. 

In terms of the input factors, all papers from this category involved very small groups of two or three members. Although the concept of fostering collaboration could be meaningful for larger teams, as in search and rescue teams or workers at a construction site, the insights from the current work may or may not scale to larger groups.
    
Mediation factors include mediator roles, which may include that of a leader. In this case, the mediator assigns roles, highlights potential of and need for working together, helps people structure their individual (inter)actions (as in \cite{Fan2021FieldAdults, Gillet2020AChildren, Kim2023ChildrobotBehaviors}), and evaluates task performance \cite{Lopes2021SocialTeams}. 
Relevant skills for robots aiming at fostering collaboration include task analysis, task scheduling, and understanding the capabilities of individual members. 
Several concepts also use interventions to explicitly manage the contributions and interactions between all members \cite{Alves-Oliveira2019EmpathicStudy, BaghaeiRavari2021EffectsLearning, Charisi2021TheDynamics}. 
The mediator may enhance motivation and group satisfaction, by emphasizing each member's contributions, fostering a shared understanding of challenges, and offering positive examples of collaboration \cite{Chen2022DesigningInterviews}.

Many studies in this category utilize the robot to provide task information and support in an attempt to bias the human interactions, for example, to help the person with lowest contributions \cite{Short2017}, or to encourage actions that are likely succeeding to build trust between group members \cite{Stoican2022LearningCollaboration}.
Regarding mediation processes, the timing of an intervention has often been described to be triggered by knowledge of participation statistics and conversational flow, but also by certain levels of task progress.
Proxy factors that could be used as means for group process optimization can mostly be categorized as quality and quantity of interactions as well as the balance of contributions among the members.

In general, the main goal of these mediations is to enhance performance outcomes. However, for certain tasks, the focus may also be on fostering group cohesion through shared experiences \cite{Gillet2020AChildren}, building mutual trust for future work on the task \cite{Stoican2022LearningCollaboration}, or improving individual outcomes like the satisfaction of working together to solve tasks \cite{Chen2022DesigningInterviews, Lopes2021SocialTeams}.

\subsubsection{Moderate Decision Making}
This category considers situations that typically occur in established teams and are defined strongly around a task. The latter is often a decision-making task, where a group has to produce a single output out of a number of possible (and valid) alternatives. In these tasks, group members provide arguments, discuss them, sometimes come up with additional alternatives or compromises, and finally apply a decision method such as voting. Decision making is also often the final or follow-up activity in many other tasks, such as ideation \cite{deRooij2023Co-DesigningDynamics} or planning \cite{Buchem2023Scaling-UpNAO}. 
Such situations usually take place in a real or virtual static meeting environment and involve medium-sized groups of 3 to 10 people. 
The mediator role for this category is traditionally described as a meeting facilitator \cite{McFadzean2002DevelopingCompetencies}.
Many task-focused mediator interventions, such as structuring activities \cite{Buchem2023Scaling-UpNAO, Rosenberg-Kima2020Robot-supportedFacilitation, Shamekhi2019}, can be used to improve performance outcomes, but participation \cite{Hitron2022AIRobots, Tennent2019} and conflict management behaviors are also often applied. It can be helpful to summarize the current state of a discussion \cite{Alves-Oliveira2019EmpathicStudy}. Future concepts might profit from utilizing research on Group Decision Support Systems \cite{DeSanctis1987ASystems, Poole1988ConflictSystems} for task-oriented support functionalities.
Common target outcomes are performance, efficiency, or decision success. At the same time satisfaction might correlate with proxy factors like perceived contribution and participation balance. 

Possible real-world examples are business strategy meetings or hiring decision making, but also everyday situations like a group of friends trying to figure out which movie to watch.
In general, concepts in this category might be a good first candidate for real world applications due to their confined environment and rigid task structure. However, all current work has been evaluated in laboratory settings with artificial tasks and participant groups.

\subsubsection{Facilitate Ideation}
In contrast to moderating decision making, which works with or towards converging task dynamics, mediators can also support creativity and divergence for creating quantity and novelty in task outputs. 
This can take place in a meeting room setting but might also involve re-locations and distributed workspaces if the task involves sub-grouping or constructing prototypes.
For leisure tasks and conversations mediation might also encourage storytelling and open discussions \cite{Chen2022DesigningInterviews, Ohshima2017Neut:Conversations} to enhance enjoyment and insights.
The role of robot is to moderate ideation activities and to encourage idea generation and participation. 
Interventions thereforee often revolve around creating an open atmosphere and providing positive feedback. 
Free ideation is usually targeting an increased quantity of results (although only formulated as target in a single related concept \cite{Fucinato2023CharismaticCreativity}), while quality might only be a secondary target.

Example mediation situation can be brainstorming sessions in product development, a band writing a new song, or children and parents jointly creating a story\cite{Chen2022DesigningInterviews}.
We separated this mediation category from decision making as the desired dynamics of convergence versus divergence seemed to require very different interventions. However, in multiple of the experiment settings \cite{deRooij2023Co-DesigningDynamics, Ohshima2017Neut:Conversations, Tennent2019}, and potentially also in real work settings, an ideation phase might be followed by the need to select a subset of the created options, so developing robots for a combined mediation task could be valuable as well.    

\subsubsection{Manage Conflict}
This category covers concepts around business or community mediation or negotiation settings where two individuals or groups have competing targets, opinions, or resource requirements. 

Common task structures can be classified as mixed motive or competitive, but in general, conflict situations can appear across almost all task types. Importantly, in the way we define this category, the mediation goal is the resolution of a defined conflict.
If there is no existing rivalry (as proposed in the concept in \cite{Druckman2021WhoHumans}) the group can consist of people without explicit relationship and only exist for a given event, but it can also contain colleagues, peers or even friends in an established group when a task touches topics of diversity in opinions or goals. 
Intervention can thereforee help to work out compromises from a task perspective or provide means to establish common ground and understanding for each other’s needs and thoughts. Although in a first step it will also be necessary to call out destructive behaviors and handle misunderstandings \cite{Jung2015UsingViolations, Shen2018}. 
In particular for this category, it is important for success that the mediator establishes itself as a neutral and trustworthy agent \cite{Druckman2021WhoHumans}. As much as the existence and the quality of a solution matters, it is almost equally important for such concepts, that the mediator helps create outcomes on individual satisfaction and perceived fairness \cite{Cropanzano2001MoralJustice, Miranda1999MeetingInterventions}. Some application examples could be couple’s therapy \cite{Hoffman2015} or arbitration of a schoolyard brawl.

\subsubsection{Stimulate Group Mood}
Mediation concepts in this group are predominantly concerned with improving affective states or enjoyment of groups and their members independently \cite{Fu2023TheCollaboration} or as part of a task \cite{Lopes2021SocialTeams}. 
This fits well for some of the social tasks, such as conversation or networking, but a humorous atmosphere can also increase decision meeting satisfaction \cite{Pham2021LaughingEffectiveness}. Besides preferences for certain tasks, input factors can be very diverse. 
In contrast, the interventions are more clustered. This includes behaviors like telling jokes, motivational speeches and aiding in emotion support as well as proposing and defining new tasks or conversation topics, while task-focused activities are rarely important. It is also possible to use specific group effects, like emotion contagion\cite{Barsade2002TheBehavior}, through the mediator showing emotions itself or mirroring emotions of others. Examples can be the host of a party facilitating conversations, firing up a crowd at a concert before the main act, or a coach cheering up the team after a defeat.
Most concepts that we find matching this group do not integrate any complex knowledge about group processes or react to dynamic changes but rather define a target mood and activate a behavioral program that can trigger it. On the other hand, the work shows potential of integrating small mediation actions, such as asking "puzzle" question to engage group members \cite{Matsuyama2010PsychologicalGame}, into robots that are mostly build for a different application (such as a museum tour guide, or an entertainer).

\subsubsection{Establish Interaction Norms}
Looking at the group development model of Tuckman \cite{Tuckman1977StagesRevisited}, many teams will enter a phase of establishing norms for accepted or encouraged behaviors and interactions within the group context. But there are also general societal norms that should be considered for interactions in crowds and loose or short-term groupings. Mediation in this category is concerned with support for establishing and adhering to those norms to improve interaction quality and prevent conflicts and negative affect. 

In newly formed groups relevant situations can occur in many different tasks, but a task with high potential for such concepts could be co-existence. Strangers interact without being bound by a common interest, and norm violations are perceived less critical, as the group is not deemed to be existing beyond the current situation, or interactions are rather accidental or initiated only by one side. 
Mediation behaviors such as managing interpersonal processes or promoting positive interactions \cite{Dorrenbacher2023} and discouraging destructive behaviors can be useful in such situations. Maintaining order in a queue or encouraging people to free a seat for someone in need could be examples with co-existence tasks, preventing interrupting or disrupting behaviors during a meeting is one for groups with a common task.
A common difficulty encountered by many concepts in this group is the balance between impacting interactions and acting in a regulatory or judgmental role, which is not perceived positively by participants. 
This supports the perspective, also backed-up by looking at the defined target outcomes, that designs in this category are often driven by a problematic group process that should be avoided, while it is not clear, which benefit the group, or its members should have after a mediation.

\subsubsection{Enhance Understanding/Social Support}
Support groups \cite{Jacob21998Counseling} are meetings in which individuals with a common problem or challenge provide support to one another. This is typically done with the help of a mediator and provides a good example of concepts in this category. In general, most people react positively when they feel like they are being listened to while talking about topics of relevance for them (for example in doctor-patient relationships \cite{Wanzer2004ActiveListening}). For task groups, it is also beneficial to create cross-understanding \cite{Huber2017}, the knowledge of its members about each other’s mental representations, which requires sharing and communication. Supporting cross-cultural understanding might also help to better society at large \cite{Gomez2024DesignCommunication}.

Groups that have shared norms and expectations are more satisfied with group processes \cite{Park2008ThePerformance}. Taking other people’s perspectives to understand their views and actions can improve relationships and interactions \cite{Peterson2015Perspective, Shih2009Perspective}. For example, individuals engaging in active perspective taking are more easily understood by others \cite{Krauss2011Perspective}. 
This description shows that such mediation activities can be beneficial for all tasks, but might be particularly useful for sharing tasks, as well as for learning/teaching and bonding tasks. 
Developing teams and short-term groups with specific roles, such as doctor-patient \cite{Takano2009PsychologicalNeeds}, might need more support, but, as shown in the referenced work, it is also a powerful intervention for established and even close relationships \cite{Erel2021}. 
Common mediator intervention types are promoting mutual understanding \cite{Birmingham2020}, handling misunderstandings \cite{Nagao1994SocialAgents}, aid in transfer messages in a receiver-aware way \cite{Dietrich2022, Noguchi2023HowRecipients}. It might also be helpful if the mediator demonstrates understanding, for example through rephrasing and backchanneling  \cite{Kobuki2023RoboticStudy, Takano2009PsychologicalNeeds}.  

\subsubsection{Facilitate Learning}
Human teachers and trainers are not only tasked with teaching multiple individuals certain skills, but they usually also have to deal with and encourage certain social dynamics within the group of learners.
One of the most explicit concepts for their role as mediators might be the cooperative learning \cite{Slavin2010InstructionLearning} or peer-learning framework \cite{Topping2017EffectiveImplementation}, where the teacher acts as a mediator on groups of children to enhance the learning outcomes.

The most common task will be of the teaching/learning type, but there might also be other group tasks that can be designed to allow "on-the-task" learning \cite{Shimoda2022ApplicationRehabilitation}.  
Interventions we have found in the robotics literature are assigning of the roles of teacher and learner \cite{Chandra2016ChildrensRobot, Mitnik2008AnMediator} or moderating and structuring activities and foster mutual understanding \cite{Edwards2018ALearning}.
Mediation target outcomes will be individual (social) skill and knowledge improvement, as well as potentially team confidence.

\subsubsection{Team Building}
Team building is one of the most commonly applied group intervention in organizations. Previous work has shown that, in general, it can have a moderately positive effect across all types of team outcomes \cite{Klein2009DoesWork}. 

The defining component for a mediator is the work on member relationships and trust (“bonding”) as well as on interaction skills, role development and team identity. A typical approach to it is to provide a group with specific tasks which require specific types of interactions. 
Mediation is required when these interactions should be actively encouraged or taught \cite{Utami2017}, or challenges have to be adapted to the group state \cite{Chen2022DesigningInterviews}. Team identity can be strengthened through providing opportunities for shared experiences \cite{Jeong2018} and mutual understanding \cite{Fu2021UsingConversations, Strohkorb2016ImprovingRobot}.
We also introduced a new task category where the group members’ goal is focused on improving or building relationships with each other: Networking/Bonding. Human networks often have some members acting as social catalysts to initiate conversations and interactions between other members \cite{Saveski2021SocialOthers}. 
As can be seen from that, the intervention is mainly targeting established groups that are meant to act together for an extended period (teams). 
The concepts we have found that work on relationship skills only work on the specific group size of two. 
Those approaching larger groups focus on shared experiences and immediate positive interactions.
However, it is not clear how the past and future relevance of these groups is considered in these concepts, as they are usually tested with un-related participants and evaluated directly after an intervention.
In general, only a minority of publications covered long-term success of the interventions and we did not find any approaches targeting professional teams.
Main target outcomes will be long-term constructs, such as cohesion, mutual trust, or team confidence, and also in-group identification. 

\subsubsection{Form Group}
Before team building activities can be pursued to strengthen a group, the group context has to be established. 
For working teams, the members are often requested to join while the team context (for example its goal) is already well defined. 

Activities such as getting to know each other, establishing relationships and shared understanding of the task, and defining first roles are common in this early phase, which Tuckman named “forming” \cite{Tuckman1977StagesRevisited}. 
In other situations, for example communities of practice \cite{Tarmizi2005APractice} or conversational clusters at a party, the reason or opportunity for starting a group has to be established or highlighted for people to consider membership. Common input factors are groups at a very early stage with low relationship status between (potential) members. Other factors, such as task or group size can vary strongly. 
Typical target outcomes are group related aspects, like cohesion and interpersonal relationships, down to the basic measure of success of establishing a group at all. 
Mediation can play a big role for establishing a group, involving, for example, match making between people with common interests \cite{Ono1999} to help them form new relationships, or ice-breaking activities \cite{Gillet2022Ice-BreakersTeenagers, Zhang2023} to ease an initial contact. Stimulating and proposing first interactions through improving physical relations has also shown useful \cite{Ono1999, Sadka2022ByOpening-encounters, Takeuchi2014Whirlstools:Affordance}.
In existing work from the group formation category, verbal expressions stand out as the primary modality for mediating, whether it is through introducing conversation topics \cite{Shin2021, Zhang2023} or guiding through collaborative tasks \cite{Fukuda2016AnalysisMediator, Fan2021FieldAdults}. 
When comparing robot setups to non-robot configurations, such as tablets or posters conveying identical content, the embodiment of the robot demonstrated equal or positive effects \cite{Zhang2023}. 

\subsubsection{Promote Inclusion}
The diversity of team members is seen as a desirable state for building teams. An important aspect to make such teams work is the inclusion of all members in the group, that is, the satisfaction of their needs for belongingness and uniqueness \cite{Shore2010Inclusion}. In contrast, exclusion happens when group members are not treated as insiders with a contribution or value to the group \cite{Shore2010Inclusion}, which and has been shown to be detrimental to work attitudes and psychological health \cite{Hitlan2006Exclusion}. This concept is not only important for working teams but also applicable to larger social groups, and more and more explicitly targeted by societies. The primary agency to make inclusion work lies with the group itself \cite{Jansen2014Inclusion:Measurement}. 

Mediators of this category work to support inclusion and prevent behaviors potentially leading to exclusion in a group. Behaviors supporting this include participation management \cite{Barak1998Diversity}, framing of uniqueness of contributions and promoting mutual understanding with targets in in-group identification and sense of belonging of individuals, while paying attention on possible impact on group measures like cohesion. 
Relevant groups are usually already established and might be growing or being interested in growing, and some of their (new) members feature characteristics that make them unique or different with respect to many other members, for example due to age, capabilities \cite{Neto2023TheChildren, Pliasa2019CanDevelopment}, or personal background \cite{Gillet2020}, or simply because they lack established relationships to group members \cite{Sebo2020TheBehavior}.
All presented concepts approach acute inclusion in a concrete task interaction measured through improved participation balance and special member satisfaction, while we did not find work evaluating long term effects on the group (i.e. cohesion).\newline

The first three categories can be seen as mostly task-oriented mediation, while in the remaining ones the dominating outcome factors targeted by the mediator are more social.
\textit{Enhancing Understanding} can provide short-term effects during individual interactions, including satisfaction and enjoyment, but can also be approached with long-term targets of a mediation, which might show overlaps with \textit{Team Building}.
The remaining 4 aim to improve a group for future activities through interventions enhancing cohesion or interpersonal relationships.

In addition, one could consider a twelfth category, where the robot mediator is meant to assist a primary mediator in its task (as in falling into one of the above categories), including the need for coordination between the two (\textit{Mediation Assistance}).
Examples could be to improve the engagement of the audience with a moderator of a group activity \cite{Karatas2020UtilizationArt, Shimoda2022ApplicationRehabilitation}, or supporting teachers in helping multiple learning groups at the same time.
These categories are not meant to be mutually exclusive, and many mediation contexts might fit two or three of them, when for example explicitly integrating conflict resolution interventions within a decision making support context. 
However, we believe that from a robotics and AI perspective, it might be helpful to focus on the technical challenges in one category first, to reduce complexity while still providing a functionality for existing real-world use-cases. This can also be seen in most of the publications we have reviewed above.

\subsection{What are the specific affordances offered by robots that enable them to shape group processes?}
As Rahwan et al. \cite{Rahwan2020IntelligentCatalysts} discuss, there are several methods for enabling robots to mediate group interactions. These include (a) replacing a human mediator (e.g., \cite{Buchem2023Scaling-UpNAO, Chew2023WhoInteractions, Druckman2021WhoHumans, Takano2009PsychologicalNeeds}), (b) leveraging unique robotic features and abilities that surpass human capabilities (e.g., \cite{Gillet2020, Jeong2018, Sadka2022ByOpening-encounters, Wang2023ExploringConflict}), or (c) implementing interventions that are either exclusively or more readily accepted when executed by a robot rather than a human (e.g., \cite{Ikari2020Multiple-RobotDiscussion, Pliasa2019CanDevelopment, Shim2017AnEvaluation, Traeger2020}). 
Welge and Hassenzahl refer to (b) and (c) as "robotic superpowers" \cite{Welge2016}. 
It will be useful to evaluate which superpowers, i.e. unique robot affordances, could be useful for mediating group interactions and the extent to which current research has explored these.

One particularly interesting affordance may be the \textit{perceived neutrality} of the robot mediator.
Neutrality of the third party is of high importance in classical mediation and negotiation settings to enable willingness to compromise among the participants \cite{Astor2007MediatorPractice}.
The idea of "fair proxy communication" \cite{Seibt2018} describes the potential to increase perceived neutrality of the human mediator but has only been applied to tele-operated robots. This was used explicitly in the robot mediator concept from Druckman et al. \cite{Druckman2021WhoHumans}. 
In contrast, a robot may be the property of a certain group member and meant to mainly act in its interest, which might lead to enhanced trust in situations when it supports during a group activity with others \cite{Dietrich2022}.
It might also be particularly suited to act as a mediator in private spaces, where the group members would not tolerate a human mediator. Examples seen in the discussed work are working with children and their parents in their homes \cite{Chen2022DesigningInterviews}, private care-giving scenarios \cite{Moharana2019RobotsCaregivers}, or for communication between doctors and patients \cite{Takano2009PsychologicalNeeds}.

Another effect from a mediator being a robot might be applicable when working with participants with autism in a group to avoid social over-stimulation and complexity \cite{Cabibihan2013WhyAutism, Scassellati2012RobotsResearch}. Although most robotics approaches replace a human interaction partner for interactions and training of social skills, two of the concepts discussed in this paper evaluate the potential of using a robot to mediate between a child with autism and another human \cite{Karatas2020UtilizationArt, Scassellati2018ImprovingRobot}.
As \textit{robots do not take things personally or feel embarrassed}, they can also be used to address social norm violations without the fear of negative reactions \cite{Ablett2007BuildBot:Teams, Dorrenbacher2023, Jung2015UsingViolations, Traeger2020} or ask questions that are relevant for some participants without fearing being in the "spotlight" \cite{Ikari2020Multiple-RobotDiscussion}.

\textit{Robots can take roles that are implausible for human mediators} and even become part of the task setting, which provides them with a special angle for mediating group processes. One common role is that of a teaching target, providing participants (often children) a means for 'learning by teaching' \cite{Jamet2018LearningAbility}. While most social robotic research considers one-to-one interaction in those scenarios, there might be interesting opportunities in challenging groups of humans to teach the robot \cite{Cumbal2022ShapingBackchannels}. A teaching activity with a group can also be used as a team building task \cite{Kim2023ChildrobotBehaviors}. 
Conversely, the robot can teach social skills during an interaction and, because of its ubiquity, provide continuous feedback on application success through later group interactions \cite{Dorrenbacher2023, Utami2019CollaborativeRobot}.

Due to a large design space, \textit{robots come in various shapes and forms}, including those that blend in with human's everyday environments. This allows for more subtle mediation interventions that might even go unnoticed by the group while still being effective \cite{Erel2021, Kang2023TheGroup, Moharana2019RobotsCaregivers, Ono1999, Takeuchi2014Whirlstools:Affordance, Tennent2019}. Such background robots can also be used as an embodied indicator of group processes, for example by reacting to aggressive communication \cite{Hoffman2015} or imply participation imbalances \cite{Fu2017TurnTurn-Taking}.
Sometimes, such designs even allow for physical interventions, as done with the rotating stool robots in \cite{Sadka2022ByOpening-encounters, Sadka2023AllConflict, Takeuchi2014Whirlstools:Affordance}, which change the physical relations between people to enhance group formation processes.
Another unique affordance offered by robots is that they do not have to be restricted to a single location or embodiment \cite{Albers2022}. This can allow a mediator robot to work with groups distributed in space \cite{Jeong2018} or work with individual local groups while also facilitating interaction between them to enhance diversity of inputs \cite{Aylett2023}.
A specific design opportunity is the use of a robot for an enhanced tele-presence of a remote group member. Robots with own agency can consider mediation targets on top of the original idea of representing the person with the rest of the group. This could mean to enhance mutual understanding through more explicit body language \cite{Hasegawa2014FacilitationExaggeration} or to transparently adapt messages with the target to prevent misunderstanding and lower the burden to initiate communication \cite{Noguchi2023HowRecipients, Wang2023ExploringConflict}.

Finally, robots can be \textit{purpose-built to compensate for the potential limitations of human mediators}. They can assist by taking over specific sub-tasks, thereby reducing the overall workload \cite{Shimoda2022ApplicationRehabilitation}. Specifically designing such robots together with mediators as users could be an interesting way to further explore the potential of this approach.

\subsection{How do specific design characteristics and intervention strategies impact a robot’s effectiveness in shaping group processes?}
We aimed to extract design aspects along the different mediation factors of the Mediation I-P-O model that would show generalizable effects on group processes or group outcomes. 
The relatively low number of papers reporting significant effects of their robot mediator and the very small overlap in measures used to assess these effects makes it difficult to extract generalizable insights about design choices that should be pursued to influence specific group processes.
However, looking at work that evaluated changes in different proxy factors in response to robot interventions, we can at least provide some successful examples that may warrant further investigations.

Shen et al. \cite{Shen2018} found that the robot mediator supporting identification of a detected object possession conflict between two children, managing the conflict and promoting mediation actions lead to a significantly higher number of constructively resolved conflicts (coded with a scheme based on \cite{Chen2001PeerPatterns}) compared to the baseline robot that did not intervene. The authors validated this effect through additionally evaluating if children paused their activities after a robot intervention, which correlated with the resolution success.
Shameki et al. \cite{Shamekhi2018} compared decision making facilitation with and without an embodiment (loudspeaker versus avatar face on a telepresence robot). Although they did not find significant effects on consensus, decision time, or choice satisfaction, they could show that having an embodied mediator improved participation balance. Participants' evaluation of the mediator with respect to social features and rapport (using established scales) showed higher ratings for the embodiment strengthening this conclusion.
Human therapists mediating social conflicts try to work with interventions that create spatial setups where group members are facing each other and can share and maintain eye contact \cite{Mohacsy1995NonverbalSession}. Robotic chairs, enabling interactions in such setups through physical rotation, have been shown to be able to positively affect perceived conflict intensity and the number of intimate interactions between couples discussing a real topic of disagreement \cite{Sadka2023AllConflict}. Although, the strict laboratory experimental setting cannot provide information about many aspects of the general role effectiveness for such a mediation robot, such physical interventions might be worth investigating more. However, In a similar experiment with strangers the same behavior could not create significant effects with respect to relational affect and general interactions \cite{Sadka2022ByOpening-encounters}. 
A bystander mediation robot showing behaviors of attentively following a conversation (active presence intervention) during a dyadic emotional support interaction could be shown to impact interpersonal synchronization, non-verbal immediacy, and verbal empathy behaviors and lead to an improved perceived interaction quality \cite{Erel2021}. They used an abstract robotic object but experiments with a humanoid bystander performing similar behaviors during a doctor-patient conversation could also trigger some changes in the perception of the group processes by the patients \cite{Takano2009PsychologicalNeeds}.
The robot mediator in \cite{Scassellati2018ImprovingRobot} showed that establishing a shared attention event with a human group member allowed it to positively manage interpersonal gaze interactions between group members, albeit for the special case of children with autism and their caretakers.

Of course, the applicability of these findings across tasks, environmental settings, robot platforms, etc. remains limited until they undergo rigorous validation through additional research.

\section{Discussion}
The insights from our research questions should inform future research on using Social Mediation Robots to target changes in group processes through directed and intentional action. The proposed Mediation I-P-O model seeks to bridge some of the gaps in existing research methodologies by providing a comprehensive framework that includes all essential components that must be considered for research on Social Mediation Robots. It identifies the core elements within the input, process, and output components, emphasizes the need to clearly define the group processes being targeted by mediation, and separates the mediation processes from group processes to enable researchers to mindfully plan and consider all the key design factors and target outcomes for their study. In doing so, it also promotes positive research practices such as complete and comprehensive reporting to tackle challenges related to reproducibility in HRI research~\cite{Bagchi2023TowardsReporting, Leichtmann2022CrisisImprovement, Schrum2020FourStudies}. Additionally, appendix~\ref{Appendix:Model} provides further details on the model and is a rich source of valuable information for the HRI community. It includes comprehensive lists that identify and describe all key elements in the model and can serve as a taxonomic guide that helps in aligning understanding and facilitating research that is consistent and comparable.

By making a clear distinction between group processes and outcomes, the model also aims to encourage researchers to select appropriate metrics for evaluating the effectiveness of their interventions. Measures should be used to evaluate the target outcomes of the mediation. However, as we already discussed, during our review we found a number of measurement constructs that were, in fact, evaluating group processes instead of the outcomes without making their relation to the outcomes transparent. One example of such proxy factors popular in existing studies quantifies participation statistics. A balance in participation from all group members may be correlated with a target outcome, such as performance, but it must be noted that there may be various other aspects of the interaction that also contribute to this outcome~\cite{Woolley2010EvidenceGroups}. Another example is the target of increasing the quantity of certain "desired" or general interactions, with an, often implicit, assumption that this results in a "good" outcome.
Additionally, studies often fail to clearly identify and report the target outcome of mediation. This may be attributed to the lack of common understanding and agreed-upon definitions within the field (see, for example, \cite{Abrams2020}) and explain, at least in part, the lack of consensus on common measures for a given outcome. 
Additionally, identifying commonalities in concept targets is hindered by the frequent use of custom questionnaires and measures, which often lack validation and may not represent the full scope of certain outcome constructs. Multiple studies also utilize users' perception of or closeness to the robot itself as a measure of success for of their mediation rather than considering the effect of this on mediation success (e.g. \cite{Ablett2007BuildBot:Teams, Fu2021UsingConversations, Kobuki2023RoboticStudy, Seaborn2023VoiceConversation}). We also see a high number of performance-related mediation targets, however, it has been emphasized that the satisfaction of group members and the overall group should be prioritized as key outcome, as it is influencing long-term group interactions beyond mere performance metrics \cite{Fine2017GroupFun, Keyton1991EvaluatingVariable, Olaniran1996AMeetings}.
This emphasizes the need for establishing a common understanding of core constructs through interdisciplinary collaborations.
 
However, there exist challenges in research on Social Mediator Robots that extends beyond what the Mediation I-P-O model can address. Among these is the limited real-world applicability of findings from current research utilizing these robots. A majority of the studies are conducted in controlled laboratory environments, a trend found in the larger HRI field as well \cite{Jung2018RobotsWild}. This implies that lab environments, abstract tasks, and unrelated, student participants in user studies may not fully represent the needs of real-world scenarios where mediation is required, and only inadequately address the complexities of these situations. Additionally, another factor that makes it challenging to assess the validity of the reported findings is that control conditions that could show the benefit of a certain mediation concept are commonly omitted from the research methodology. While a small number of studies include control groups without mediation \cite{Edwards2018ALearning, Rosenberg-Kima2019Human-Robot-CollaborationGroups, Shamekhi2018, Shamekhi2019, Zhang2023} (and to some degree \cite{Ablett2007BuildBot:Teams, Druckman2021WhoHumans}), it is not a widespread practice in studies targeting social mediation, leading to potential biases and limiting the robustness of the conclusions drawn.

Related to this is the need to perform longer-term investigations producing insights beyond the novelty effect, which have the potential to benefit the field tremendously \cite{Gallego2013Novelty}. Not only is there a need to evaluate the long-term impact of mediation, but studies should also be thoughtfully designed with respect to the time scales in which mediation takes place. For example, the role of a robot mediator must be defined in the context of whether it is intended for short-term interventions, such as team-building exercises with new groups, or for continuous support of a single group over time. In the latter case, robotic mediation might need to adapt to groups in different developmental phases, which likely require different interventions. Understanding these dynamics is crucial for developing effective and sustainable mediation strategies.

Related to this is another trend in the field where many reviewed studies do not explore the effects of mediation concepts beyond a single experiment. We did find several mediation concepts that underwent multiple rounds of development and yielded follow-up publications (for example, \cite{Pettinati2015TowardsRelationships}$\rightarrow$\cite{Shim2015AnRelationships}$\rightarrow$\cite{Shim2017AnEvaluation}, or \cite{Kim2018DesigningOpportunities}$\rightarrow$\cite{Kim2023JointInteraction}) or research groups that reused successful concepts from previous studies (for example, \cite{Zuckerman2015EmpathyCompanions}$\rightarrow$\cite{Hitron2022AIRobots}\&\cite{Hoffman2015}\&\cite{Erel2021}). However, most concepts appeared to be limited to a single experiment without subsequent investigations. While it is possible that follow-up work was conducted but not yet published, this trend may be driven by tendencies in the field and publication venues to value new experimental designs over longitudinal studies yielding generalizable findings. 

Of course, it is not difficult to understand the challenges associated with long-term HRI studies \cite{Leite2013SocialSurvey} -- they pose significant logistical difficulties, require intensive human resources and time, and yield large sets of data that can be difficult to analyze. However, long-term studies investigating the impact of social mediation through robots come with their own unique research questions. Mediation efforts must take into consideration the relationship history of the participants as they interact with each other repeatedly and develop rapport. They must evolve as participants collect more and more experience together, build an understanding of each other, and become familiar with each other's needs and preferences. This requires advanced algorithms that capture and utilize this nuanced understanding to equip Social Mediation Robots with the social intelligence needed to leverage these evolving relationships to provide effective mediation. Personalization of this nature must be able to handle highly subjective human data. Increasingly personalized mediation also requires that algorithms be trained to handle diverse cultural and social contexts in order to mitigate the risk perpetuating and amplifying human bias. These risks are similar to, for example, the risk of perpetuating biases in increasingly personalized algorithms utilized by robots providing care for people with dementia (see \cite{Kubota2021SomebodyCare}).

The ethical and social implications of robotic mediators have been explored in several studies, primarily through theoretical and design perspectives. For instance, the design fiction approach employed by \cite{Dorrenbacher2023} revealed participants’ concerns about robotic mediators, including provocation, feelings of exclusion, exposure, social de-skilling, and the potential for negative behavior manipulation. These findings underscore the complexity of integrating robotic mediators into social settings, as they may inadvertently exacerbate existing social issues or create new ones. Dietrich and Weisswange \cite{Dietrich2022} examined the impact on privacy and the perceived appropriateness of robots sharing personal information in the presence of or towards others. Their study highlighted the delicate balance between the benefits of robotic mediation and the potential risks to individual privacy (also see \cite{Dietrich2023Privacy}. Similarly, Jeong et al. \cite{Jeong2018} explicitly evaluated participants’ privacy perceptions regarding the sharing of living noises within a group and emphasized the need for careful consideration of privacy concerns when designing and implementing robotic mediators. Given that many mediator concepts assume access to personal information or consider groups of children as potential participants, it is crucial for future research to more carefully and explicitly consider these implications. For inspiration, researchers can look to studies such as \cite{Charisi2022ChildRights}, which provide valuable insights into the ethical considerations of involving children in robotic mediation. Furthermore, the social relationship between the robot and group members during interventions, particularly concerning bonding or deception, has also been briefly addressed in some studies \cite{Moharana2019RobotsCaregivers, Alves-Oliveira2019EmpathicStudy, Tennent2019}. These studies refer to prior work on ethics in social human-robot interaction \cite{Kwon2016Deception, Riek2014Roboethics, Serholt2017Ethics, Sharkey2016Ethics}, highlighting the importance of ethical guidelines and frameworks in the development and deployment of robotic mediators. The potential for robots to influence social dynamics and relationships necessitates a thorough examination of their ethical and social implications to ensure their responsible and beneficial integration into society.

A significant limitation in the field of robotic mediation is the perception and understanding of situational contexts and ongoing group processes. Many approaches rely on human perception through wizard-of-Oz control or simplified perception due to controlled environments, such as laboratory settings or meeting rooms, and well-defined tasks. This highlights the need for stronger interdisciplinary collaboration with researchers from fields like computer vision, machine learning, affective computing \cite{Picard1997AffectiveComputing}, and social signal processing \cite{Pentland2007SocialProcessing}. Additionally, the semantic understanding of conversational input can benefit from integrating recent advances in large language models \cite{Williams2024LLM}. 
Although current work depends heavily on pre-determined input factors through the use of static interaction and mediation scenarios, it would be interesting to see how future research addresses enables robot capabilities of autonomously detecting and inferring input factors for effective mediation. 
Additionally, to leverage machine learning modeling capabilities, there is a need for datasets that capture relevant dynamics. Beyan et al. \cite{Beyan2022Face-to-FaceSurvey} highlighted limitations in existing group activity datasets, such as the lack of real-world scenarios and repeated interactions with the same groups. Research on group affective states has its own dedicated community with numerous datasets available (see, for example, \cite{Veltmeijer2023AutomaticReview}), although these often focus on static classification of states, with limited research on capturing ongoing state changes during group interactions (but see \cite{Sack2023CORAE:Interactions} for a tool to capture continuous group affect dynamics). Addressing these gaps will be crucial for advancing the field and developing socially-aware robotic mediators.

Another interesting challenge is assessing the necessity of embodiment in mediation applications and understanding the need for specific embodiment features in comparison with a smart speaker or computer screen, especially when the primary intervention modality is speech (as done in \cite{Druckman2021WhoHumans, Shamekhi2018, Zhang2023}). Even when the need for embodiment is established, determining whether and to what extent effects transfer across different robotic platforms may be difficult. It is also unclear which specific features and behaviors in a given form factor, such as a humanoid, are necessary to achieve a desired effect. This ambiguity complicates the ability to draw concrete generalizations about the design characteristics and mediation strategies that impact a robot’s effectiveness in shaping group processes.

When evaluating the appropriateness of robotic mediation concepts and interventions, participant feedback is crucial. For instance, feedback on the caregiver-caretaker mediation concept presented in \cite{Shim2017AnEvaluation} suggested that neutrality should be part of the functional design, and the robot should not "judge" or "command" humans. Similarly, while the robotic co-reading mediator in \cite{Chen2022DesigningInterviews} received positive feedback from many users, some parents reported increased effort and multiple undesired side effects. The authors recommend that “desired robot roles should be explored from each stakeholder’s and group’s perspectives” \cite[pp 407]{Chen2022DesigningInterviews}, a practice not often carefully executed in other reviewed papers. Another open question not addressed by current mediator concepts is the intended ownership and related "intent" of the robot. Meaningful target settings might differ significantly depending on whether the robot is owned by an individual group member, the entire group, or the group’s superior. Addressing these questions is essential for developing effective and ethically sound robotic mediation strategies.

It is interesting to note that the mediator concepts reviewed in this work often extend beyond the mere design of the robot or its mediation behavior. Multiple authors approach target outcomes by additionally or predominantly designing specialized tasks, controlling the environment, or utilizing the robot as a support for mediation, for example games and training schedule to improve social skills of children with autism \cite{Pliasa2019CanDevelopment, Scassellati2018ImprovingRobot}. While this mirrors the practices of human mediators, it can be argued that it limits the robot to an assistant or tool for human designers.

Some publications have addressed the potential side effects of social robots on group processes and outcomes. For instance, Shen et al. \cite{Shen2017RobotConformity} explored the subconscious effects on conformity in decision-making, revealing how robotic mediators might inadvertently influence group dynamics. Hitron et al. \cite{Hitron2022AIRobots} explicitly evaluated the effects of human biases transferred to AI agents when trained on real-world data \cite{Howard2018AIBias}, highlighting the ethical implications of such biases in robotic mediation. Other studies have shown that a robot mediator can make a group aware of conflicts that might otherwise go unnoticed \cite{Jung2015UsingViolations}, or conversely, cause exclusion through sub-optimal decision-making \cite{Neto2023TheChildren}. These findings underscore the importance of considering unintended consequences when designing robotic mediators. Taking such effects into account is crucial even for robots not targeting mediation tasks (for details, see \cite{Erel2024RoSI:Influence, Gillet2024Interaction-ShapingAgents}), emphasizing the need for a comprehensive understanding of the broader impacts of robotic interventions. By addressing these potential side effects and integrating feedback from diverse stakeholders, researchers can develop more effective and ethically sound robotic mediation strategies. This holistic approach ensures that robot mediators not only achieve their intended outcomes but also mitigate any unintended negative consequences, ultimately enhancing their acceptance and effectiveness in real-world applications.

While social robots will likely remain limited in their interaction capabilities compared to human mediators in the near future, this limitation presents a unique opportunity for researchers. By focusing on specific interventions, researchers can evaluate the explicit effects of robotic mediation on group processes and outcomes, a task that has traditionally been challenging in group research due to numerous confounding factors introduced by human mediators. However, to ensure general applicability, such experiments must be meticulously designed to extend beyond single, abstract tasks and context-free group settings. Moreover, to explore the potential of robot mediators in future applications, it is essential to consider their roles holistically, as outlined in this work. A narrow focus on only one or a few aspects highlighted in the Mediation I-P-O model may yield an incomplete design and risk producing results that may not transfer to different settings. By adopting a comprehensive approach, researchers can develop more robust and transferable insights, ultimately advancing the field of robotic mediation and its practical applications.

\section{Conclusion}
We presented a scoping review of concepts about artificial agents mediating human group processes.
For our analysis, we extended, structured, and mapped existing models and taxonomies from group research to the field of mediation, human-robot interaction and social robotics. 
The extensive search process identified more than 100 papers with 89 unique concepts for robotic group mediation.

Analyzing these concepts showed a large diversity in targeted group factors, outcomes and intervention approaches. 
Due to a common lack of references to existing models and unclear taxonomies, many concepts and their results are difficult to relate to each other, and transferring insights of the corresponding experiments and approaches to other social robot mediation work has proven difficult.
We have condensed our insights into 11 categories of group mediation roles that allowed us to discuss possible application targets. 
We also identified specific features that robots can contribute to mediation settings as compared to humans due to, for example, special embodiment designs. This might be an interesting direction to further explore unique roles for robotic mediators.
Discussing positive effects of robot mediator concepts and quantifying potentially successful improvements to groups proved difficult and we propose that target outcome measures driving a mediation concept need to be considered more intensively in the future.

We hope that the present work, can provide insights for a common framework that future research can use during both design of new approaches and the description and reporting of robotic mediation work, and to identify gaps in our general understanding of mediator--group interactions.

\begin{acks}
The authors would like to thank Tuan Vu Pham, Maria Teresa Parreira and Michael Sack for contributions to the initial discussions about organizing the space of robotic group mediation and Randy Gomez for valuable input about applications and open issues. 
This research was funded by the National Science Foundation through the CAREER Award (\#1942085) for MFJ. Any opinions, findings, conclusions, or recommendations in this material are those of the authors and do not reflect the views of the foundation.
\end{acks}

\bibliographystyle{ACM-Reference-Format}
\bibliography{main.bib}

\appendix

\section{Defining Social Robot Mediation}

Existing terminologies may broadly encompass social mediation but are unable to capture the nuances of the function discussed in this review. 

\begin{figure}
     \centering
     \includegraphics{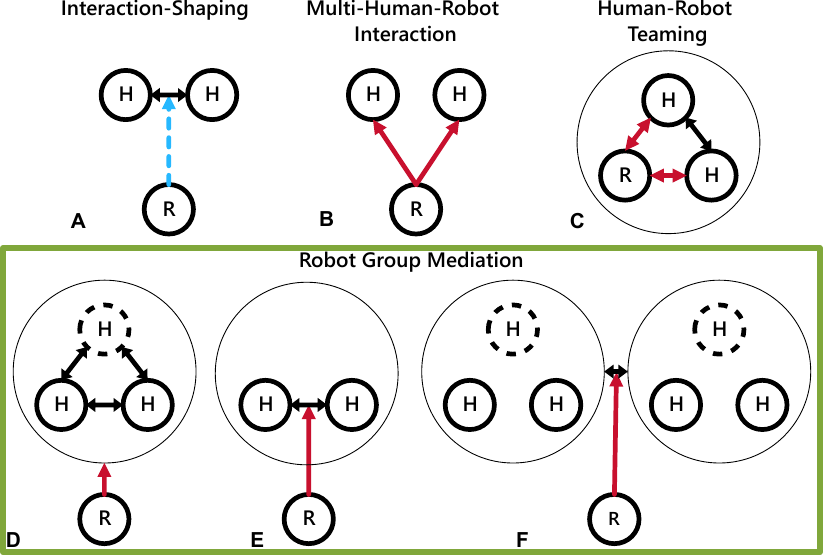}
     \caption{Interaction graphs for different robot group patterns. Arrows show interaction directions between robots (R) and multiple humans (H). Dashed arrows show implicit influence, solid arrows show explicit/intentional interaction. A-C: Representation of dominant interaction pattern found in related work. (A) Human-human interaction influenced by robot behavior that is not necessarily directed towards the humans. (B) Robot interacting with multiple humans independently. (C) Robot interacting reciprocally with humans as part of a joint group. D-E: Interaction patterns in focus for social robot mediation. (D) The robot takes action on the group as a whole. (E) The robot explicitly influences the interaction pattern between two (or more) humans. (F) The robot mediates the interaction between two (sub-)groups of humans.}
     \Description{Figure 10. Schematic depiction of the three interaction graphs used in related work and the three patterns relevant for this work. The interaction-shaping approach can be depicted by an indirect connection between the robot and the interaction between two humans. Multi-human-robot interaction uses separate connections between the robot and two individual humans, while robot teaming has the robot as part of a group with mutual interaction between all group members. The three graphs covering robot mediation as discussed in this work show the interaction of the robot with the group as a whole, direct action on the interaction of humans in a group and influencing the interaction between two groups.}
     \label{fig:Graphs}
\end{figure}

Multiple prior publications have tried to analyze interactions between multiple humans and/or multiple robots. 
Sebo et al. were the first to provide a review of HRI literature~\cite{Sebo2020} and addressed the scalability of lessons learnt from one-to-one human-robot interactions to one-to-many, many-to-one, and/or many-to-many configurations. 
Schneiders et al. also identified trends in HRI literature~\cite{Schneiders2022Review}, such as an increasing focus on simultaneous interactions beyond the one-to-one configuration and the popularity of one-to-many configuration in interactions between humans and digital artefacts, including robots. 
Recent work from Dahiya et al.~\cite{Dahiya2023} characterized literature in terms of team structure, interaction style, and robot control methods, and used interaction graphs~\cite{Yanco2004ClassifyingTaxonomy} to highlight direct and indirect interactions between group entities. 
The definition of \textit{Socially Assistive Robots}~\cite{Mataric2016SociallyRobotics} may broadly encompass a social mediation function but is traditionally only utilized to provide assistance in care-giving scenarios to individuals in need, such as older adults with dementia~\cite{Abdi2018ScopingCare} or children with autism~\cite{Syriopoulou-Delli2021RoboticsEducation, Cabibihan2013WhyAutism}. 
Chita-Tegmark et al.~\cite{Chita-Tegmark2020ElderlyMediationReview} reviewed research on such robots assisting social interactions, specifically within groups of older adults and individuals with health conditions. 
More recently, Gillet et al. introduced the formulation of interaction-shaping robots~\cite{Gillet2024Interaction-ShapingAgents}.
By including social influences implicit in robots and their behaviors, both \textit{Interaction-shaping robotics}~\cite{Gillet2024Interaction-ShapingAgents} and also the \textit{Robot Social Influence model}~\cite{Erel2024RoSI:Influence} do not necessitate behavior to be specifically designed to actively target changes in human social behavior. Instead, they allow for the incidental emergence of the robot's social influence without direct intentionality. 
Additionally, \textit{Interaction-shaping robotics} does not require the robot to play a specialized role, thereby including scenarios where the robot may serve as a peripheral companion or an equal peer to the humans. 
Described as robots that assist through social rather than physical interaction, this paper adopts the term \textit{Social Robot Mediation}\includecomment{for group interactions} to more specifically reflect the active and intentional facilitation of group social interactions. 
Where possible, we seek to integrate models and formulations from prior work to further push for a common taxonomy in the field. Extending the interaction graphs from Dahiya et al.~\cite{Dahiya2023}, we highlight our emphasis on robots mediating social interactions at the group level, rather than interacting with individual members in Fig. \ref{fig:Graphs}D-F.

\section{Social Group Mediation Model - Details}
\label{Appendix:Model}

This section contains a short description of every aspect within the different factors of the Social Group Mediation Model which we have used to compute statistics on the reviewed concepts.
We do not claim completeness of these sets, however, we believe that it will be a good starting reference when trying to unify reporting practices in future studies.

\subsection{Input Factors}
There exist many different input factors that likely influence group process \cite{Forsyth2018GroupDynamics}.
For our analysis, we selected the ones that we have encountered or deemed directly relevant for work in robot group mediation. We believe that future work should consider defining those within their group mediation concepts, although many others might become relevant in specific applications.

\subsubsection{Individual input factors}
\begin{itemize}
    \item Abilities/Skills/Knowledge - task- and social interaction-related capabilities that each group member has before the group activity. This can also include physiological or cognitive limitations as well as relative skill levels with respect to the majority of the members.
    \item Age - for this work we consider five categories young children (age 3-6 years), children (age 6-12 years), teenagers (age 13-18 years), adults and older adults (age 64+). Groups can also contain a mix of age groups ("inter-generational").
    \item Conditions - considers medium- and longer-term health conditions which imply a variety of needs and restrictions and might imply special interaction concepts for both groups and mediators. Examples that we find in the reviewed work are Parkinson's disease, dementia, aphasia, autism, and visual impairments.
    \item Mood/Motivation - general affective states that group members are in when starting an activity, including being happy or stressed and if they have positive feelings towards the group and the upcoming activity.
\end{itemize}
\subsubsection{Group input factors}
\begin{itemize}
    \item Group Size - there has always been discussions about the minimal size of a group \cite{Moreland2010AreGroups, Williams2010DyadsAre}. We include dyads as the smallest unit that allows a mediator to work on interpersonal processes. However, we acknowledge that certain processes will only be found with larger groups, and that results found with group sizes of two might not transfer to larger groups. As for many concepts it might not be meaningful to provide exact numbers for the intended group sizes, we use the following clusters: 
    \begin{itemize}
        \item dyad - exactly 2 members
        \item triad - exactly 3 members
        \item small group - scaling between 2 and 5 members
        \item bigger group - between 5 and 10 members
        \item large groups - undefined number, larger than 7
    \end{itemize}
    \item Relationships - the social connections between the group members that exist prior to the group activity. We differentiate between (similar to e.g. \cite{Lv2018Multi-streamVideos}:
    \begin{itemize}
        \item team - individuals that share predominantly professional relationships \cite{ Forsyth2018GroupDynamics} bound by shared role or task. 
        \item peers - individuals belonging to the same societal group (e.g. students) or with the same affiliation (e.g. sports club, school)
        \item authority - relationship defined through hierarchical structures (e.g., superior - staff or teacher - student) or dependencies (e.g. caregiver - caretaker).
        \item family - long-term socio-affective bonds mostly determined through common ancestry \cite{Strodtbeck1954TheGroup}, often multi-generational.
        \item couples - (mostly) dyads with affective relationship and co-living conditions.
        \item friends - members with long-term interdependence based on shared interests and experiences \cite{OConnell2008GroupApproach}
        \item strangers - co-located people without a shared social identity or goal and no prior interaction history or relationship as it can be found in crowds \cite{Adrian2019Crowd}.
        \item rivals - people with opposing or competing (long-term) interests or affiliation to entities with such interests, as in classical community mediation settings.
        \item mixed - combinations of the above, for example for a group of two unrelated families.
    \end{itemize}
    \item Group Composition/Roles - composition describes specific ratios between or the presence of individual factors of group member. Example compositions that we find in the discussed concepts are a combination of children with adults or older adults, a combination of experts and novice members, or compositions with a specific minority that might lead to imbalances in group membership.
    Additionally, different roles and responsibilities within a group can be established before starting an activity, for example based on the processes of previous group interactions or explicit assignments from a superior. We can find predetermined roles of group leader, caregiver and caretaker, or performer and audience.
    \item Group History/Future - refers to the overall lifetime of a group which determines, for example, if outcomes of an activity are relevant for further interactions or if the group might have gone through some of the group development stages \cite{Tuckman1977StagesRevisited} already. We classify this based on if a group was established prior to the current activity or is newly forming or adding members during the activity. Groups that only exist for an activity and neither play a role before nor after will be framed as "random".
\end{itemize}

\subsubsection{Environment input factors}\label{Appendix:TaskList}
One major influence factor that is determined through the environment of the group is the task at hand. Prior work has proposed a taxonomy for tasks typically approached by teams \cite{McGrath1984Groups:Performance}. We use these as the first 8 categories, while the remaining ones are proposed by us to also cover social group settings.
\begin{itemize}
    \item Planning – generating plans, with a primary target on high quality outcome with respect to details, feasibility, resource requirements, etc.
    \item Creativity – generating ideas, targeting high quantity and novelty of proposals.
    \item Intellective – solving problems for which a single correct answer exists, where, besides success, solution time has high relevance.
    \item Decision making – deciding on issues with no objectively correct answer, where agreement and satisfaction could be used to evaluate success.
    \item Cognitive conflict – resolving conflict of viewpoint or opinion, which often requires mutual understanding to be reached.
    \item Mixed motive – resolving conflict of interest or targets as in negotiations, where the quality of the compromise usually has an external and multiple internal aspects.
    \item Contests/competitive – resolving conflicts of power, opposing targets, where only one participant (or sub-group) can win.
    \item Performances/psycho-motor – executing tasks, striving for excellence, physical collaboration.
    \item Conversation/Leisure/Free Play – improving mood/well-being/enjoyment through social interaction
    \item Learning/Teaching – improving individual knowledge/skill through sharing knowledge.
    \item Information Exchange – synchronizing/updating mental models through disclosing “private” information to others .
    \item Networking/Bonding – improving relationships and cohesion through interaction and shared experiences.
    \item Care-taking – enabling someone to perform everyday individual task despite limitations.
    \item Co-existing - individual tasks and goals which however imply interaction with others through sharing space or resources. A social goal in such setups is to limit negative impact to other and not violate social norms \cite{Rother2023Coexistence} .
\end{itemize} 

other considered environmental input aspects are:
\begin{itemize}
    \item Setting - describes the location of the group activity, which implies certain spatial relationships of the team members and constraints the dynamics of the (inter-)actions. We used the following categories:
    \begin{itemize}
        \item Virtual/Remote Space - Group members are (at least partially) located in separate places, which prevents direct physical interaction and requires technical means for communication. One such technical means can be the full virtualization of the meeting space, where users (and mediators) interact through digital avatars.
        \item Static Seating/Standing Arrangement - The classical office setting, where people sit or stand around a table screen. An activity will usually be performed within the same room and address conceptual tasks.
        \item Dedicated Group-Space - The group activity is approached in a room that is specific for a task or more general target of the group. Examples could be a production hall or a gymnasium. In contrast to the previous setting, tasks in this one are usually of behavioral type.
        \item Casual Interaction Space - Confined locations in which groups can come together and interact without a specific task implication. This will frequently be found for social tasks and for groups without business context. Examples can be the home or a night club.
        \item Public Space - Casual groups can be forming without an organized intent through public encounters and crowds can usually interact in any place where many people come together. Such settings might be a bus stop, a museum or on the street.
    \end{itemize}
    Related aspects, like the explicit seating arrangement or a place's atmosphere or appeal have not been explicitly considered in this paper.
    \item Temporal conditions - Independent of the task, there might be time pressure on the group, for example through a deadline set by a superior, which can affect group performance \cite{Durham2000EffectsPerformance}. Other temporal conditions are the re-occurrence of a task and the frequency of group activities, which might promote habitual group processes \cite{Gersick1990HabitualGroups}.
    \item Embedding - describes general aspects of the environment in which a group exists, and which might influence how tasks might be approached, or which social processes are accepted. Aspects that influence this category can be cultural traditions and societal norms, company culture, or legal and regulatory boundaries.
    \item Resources - limitations of material and tools required for a given task will influence a group's approach but also induce additional requirements (e.g. coordinated use) and potential conflicts. Missing or constraint resources have been discussed to, for example, influence creative processes \cite{Keupp2013ResourceSector}.
\end{itemize}

\subsection{Group Processes}
There exist behavioral coding schemes to assess group processes for short observational units. Popular examples are the Interaction Process Analysis \cite{Bales1950InteractionGroups.} (which has also been adapted to e.g. analyze role dynamics \cite{Pianesi2007RoleAnnotation}), the Hill Interaction Matrix \cite{Hill1971InteractionMatrix}, 
or the Interaction Dynamics Notation scheme \cite{Miller2021IDNReview, Sonalkar2013InteractionDynamics} (which has also been used in research on ideation mediation \cite{Maier2020IDNMediation, Pham2024EmbodiedConsensus-Building}). 
Fuhriman and Packard \cite{Fuhriman1986GroupIssues} provide an overview about a large number of such schemes. 
There is also work on trying to automatically recognize some group processes using speech or vision-based models \cite{Gatica-Perez2009AutomaticReview, Javed2023GroupMediation}, which often also use description spaces from proxemics and kinesics.
However, it is challenging to find established taxonomies for group processes for larger observational units covering a diversity of tasks, which has been mentioned and criticized before \cite{Marks2001AProcesses}.
We consider group processes to describe types of dynamics within the states of any aspect involved in the current activity, similar to \cite{Ilgen2005TeamsModels, Marks2001AProcesses} but in contrast to the more event-based perspective used by the behavioral coding schemes or to emergent states that describe a certain parameter or measure of group activity at a given time.
Each of such processes might come in multiple different qualities, for example verbal conflict might be constructive or involve personal attacks, and a mediator might want to influence the process towards a certain direction.
We do not intend to provide a comprehensive list of group processes in this paper though, as we have seen few of the reviewed papers explicitly discussing group processes beyond their possible use as an intervention trigger (see appendix \ref{Appendix:InterventionList}).
For completeness, we still want to share the current state of such a list, which we tried to extract based on a large amount of prior work \cite{Aldag1993BeyondProcesses, Amici2015CoordinationPerspectives, Becker2016TeamarbeitTeamentwicklung, Benne1948FunctionalMembers, Biber2021TowardsAnalysis, Cini2001GroupInnovation, Farley2018TheSkills, Forsyth2018GroupDynamics, Kauffeld2018TheScheme, Ilgen2005TeamsModels, Maier2020IDNMediation, McGrath1984Groups:Performance, Marks2001AProcesses, Mathieu2008TeamFuture, Miller2021IDNReview, Scheerhorn1994, Zaki2013InterpersonalRegulation} and structured along the four types of \textit{task-level}, \textit{individual-level}, \textit{interpersonal-level}, and \textit{group-level processes}. Some processes that have been discussed for groups in prior work will instead be considered in the mediation processes (see appendix \ref{Appendix:InterventionList}) as they tend to work on top of another group process.

\subsubsection{Task-level processes} revolve around the group's aim to advance in the current task:
\begin{itemize}
        \item Task assignment/Planning - defining individual and sub-group work packages and assigning them according to skills and availability of the members.
        \item Information exchange - group members communicate relevant facts about the task towards the group.
        \item Elaboration - provide additional details and background to a task-relevant information item.
        \item Problem identification - analyzing the task and finding possible strategies.
        \item Resource identification - collecting which items, tools and member skills are available that could potentially be used during the activity.
        \item Joint action - group members carry out an action that requires contributions from multiple people.
        \item Coordinated task execution - group members carry out interdependent actions that progress the task.
        \item Individual task execution - individual group members work on their own on sub-tasks that have no interdependence with other members.
        \item Creating ideas - individual group members develop proposals through cognitive, communicative or physical activity.
        \item Exploration - individuals perform behaviors to gain information and test strategies.
        \item Alternative generation - responding to issues and problems that occurred during implementation or planning phase.
        \item Evaluation - discussing (intermediate) results during an activity, including individual contributions, as well as possible learning for future activities.
        \item Choice selection - deciding for a single proposal out of multiple options, for example through a voting procedure.
\end{itemize}

\subsubsection{Individual-level processes} describe dynamics within group members' states which result from or drive the current group activity:
\begin{itemize}
        \item Changing engagement - reducing or increasing participation in group and task processes, can, for example, be caused by physical or cognitive exhaustion or excluding behaviors of other members.
        \item Changing motivation - reducing or increasing effort, for example, due to dis-/encouraging feedback from others, or success/failure experiences within the task.
        \item Changing affect - includes changes of the emotional state due to need fulfilment, but also, for example, becoming stressed due to a tightening time limit
        \item Learning - focusing and practicing a certain topic to increase skill and knowledge.
        \item Fatigue - losing physiological or cognitive resources due to ongoing activity.
\end{itemize}
There might seem to be a high overlap of some processes with certain individual outcome factors. However, we separate the process of "getting angry" or "loosing motivation" from the final state of "being angry" or "being de-motivated" after an activity, as the latter might transfer to future activities, potentially with different tasks or even different groups.

\subsubsection{Interpersonal-level processes} consolidate dynamics in the interaction between two or more group members: 
\begin{itemize}
        \item Argument - debating for and against opinions present among the group members.
        \item Negotiation - group members make alternating proposals that approach a trade-off between each other's needs.
        \item Competition - group members try to perform sub-tasks better than others, even though this might not be necessary for the task.
        \item Verbal conflict - group members attack each other with words
        \item Physical conflict/fight - group members attack each other physically
        \item Bullying - group members intentionally cause another member's discomfort.
        \item Disagreement - individual members or sub-groups cannot find a common ground and the group will need follow at most only one of the preferences.
        \item Disruption - a group member interrupts an ongoing process, which might require adaptation of actions for the other members.
        \item Feedback - group members react to an action of individual members.
        \item Recognition seeking - an individual communicating to receive praise from the group
        \item Leader-Follower/Power Dynamics - a group member uses its special role or authority to get others to follow its opinion or plan.        
        \item Inspiration - some members perform actions or tell stories with the target to motivate the group.
        \item Affect management - actions to release tension, counter disappointment, increase happiness of group or individual members
        \item Solidarity Display - group members become aware of negative affective states or experiences of a member and provides emotional support through communication.
        \item Help - group members become aware of struggles of a member and provides support through actions.
        \item Protection - group members take actions to prevent another individual from being exposed to risk or high workload or influence the group processes as to provide special benefit to this individual.
        \item Relating - a group member takes the perspective of another one to improve understanding.
        \item Inclusion - a group member performs an action towards another (side-)participant that allows it to participate/increase participation.
        \item Exclusion - group members engage in an activity that prevents another one to participate due to lack of skills or knowledge or actively prevent it from contributing to the activity.
        \item Norm negotiation - group members work on defining which actions should be generally reinforced and discouraged for any activities within this group.
        \item Grounding - group members discuss each other's understanding and work towards a shared meaning of ambiguous information.
        \item Teaching - a group member tries to convey knowledge or skill information towards others to trigger learning processes.
        \item Introduction - group members that are not yet known to everyone in the group provide information about themselves (both task-relevant and personality-wise).
        \item On-boarding - group members that are not yet known to everyone in the group are introduced by another member.
        \item Story exchange - sharing real or imagined events with the group to improve knowledge about individual personalities and backgrounds.
        \item Sharing/listening - a group member talks about a personal issue or experience while others actively listen and provide the feeling of "being heard".
        \item Handling misunderstandings - group members discover communication flaw between each other and use a different communication strategy to try to correct this.
        \item Bonding - group members interact to deepen their relationship through affective touch or other interactions.
        \item Flirting - group members interact to signal romantic or sexual interest in each other.
        \item Play/Joking around - group members engage in playful interaction/communication aside from the task
        \item Experience sharing - group members perform or expose themselves to a sub-activity or external event together to create shared memories and affective states.
        \item Socializing - group members are pursuing the proximity of others without interaction.
        \item Co-Living - group members perform non-task related actions around others, while actively trying to not interfere with them.
        \item Incidental Interaction - group members unintentionally interfere with one an-other's actions.
\end{itemize}

\subsubsection{Group-level processes} change the structure or underlying aspects of the whole group and they cannot be assigned to any members or their conscious actions (although an effect resulting from the process might be):
\begin{itemize}
        \item Changing Group composition - a group member leaving the group during an activity or returning from an absence
        \item Sub-group formation - due to ongoing group processes and repeated structured interactions compositions of group members start to separate from the rest of the group.
        \item Role emergence - through continuous interaction, certain communication roles \cite{Benne1948FunctionalMembers} are establishing.
        \item Norm diffusion - all group members intuitively follow group norms.
        \item Evaluation apprehension - group members are becoming resistant to sharing ideas due to the concern of being negatively judged.
        \item Social loafing - group members start exerting less effort due to working in groups than they would  working individually \cite{Simms2014SocialLiterature}.
        \item Groupthink/conformity - group members start following other members opinions due to own uncertainty or desire to fit in the group \cite{Crutchfield1955ConformityCharacter.}.
        \item Social Facilitation - The presence of others (watching) causes changes in the way group members perform a task \cite{Bond1983SocialStudies}.
        \item Synchronization - group members sub-consciously start acting more similar to each other and with a common timing \cite{Oullier2008SocialBonding}. This is also involving behavioral mimicry effects \cite{Bavelas1986IAct, Chartrand1999Chameleon}.
        \item Inter-group interaction - the group encounters and interacts with another group, group members act based on the group's identity instead of their own.
        \item Group emotional contagion - the transfer of moods among people in a group \cite{Barsade2002TheBehavior}.
\end{itemize}

\subsubsection{Proxy Factors} 
As discussed before, we find a number of measurement constructs that are frequently used to describe group activities, but which rather evaluate group processes instead of their outcomes. 
We have termed these "Proxy Factors", as they likely have a high correlation with specific outcomes. One common example is distribution of contributions among the group members, which has been shown to be a big factor for group performance in problem solving tasks \cite{Woolley2010EvidenceGroups}. This makes proxy factors a potential measure for mediators to evaluate their progress during an activity or to assess the need for an intervention. However, many of the reviewed robotic mediator concepts seem to also use them as the main target of a mediation process, while we believe that mediation should always aim for an improvement of a lasting outcome of the group activity. 
Here is a list of common proxy factors analyzing group process qualities that can be assessed from observations:
\begin{itemize}
    \item Interactions Quantity - counting how often two or more team members engage with each other, for example through joint action, conversation, or eye contact.
    \item Interaction Qualities - there is a variety of ways to measure the quality of individual interaction between team members, for example conversational involvement \cite{Coker1987}, immediacy \cite{Andersen1979TheImmediacy}, describing intimacy and closeness of interactions, or the synergy of joint actions \cite{Wollstadt2022}.
    \item Engagement - is a measure for how much individual members are physically, mentally and emotionally involved in a group activity \cite{Hoffman2004CollaborationTeams}.
    \item Participation Balance - evaluating the distribution and relative length or frequency of activities of individual members during the group activity \cite{Woolley2010EvidenceGroups}.
    \item Contribution Balance - focuses on how much each group member's individual activities contributes to a joint results or gets represented in the group's strategy towards a target.
    \item Conversation/Interaction Flow - describing if interactions are interrupted by phases of silence or in-activity, which might signal misalignment of plans, understanding or interests.
    \item Consensus (Conformity/Agreement) - measures similarity of group member's opinions and decisions within an activity. This can include assessing conformity effects, where individuals might change opinions to fit with the group \cite{Asch1963, Castore1978DeterminantsDecisions}.
    \item Socialness of behavior/ Helping - observing the quantity and/or quality of individual actions that support other members in reaching their sub-goals or that aim to improve other member's state without own benefits (e.g. \cite{Bachrach2006EffectsPerformance, Webb2003PromotingGroups}).
    \item Constructiveness of Behavior - measuring how much of the group's activities has been contributing to task progress \cite{Shen2018}.
    \item Number/Intensity of Conflicts - evaluating non-constructive physical or conversational interactions that results from disagreements between group members or represent violations of group or social norms.
    \item Coordination/Synchronization - the number of actions with interdependence in, for example, timing or sub-target selection, as well as more subliminal processes of multiple members using similar behaviors \cite{Bernieri1994InteractionalAffect} or finishing another one's sentences.
    \item Quality/Quantity of Affect Displays - observing frequency and intensity of behaviors that signal emotional states, for example smiles; can be used to assess the group's affective dynamics.
    \item Task Cohesion - Cohesion, the forces that bind a group together, is a complex construct that addresses multiple aspects with respect to group processes \cite{Abrams2020}. While the social aspects can be considered an outcome of a group activity and translate to future group interactions, there is also a part that is specific to the task at hand. This describes a group member's feeling with respect to the task as a group activity rather than an abstracted item and the similarity of this perception among the members.
    \item Adherence to Structure - if an activity involves a pre-defined structure, it can be evaluated if and how many of the individual steps are followed or how often the group deviates from it.
\end{itemize}

Another type of proxy factors describes subjective quality estimations of group processes from the perspective of the members:
\begin{itemize}
    \item Perceived Contribution - the subjective feeling of how much an individual contributed to the group's outcomes \cite{Gouran1973CorrelatesDiscussions}. This might be prone to attributional bias \cite{Bradley1978Self-servingQuestion}, the perception that one's own contribution was more important than objectively true.
    \item Perceived Cooperation/Conflicts - subjective feeling about intensity and number of conflicts during an activity \cite{Jehn2001ThePerformance}.
    \item Perceived Fairness/Justice - a group member's perception of fairness and justice of work-related processes, not necessarily based on a final outcome, but how the outcome was reached \cite{Colquitt2002JusticeClimate., Cropanzano2001MoralJustice}.
    \item Role Ambiguity - if group members are not clear about which roles they have to follow during an activity this can lead to inefficient actions and individual stress \cite{Bauer2000RoleLiterature.}.
\end{itemize}

Similarly, we can look at subjective measures that evaluate the mediation processes ("Mediation Proxy Factors").
\begin{itemize}
    \item Perceived Neutrality - the evaluation of the team members towards the neutrality of a mediator in a negotiation can correlate with individual satisfaction \cite{Szejda2019NeutralityPerspectives}.
    \item Perceived Usefulness - evaluating the subjective assessment of the group members towards how much of the outcome can be assigned to mediator interventions.
    \item Mediation Satisfaction - a group's evaluation of the mediation process can be correlated with overall satisfaction \cite{Alberts2005DisputantProgram}.
\end{itemize}

\subsection{Outcomes} \label{Appendix:OutcomeList}
We present a list of possible outcome factors.
We are aware that several of the concepts presented here, might share meanings, partially because they have not been defined consistently throughout the literature (e.g. for cohesion \cite{Abrams2020}).
We tried to add references to definitions and measurement constructs for assessing the different factors to guide the reader to further investigations.

Individual Outcomes are:
\begin{itemize}
    \item In-group Identification - evaluates the perception of an individual as a group member and the positive impact and importance of being in a group \cite{Abrams2020, Leach2008Group-LevelIdentification}. A related measure is perceived inclusion \cite{Jansen2014Inclusion:Measurement}, “the degree to which individuals experience treatment from the group that satisfies their need for belongingness and uniqueness” \cite[p. 1265]{Shore2010Inclusion}.
    \item Satisfaction - is considered an aggregate of users’ weighted reactions to a set of situational variables, including job/life satisfaction, process/job facet satisfaction \cite{Hecht1978TowardSatisfaction, Wanous1972MeasurementSatisfaction}. Specifically, it can be related to the satisfaction of psychological needs, such as esteem, security, or autonomy. A scale to assess it was for example presented in \cite{Sheldon2012NeedSatisfaction}. It might additionally be helpful to evaluate dissatisfaction explicitly, which was shown to not properly being assessed as the polar opposite of satisfaction \cite{Keyton1991EvaluatingVariable}. It has been shown to have a high correlation with another possible outcome -- well-being \cite{Ryan2000Self-determinationWell-being, Sheldon2006Needs}. 
    \item Well-being - is a complex construct covering different aspects of assessment of one's life quality, such as evaluative well-being (or life satisfaction), hedonic well-being (or moods), and eudemonic well-being (sense of purpose and meaning in life) \cite{Cooke2016Wellbeing}.
    \item Motivation (long-term) - can be disrupted by various non-supportive conditions. It can be enhanced e.g. by positive performance feedback, but different effects of extrinsic and intrinsic rewards have been discussed \cite{Ryan2000Self-determinationWell-being}. It has also been related to self-esteem and perceived competence \cite{Deci1981IntrinsicMotivation, Harter1982SelfEsteem}, which in turn enhance intrinsic motivation in an environment with sufficient autonomy \cite{DeCharms1983Motivation}.
    \item Knowledge - learned and successfully re-callable information.
    \item Individual Skill - including social, cognitive, and physical skill gains after learning, practicing, or observing.
    \item Mood - is describing longer-term, global affective states including being happy or stressed or increased self-esteem \cite{Desmet2016Mood}.
    \item Enjoyment - "a positive feeling, when engaged in a pleasurable activity" ( \cite{Davidson2023TheScale} pp 17742). It has many correlations with other constructs, such as need satisfaction or motivation and affect but in particular leads to an increased interest of repeating the experienced group activity or group embedding.
    \item Social State/Status - describes the general relationships over all the groups a person is part of. In terms of social hierarchy, it has been determined based on sociograms (of the group but also among a wider range of related individuals) \cite{Leung2006Sociogram}. A related construct is social loneliness \cite{Green2001Loneliness}, which is correlated with the size and quality of groups that a person is part of.
    \item Physiological States - cover safety measures guaranteeing physical integrity, means to improve health conditions as in group rehabilitation, as well as states with more immediate, short-term effect, such as being exhausted or warmed-up.
\end{itemize}
Group Outcomes are:
\begin{itemize}
    \item (Social) Cohesion - describes the strength of the forces that keep a group together, including similarities, relationship qualities and involvement of its members \cite{Carron1985TheQuestionnaire, Dion2000GroupConstruct, Salas2015Cohesion, Severt2015OnCohesion}. It has been considered a strong influence factor for team performance. Sometimes the construct is further divided into task-related and social-related aspects \cite{Abrams2020}. This construct focuses on parts that will transfer to a next group interaction, while explicit task cohesion might rather be considered a measure of ongoing group process within an individual task \cite{Marks2001AProcesses} and is considered a proxy factors instead.
    \item Interpersonal Relationships - describes the closeness and affective connection between two group members. On the group scale this could be assessed for example using sociograms \cite{Hale2009Moreno} or inter-individual ratings of warmth and competence \cite{Aragones2015MeasuringLevels}. A measure for individual pair-relationships is the “inclusion of other in self” construct \cite{Aron1992InclusionCloseness}. For closer partnerships, intimacy measures can also be used to evaluate relationship strength \cite{Schaefer1981Intimacy}. A general overview of different measures can be found in \cite{Dunn1995Relationships}. Another very related construct is rapport \cite{Tickle-Degnen1990TheCorrelates}.
    \item Mutual Understanding/Cross-Understanding/Shared Mental Models/Transactive Memory - describes how much group members know about each other, allowing them to adapt to each other's cognitive states and intentions. A shared mental model is the common understanding of the environment and task context among the group members \cite{Klimoski2016}. Cross-understanding in turn tries to assess the awareness of each other's mental models of different aspects \cite{Huber2017}. Transactive memory describes the efficient distribution of relevant knowledge across a team \cite{Wegner1987TransactiveMind}. Another, more general, construct is familiarity \cite{Janssen2009Familiarity}.
    \item Group Trust/Team Climate - Group Trust is a belief about the dependability of the people in the group and willingness of a member to be vulnerable to the actions of the group \cite{Mayer1995Trust}. For dyads, there exists some common measures like the Dyadic \cite{Larzelere1980TheRelationships} or Interpersonal \cite{Johnson-George1982MeasurementOther} Trust Scales. Team Climate describes the perceived norms, attitudes, and expectations in a group and their implications for individual actions \cite{Schneider2013TeamClimate}. A related construct is psychological safety, the perceptions of the consequences of taking interpersonal risks \cite{Edmondson1999PsychologicalTeams}.
    \item Team Confidence - covers two related constructs: team efficacy is the shared belief in a group’s collective capability to execute what is required to produce a given level of performance \cite{Kozlowski2006EnhancingTeams, Salanova2003}. Group potency is the collective belief regarding the team’s ability to be successful in general \cite{Guzzo1993Potency}.
    \item Norms \& Roles -  especially during the early phases teams have to establish their own norms and rules that they want to use through all future assignments \cite{Tuckman1965DevelopmentalGroups}. Every resulting agreement can be considered an outcome of the group processes. Similarly, groups might aim to establish a certain distribution of roles within a group or towards the outside that transcends specific tasks or is relevant along a larger task construct.
\end{itemize}
Possible Performance Outcome Measure are:
\begin{itemize}
    \item Solution Quality - task outcomes that can be evaluated based on an explicit metric. For some tasks, this might also reduce to the binary quality of success/failure.
    \item Solution Quantity - task outcomes that cannot (yet) objectively evaluated for their quality. A higher number of results is usually considered preferable, for example in creativity tasks.
    \item Solution Speed - Outcome quality is usually important, but sometimes creating sub-optimal solutions in a short time can be desirable, for example to beat an opponent or in tasks with temporally limited resources and opportunities.
    \item Constraint violations - sometimes, one can reach a good result in many different ways. Instead of evaluating the outcome, it might also be relevant to rate the approach, in particular, if considering negative collateral effects. Negative evaluations can also result from violating explicit or implicit constraints set by a superior entity.
\end{itemize}
Finally, there is the construct of \textit{entitativity}, which measures the "groupness" as perceived from the outside \cite{Abrams2020}, which can be a relevant outcome, if the group is to interact with the outside, but one might also argue, that it should rather be considered a proxy factor, as its assessment includes observation and evaluation of processes, such as joint actions or spatial relation dynamics.

\subsection{Mediator Inputs \& Interventions} \label{Appendix:InterventionList}

\subsubsection{Specific input factors for the mediation concepts that we consider in this review}
\begin{itemize}
    \item Role - although we require a mediator to have a specific role with respect to the group, there are multiple ways in which it can be involved in a group activity.
    \begin{itemize}
        \item Facilitator - the classical role of (human) mediators, which come from outside of the group to help the group reach their goals
        \item Group Member - the mediator can also work on improving group processes from within the group, holding a specific role \cite{Benne1948FunctionalMembers}. One example could be as the leader of a work team.
        \item Bystander - group processes can also be influenced by people that are not directly interacting with the group members. As long as such a bystander chooses its actions intentionally to influence the group, this can also be considered mediation.
        \item None - some concepts might act in the background and cannot be assigned a particular role in the perception of the group. This can likely only be relevant for robotic mediators due to their flexible form factors.
    \end{itemize}
    \item Reputation - what group members know or infer about a mediator's skills and targets might influence the success of its interventions. It is, for example, considered very important for conflict mediation that both parties perceive the mediator as neutral \cite{Astor2007MediatorPractice}. Likely, this aspect will mainly influence concepts that perform explicit mediation. We structure reputation along task skills/experience and mediations skills/experience, as well as mediators that are known to be novice. 
    \item Relation - targets the understanding of the group with respect to the alignment of the mediator's targets for the group with their own targets. If these are fully aligned, the mediator might even be considered a group member. The classical mediation setting assumes mediator neutrality, while in a business setting there might also be an outside party requesting the mediator to facilitate certain outcomes. We might also find cases, where the mediator has a special relationship to one or a few individuals within the group.
    \item Capabilities - generally refer to specific skills and tools available to a mediator. In the context of robotic mediators, we mainly focus on implications of robotic interfaces, sensors and algorithms that influence how it can interact with a group.
    \item Embodiment - this category is not very important for human mediators (although they might also act through a communication interface or telepresence device instead of being co-located with the group), but it will play a stronger role for robot mediators. We use the following five categories here, although there might be other additional (sub-)categories (e.g. anthropomorphic vs humanoid, or zoomorphic \cite{Fong2003ARobots}) that could play a role:
    \begin{itemize}
        \item Tabletop Robots - most common social robot morphology, used to establish a physical social presence without non-verbal functional affordances. Can be modelled to resemble a human head as in Furhat \cite{AlMoubayed2012Furhat:Interaction}, an animation character as for Haru \cite{Gomez2018}, or using just basic shapes as for Keepon \cite{Kozima2009Keepon:Entertainment}.
        \item Mobile/Functional Robots - are mostly designed to approach certain tasks and have the affordances to physically interact with the world. Typical examples will be wheeled platforms that can move through a space or robotic arms that can move objects on a table.
        \item Humanoid Robots - resemble human body shapes, usually including a head, arms and potentially legs or at least the ability to move around. This category covers both android designs, such as ERICA \cite{Glas2016ERICA:Android}, and more robotic humanoids, like Nao.
        \item Actuated Objects - designed to resemble an everyday object such as a lamp \cite{Hoffman2015}, seat \cite{Takeuchi2014Whirlstools:Affordance}, or microphone \cite{Tennent2019} with some additional means to move parts to interact with its surrounding. 
        \item Virtually Embodied Agents \cite{Thellman2016PhysicalInteraction}/unspecified embodiments - for interaction with remote groups through a video communication platform or groups acting in a virtual world, the artificial agent might be embodied through a virtual avatar. Some concept designs might also not involve a specific requirement for the final embodiment.
    \end{itemize}
\end{itemize}

\subsubsection{Mediation processes}
Based on prior proposals on intervention types in human mediation \cite{Bostrom1993GroupSystems, Clawson1993TheMeetings, Dickson1996FacilitatingEnvironment, Moore2014Mediation}, group facilitation \cite{McFadzean2002DevelopingCompetencies}, community facilitation \cite{Tarmizi2005APractice}, team coaching \cite{Hackman2005ACoaching}, and human group leadership \cite{Becker2016TeamarbeitTeamentwicklung}, we compiled a list of mediation intervention categories in three categories. 
In \textit{task interventions} the mediator works on functional, task-level group dynamics, \textit{interactional interventions} influence and are triggered by interpersonal or group-level processes, and un-directed \textit{contextual interventions} aim to change the framing and atmosphere of the activity:
\begin{itemize}
    \item Task interventions
    \begin{itemize}
        \item Structure activities - the mediator helps the group to approach the task following established processes. This also involves time management, and the explanation or definition of processes and advice how to follow them.
        \item Moderation - the mediator guides the group through an established agenda.
        \item Support Problem identification - the mediator introduces a task and make sure that all group members understood the conditions, potentially provides further background information, possible sub-tasks and open questions.
        \item Keeping focus/Thread management - the mediator keeps the group moving towards the current target, recognizes tangential activities and redirect the group towards task-progressive ones.
        \item Summarize progress - the mediator keeps track of the group processes and task advancements and regularly or on request provides an overview to the group. This also includes identifying arguments and options and possibly expansion and rephrasing and requests for clarifications. It acknowledges significant moments and good ideas, highlights remaining sub-tasks, and might create a retrospective narrative about the group's approach.
        \item Facilitate decision making - includes, listing decision candidates and guiding the group to consensus by proposing and moderating decision-making methods.
        \item Encourage idea generation - the mediator can provide discussion cues, promote generative discussion contributions and exploratory thoughts, and encourage individuals to expand on their ideas.
        \item Support diverse perspectives - includes highlighting the benefits of diversity for the group, actively respecting different backgrounds, and asking to explicitly incorporate them in proposals. the mediator might also propose and assign, or even represent relevant personas for perspectives not present in the group.
        \item Provide information \& Task support - a mediator with expertise in the current group task might support by demonstrating task behavior in an educational way, suggesting resolution strategies or useful actions if the group gets stuck, or search for and share additional information or actively helping with certain sub-tasks. In particular, if the task is part of the mediation concept, the mediator also has access to interventions that provide guidance, hints, or even solutions for sub-tasks, and can correct wrong information and make sure the group is keeping task boundaries.
        \item Assign roles - the mediator can support the group during role development, both with respect to task and potential sub-task distribution or leader assignment, and in working out and highlighting benefits of positive communication roles.
        \item Provide and explain tools/skills - group process can benefit from a range of technical and methodological tools; the mediator can help understanding and applying/using them.
        \item Evaluate performance/provide feedback - the solutions provided by the group might be directly assessed with respect to quality, quantity and/or speed through the mediator. It might also apply certain tests during the activity to assess intermediate performance or other group targets, as well as point out or facilitate discussion on learnings for future activities.
        \item Provide new task - if the mediator has certain control on the task of the group, it might assign a new task or propose alternative task options if group processes fulfil/miss certain qualities in the current task. One option could be to adjust task difficulty or add side-goals to improve the chance for successful learning activities.
        \item Spark conversational topic - similarly, during social activities, it might support task progress by proposing new conversational topics.
        \item Kick-start - for certain tasks or after certain events, there might be an initial phase of uncertainty which leads to silence or in-activity of the group members. A mediator can break such state through initiating a first action or starting a conversation.
    \end{itemize}
    \item Interactional interventions
    \begin{itemize}
        \item Participation management - the mediator might encourage participation of group members to improve a certain balance. Alternative interventions are the coordination or modulation of turn taking, or explicit speaker appointment and speaking time limitation. 
        \item Handle misunderstandings - if a mediator detects failed communication that leads to diverging mental representations, it can highlight the existence of misunderstandings, propose an alternative description of the information, or open a discussion for correction. If not corrected early, misunderstandings might cause further issues, and when only addressed/noticed at that time, the mediator might also need to manage resulting discomfort or dissatisfaction with the communication and positively frame learnings.
        \item Promote mutual understanding - being aware of other group members backgrounds, intentions and issues can support positive group outcomes. The mediator can support this through ask relating, affect-sharing or relationship-reflecting questions, or explicitly share states and current and past activities of group members with the others. Similarly, it can make intentions (but also expectations) of participants transparent or ask questions about preferences. Finally, it might encourage positive regard for the experience and perception of everyone.
        \item Solicit feedback - the mediator can assist the group in reflection on its experience and encourage constructive criticism.
        \item Manage conflict - for groups that start in a conflict situation, the mediator will help or prepare compromises and highlight commonalities and identify sources of conflict. If conflict is not part of the task, the mediator will usually provide interventions for different types of conflicts. It has to handle interruption and manage disruptive behavior. It might provide or ask for repairing actions and if it cannot be avoided even try to divert the attention of the group from the conflict. 
        \item Aid in emotion support - if group processes trigger emotions or group members enter an activity in certain affective states, the mediator can facilitate empathy behaviors of other group members and contribute with own empathetic behavior. It might also point towards and help to weaken potential sources of negative affect states within ongoing group processes.
        \item Manage interpersonal processes - the mediator can help to improve and establish social interactions between group members. It can encourage and reward helping behaviors and intervene or correct negative interactions. The former includes encouraging and providing help for maintaining active listening behaviors. It can help the group establish and keep social interaction norms and criticize violations. It might also use opportunities to teach social skill and raise awareness of effects of social behavior. 
        \item Enable interactions - the mediator might change or highlight aspects in the environment to guide joint attention or change the spatial relations between group members to ease or trigger initial interactions or enable interactions with better quality. It might also explicitly guide the individuals towards collaborative or joint activity. 
        \item Frame contributions - sometimes contributions with no direct connection to a task can still improve the functioning of the group, and the mediator might highlight these positive effects to make others appreciate this member's activity.
        \item Encourage group identity - to enhance in-group identification and cohesion, the mediator can point out similarities among the members, support the establishment and adherence to team values and processes, encourage trust in the group and the experience of the participants as well as mutual support. Group traditions, for example for welcoming new members can also be established and promoted by a mediator, as can be differences to outgroups.
        \item Manage relationships - to provide initial support for creating or enhancing interpersonal relationships, a mediator can determine likely matches with respect to common interest and provide mutual introductions or refer people to one another.
        \item Team building - To strengthen the overall bonds among all group members, the mediator might provide opportunities and task that help to build shared memories and experiences and facilitate trust building.
        \item Encourage learning - to help improve individual skills and knowledge and improve performance of future group activities, the mediator can highlight or provide opportunities for learning or practicing skills and encourage the exchange of learnings among the group members. 
        \item Promote mediating actions - sometimes group members provide positive mediating actions themselves, a mediator might want to recognize and reward such attempts.
        \item Transfer messages - the mediator might act as a liaison to handle communication between group members with strongly different backgrounds to prevent misunderstandings or to provide a certain anonymization or privacy filter for group members.
        \item Inclusion - certain group members might require additional attention to participate in group processes, for example due to skill or knowledge limitations. A mediator might provide this support or adapt processes to people with special needs or it raises awareness of such needs to allow the group to actively integrate people into the group. In addition, it might work to prevent separation of the group or a dynamics that can lead to the creation of subgroups. 
        \item Moderate power dynamics - highlight and encourage positive effects of contributions from people with different hierarchical levels and prevent negative effects from excessive use of power.
        \end{itemize}
    \item Contextual interventions
    \begin{itemize}
        \item Welcoming/Closing - to provide a framing for the activity, the mediator might make statements to welcome the group, start the task or respectively close the activity and bid farewell to the group.
        \item Creating open atmosphere - the mediator makes statements that it will promote psychological safety, mutual trust, and communicate procedural justice during the activity. Other interventions can be ice-breaking (i.e. non-task) games or telling jokes.
        \item Asking in place of the group - to reduce possible anxiety for embarrassing situations among the group members and prevent negative group processes in response to difficult or controversial contributions, a mediator might perform the contribution or ask a question in place of the group (act as "devil's advocate" or "scapegoat").
        \item Motivating - to (re-)activate a group's energy, the mediator might highlight the direct or societal impact of performing the task, recall prior successful activities of the group or provide additional incentives.
        \item Reducing stress - similarly, the robot might call for a break or provide relaxing side tasks to reduce stressful situations.
        \item Mirroring affective atmosphere - to help the group become aware of ongoing group processes and their effects, a mediator might mirror the (majority of the) group's affective state
        \item Active presence/active listening - to be perceived as (positively) contributing to the group, a mediator might need to perform subtle actions that keep a group aware of its presence and shows active following of the group's actions. Such actions can be verbal or non-verbal signs of interest in a topic, backchanneling  and attentive gaze or body poses, as well as eye contact \& shared attention and mimicry of facial expressions and postures of group members. It can also provide physical clues of its presence through haptic communication (hugging, touching hands), non-task interactions with the environment and non-verbal behavior routines.
        \item Showing own state - to establish and be transparent about its own intentions and goals, a (non-neutral) mediator might show signs of (dis)agreement with the group or affective states triggered by group processes. It might also disclose own initial conditions (e.g., mood) that potentially affect its behavior during an activity.
    \end{itemize}
\end{itemize}

Additional aspects that are to be considered within mediation processes are:
\begin{itemize}
        \item Modalities - depending on the mediator's capabilities, it might use different channels to provide its interventions:
        \begin{itemize}
            \item Speech - the intervention is delivered through verbal communication (or text messages).
            \item Non-speech utterances - the mediator produces sounds which are not speech, but still imply meaning for an intervention. Examples can be backchanneling  or affect-implying sounds.
            \item Gesture/Facial expression/Posture - communicative body expressions of the mediator, often used to emphasize spoken messages or transmit emotional meaning.
            \item Gaze/Attention - mediator motion, usually of the head/eyes or another explicit sensing body part that is oriented towards a specific group member or object.
            \item Social touch - explicit physical interaction between the mediator and group members, communicating socio-affective states.
            \item Physical action - physical interaction of the mediator with the environment or to progress a task, which might provide an intervention through demonstration or an implicit message.
            \item Other - technical communication modalities that might be specific for robotic mediators, such as light signals, projection, change of body shape, or activation of other technical means.
        \end{itemize}
        \item Trigger Condition - the mediator will observe the different levels of group processes and states to determine when and how to intervene in which way. Many proxy measures can also be used as trigger conditions as they imply positive/negative qualities of group processes. 
        \begin{itemize}
            \item (Individual level) Participant Intention/Need - if the group processes prevent one member from following its intentions or fulfilling its needs, the mediator might want to intervene. Examples could be a participation intend being blocked by ongoing speaking of another member, or the need to feel included in the group, when other members are having a conversation about a topic of only their expertise.
            \item (Individual level) Individual Skill - if a team member cannot follow the progress of the rest of the group or is demonstrating the lack of a (social) skill that is considered relevant for the group activity, the mediator might facilitate a change of approach for the group.
            \item (Individual level) Participant State/Activity - certain specific activities or states of individual members might signal an opportunity for an intervention of the mediator, for example to highlight synergies.
            \item (Individual level) Engagement/Affect - similarly, the affective state or engagement of individuals can provide information about an imbalance in the current group process or an opportunity to help the group to address it on their own.
            \item (Individual level) Contribution Semantics - some types of contributions of individual members can negatively or positively influence the following group processes and might thereforee be dis- or encouraged by a mediator. Examples can be personal attacks towards others or demonstrating prototypical behavior to be followed by others.
            \item (Interpersonal level) Participation Statistics - as there exists some evidence that a certain balance of participation or contribution among the group members can be beneficial \cite{Woolley2010EvidenceGroups}, deviations from an equal distribution can be used to signal a mediator a need for intervention.
            \item (Interpersonal level) Conversation Flow - conversations where only sub-groups interact with each other might undermine overall joint processes. Moments of "awkward silence" can also show missing commonalities or mutual understanding. 
            \item (Interpersonal level) Conflict Detection - supporting the resolution of conflicts is a classical task for a mediator. In many cases such conflicts might be the reason for a group to come together. However, conflicts can also appear as part of the group processes during an activity and might act as a trigger for the mediator to act. This also includes instances of group or social norm violations which might lead to unwanted side effects and should be addressed by the group or the mediator directly.
            \item (Interpersonal level) Common Interests - detection and highlighting of common interest among group members can help with group formation and stabilization processes.
            \item (Group level) Group Affect - affective states of multiple group members and towards the group or related processes can be assessed on a holistic level and might point towards certain group dynamics \cite{Menges2015GroupProcesses}.
            \item (Group level) Group Structure Changes - sometimes group members might leave or not be available  for a certain time during an activity. The missing of an existing or establishment of new role distributions during an activity can also aid group processes and might be an important source of information for a mediator.
            \item (Task level) Performance Targets - many tasks provide access to intermediate information predictive of the final outcome. Particularly low (or high) values might point mediators to a need for intervention.
            \item Explicit Request - in complex situations (or with limited sensing capabilities) a mediator might not always be able to follow all group processes. It can therefore be beneficial to also allow the group member to actively request an intervention or point out a problematic process or state towards the mediator.
            \item Pre-determined Structure - if the group activity follows a strict structure that has been set up prior to it, a mediator intervention can be explicitly made part of such structure, for example, to facilitate a switch of task context or simply handle an initial introduction phase.
            \item Task State - similarly, if the mediator has control over the task (for example during team building activities), interventions might be known to be most effective during certain states within this task and can therefore be triggered independent of group processes.
        \end{itemize}
        \item Intervention Style - this aspect describes how transparent a mediator makes its interventions. For explicit interventions the group will be aware that the mediator tries to influence their processes. In contrast, implicit interventions try to bias the group through background activity. Two other ways are to provide an "un-directed" intervention, which the group can take up, and "accidental" interventions, where the changes can be considered a side-effect of a non-mediation-focused action.
\end{itemize}

\section{Experiment Categorizations - Details} \label{Appendix:ExperimentList}
Although the main focus of this review is on the concepts for robot mediation, we have also analyzed the experiments that have been performed and how the concepts have been implemented. 
The details for every paper can be found in tables \ref{tab:allpapers_experiments}-\ref{tab:allpapers_experiments11}.
For some aspects we also used categories, which we would like to explain in the following.

We classified Experiment Types into one of four classes:
\begin{itemize}
    \item Design Only - the paper either just describes how the authors envision a robot mediator to act, while the main focus of the paper is on something else, or reports a methodological design study, for example involving users in drafting a mediator concept.
    \item Proof of Concept - the mediator concept is implemented and the paper reports on technical details, while tests with humans are, if at all, performed only to demonstrate the basic effect, but there is no experimental study.
    \item User study (Lab) - the concept is implemented and tested in a structured experiment with human groups inside a controlled research lab environment and includes an analysis of the resulting effects.
    \item Field Study - the concept is implemented and tested in an experiment with human groups in the native environment for the group/mediation activity, for example at private homes, retirement homes, or public spaces.
\end{itemize}
In addition, we check if the participant types targeted in the mediation concepts have been covered in the experiments (\textit{Participant Type Match}), for example if a concept for older adults is tested with students or a concept designed for established teams is evaluated with randomly recruited strangers.
We also report if the robot was working \textit{autonomous}ly or was controlled through an experimenter/expert (\textit{Wizard}-of-Oz\footnote{Named for convenience, although we did not evaluate the appropriate application of the method with the same name \cite{Steinfeld2009TheResearch, Kelley1984WoZ}}).

If the experiments have been performed with one or multiple control conditions (or "None"), these have been labeled as:
\begin{itemize}
    \item No Mediator - the groups performed the same activity without any additional support.
    \item Human Mediator - the groups performed the same activity with a human mediator.
    \item Ablated Robot - the groups performed the same activity with the robot mediator but with reduced functionality. This can, for example, include random intervention timing, reduced expressivity, or a simplified intervention behavior.
    \item No Embodiment - the groups performed the same activity with only a chat- or voice-based mediator intervention, i.e. functionality without a robotic embedding.
    \item Intervention Type - the groups performed the same activity with different mediation strategies implemented for the robot, for example comparing the transactional and transformational leadership styles \cite{Lopes2021SocialTeams}.
\end{itemize}
In addition, we report the number of experiment groups and participants (originally recruited) and if the study used an inter- or intra-subjects design and which measures have been reported. 
For measures, we focus on those concerning the mediation task (i.e. addressing outcomes or proxy factors). In particular for the use of questionnaires, we mention if the authors used a validated measure or performed validation themselves and if they make use of constructs that have been previously used by others (marked with \textbf{bold} font.
If the results showed a significant effect with respect to relevant baselines, these measures are marked \textcolor{lime}{green}, or, if there was at least some data showing the influence of the mediator, in \textcolor{cyan}{cyan} in tables \ref{tab:allpapers_experiments}-\ref{tab:allpapers_experiments11}.

\section{Mediator Concept Classification}
Four of the authors met to discuss the categorization of the social mediator concepts across the groups proposed in \ref{subsec:concepts}. 
For each paper we evaluated candidates and assigned it to all groups that were still upheld after a short discussion (final assignments, see Table \ref{tab:mediation-categories}). 

\begin{figure}
    \centering
    \includegraphics{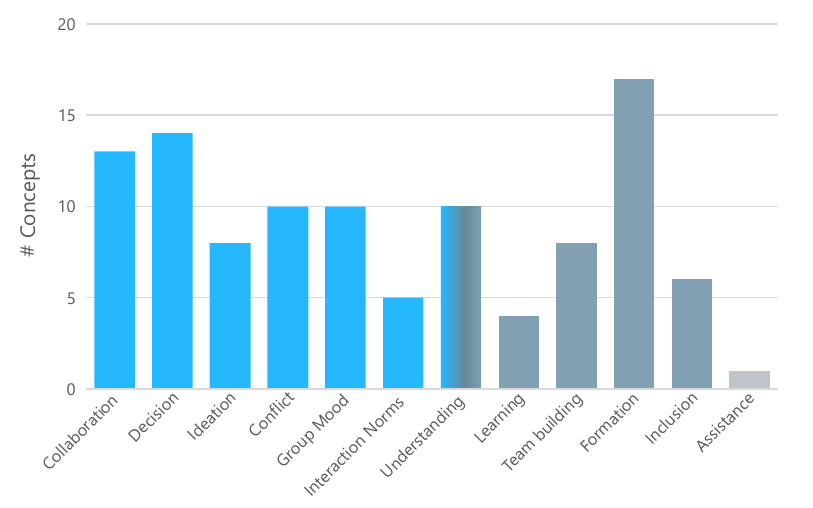}
    \caption{Distribution of social mediation concepts across the categories proposed in this paper.}
    \Description{Figure 11. Histogram showing how the reviewed concepts are categorized into the mediation roles introduced in this paper. The numbers are: Collaboration 13, Decision 14, Ideation 8, Conflict 10, Group Mood 10, Interaction Norms 5, Understanding 10, Learning 4, Teambuilding 8, Formation 17, Inclusion 6, Assistance 1.}
    \label{fig:Stats_Targets}
\end{figure}

About one fifth (16) of the concepts appeared to fall into multiple categories. Reasons for this were a more complex application story spanning multiple activities or the fact that the proposed categories are not necessarily exclusive in their formulation.
With the exception of the special category of Mediation Assistance, which only contains a single paper \cite{Shimoda2022ApplicationRehabilitation}, at least 4 concepts could be classified for any given group (Fig. \ref{fig:Stats_Targets}). 
Major clusters are Collaboration, Decision Support and Group Formation. 
More than half of the concepts (53\%) focus on providing immediate support for the group, 44\% work on more long-term dynamics and 7 concepts cover aspects of both.

\begin{table}[!htbp]
    \caption{Assignment of reviewed robot mediation concept to the proposed mediation categories}
    \label{tab:mediation-categories}
    \begin{center}
    \begin{tabular}{|p{0.16\columnwidth}|p{0.29\columnwidth}||p{0.16\columnwidth}|p{0.29\columnwidth}|}
        \toprule
        Category           & References                               & Category        & References \\
        \hline \midrule
        Collaboration      & \cite{Alves-Oliveira2019EmpathicStudy, BaghaeiRavari2021EffectsLearning, Charisi2021TheDynamics, Chen2022DesigningInterviews, Fan2021FieldAdults, Fu2017TurnTurn-Taking, Gillet2020AChildren, Jung2020Robot-assistedGroups, Kim2023ChildrobotBehaviors, Lopes2021SocialTeams, Shen2018, Short2017, Stoican2022LearningCollaboration}   & 
        Understanding      & \cite{Aylett2023, Birmingham2020, Dietrich2022, Erel2021, Hasegawa2014FacilitationExaggeration, Kobuki2023RoboticStudy, Nagao1994CommunicativeModality, Noguchi2023HowRecipients, Scassellati2018ImprovingRobot, Takano2009PsychologicalNeeds} \\ 
        \hline
        Decision           & \cite{AlMoubayed2013TutoringTutor, Alves-Oliveira2019EmpathicStudy, Basu2001TowardsSettings, Buchem2023Scaling-UpNAO, Fu2017TurnTurn-Taking, Gillet2022Ice-BreakersTeenagers, Hitron2022AIRobots, Li2023ImprovingGroups, Ohshima2017Neut:Conversations, Rosenberg-Kima2020Robot-supportedFacilitation, Shamekhi2018, Shamekhi2019, Shen2017RobotConformity, Tennent2019}  & 
        Learning           & \cite{Edwards2018ALearning, Gordon2022InvestigationAgent, Chandra2016ChildrensRobot, Mitnik2008AnMediator}  \\ 
        \hline
        Ideation           & \cite{Chen2022DesigningInterviews, Chew2023WhoInteractions, Cumbal2022ShapingBackchannels, deRooij2023Co-DesigningDynamics, Fucinato2023CharismaticCreativity, Ikari2020Multiple-RobotDiscussion, Ohshima2017Neut:Conversations, Tennent2019}  & 
        Team building       & \cite{Chen2022DesigningInterviews, Fu2021UsingConversations, Jeong2018, Moharana2019RobotsCaregivers, Short2017, Strohkorb2016ImprovingRobot, Traeger2020, Utami2019CollaborativeRobot} \\ 
        \hline
        Conflict           & \cite{Ablett2007BuildBot:Teams, Druckman2021WhoHumans, Hoffman2015, Jung2015UsingViolations, Kanda2012ChildrenRobot, Sadka2023AllConflict, Shen2018, Stoll2018KeepingFormat, Wang2023ExploringConflict}, \cite{Dorrenbacher2023TheSettings}:PARFAR     & 
        Formation          & \cite{Birmingham2020, Fan2021FieldAdults, Fukuda2016AnalysisMediator, Gillet2022Ice-BreakersTeenagers, Hayamizu2013AnControl, Isbister2000HelperSpace, Kim2023ChildrobotBehaviors, Kochigami2018DoesCommunication, Nakanishi2003CanCommunities, Ono1999, Pliasa2019CanDevelopment, Sadka2022ByOpening-encounters, Shin2021, Takeuchi2014Whirlstools:Affordance, Uchida2020ImprovingExperience, Xu2014, Zhang2023}  \\ 
        \hline
        Group Mood         & \cite{Fu2023AResolution, Karatas2020UtilizationArt, Leite2016AutonomousInteraction, Lopes2021SocialTeams, Matsuyama2010PsychologicalGame, Shimoda2022ApplicationRehabilitation, Sinnema2019TheInteraction, Traeger2020, Yamazaki2012AVisitor, Zheng2005DesigningGuide}    & 
        Inclusion          & \cite{Gillet2020AChildren, Inoue2021AParticipants, Matsuyama2015Four-participantParticipant, Neto2023TheChildren, Pliasa2019CanDevelopment, Sebo2020TheBehavior}  \\ 
        \hline
        Interaction Norms  & \cite{Kang2023TheGroup, Shim2017AnEvaluation, Tahir2020AConversations}, \cite{Dorrenbacher2023TheSettings}: SEYNO \& SEATY   & 
        Assistance         & \cite{Shimoda2022ApplicationRehabilitation}  \\ 
        \bottomrule
    \end{tabular}
    \end{center}
\end{table}

\newpage

\section{Reviewed Literature -- Details} \label{Appendix:AllPapers}

For each of the concepts included in this review, we provide the reference and an overview of input (Tables \ref{tab:allpapers_inputs}-\ref{tab:allpapers_inputs6}), output and mediator factors considered (Tables \ref{tab:allpapers_dynamics}-\ref{tab:allpapers_dynamics9}), as well as information about processes and experimental setup (Tables \ref{tab:allpapers_experiments}-\ref{tab:allpapers_experiments11}). 

\begin{landscape}
\begin{table}[!htbp]
    \caption{Overview of the input factors of all concepts included in this review (Part I)}
    \label{tab:allpapers_inputs}
    \centering
    \begin{tabular} {|p{0.1\columnwidth}|p{0.04\columnwidth}|p{0.15\columnwidth}|p{0.08\columnwidth}|p{0.09\columnwidth}|p{0.07\columnwidth}|p{0.12\columnwidth}|p{0.1\columnwidth}|p{0.09\columnwidth}|}
        \toprule
        Reference      & Group \newline Size & Member Specs  & Group \newline History & Relationships & Setting   & Task & Embodiment   & Role\\
        \hline \midrule
        Ablett et al \cite{Ablett2007BuildBot:Teams} & 3-10 & None & Established & Team & Dedicated Space & Performance & Mobile \newline (AIBO) & Facilitator\\ 
        \hline
        Al Moubayed et al \cite{AlMoubayed2013TutoringTutor} & 2 & None & Established & Team & Static & Decision & Tabletop (Furhat) & Facilitator\\ 
        \hline
        Alves-Oliveira et al \cite{Alves-Oliveira2019EmpathicStudy} & 2 & Teenagers & Established & Peers & Static & Learning/ \newline Planning/ \newline Intellective & Humanoid (Nao) & Facilitator\\ 
        \hline
        Aylett et al \cite{Aylett2023} & 2 x 3-10 & Teenagers & Forming & Mixed & Virtual & Leisure & Tabletop (Haru) & Facilitator\\ 
        \hline
        Baghaei Ravari et al \cite{BaghaeiRavari2021EffectsLearning} & 2 & None & Forming & Team & Virtual & Learning & Tabletop \newline (static Nao) & Facilitator\\ 
        \hline
        Basu et al \cite{Basu2001TowardsSettings} & 3-5 & None & Established & Team & Static & Decision/ \newline Cognitive Conflict & Actuated \newline Object (room) & Facilitator\\ 
        \hline
        Birmingham et al \cite{Birmingham2020} & 3-5 & Sharing Problem & Forming & Peers & Static & Sharing & Humanoid (Nao) & Facilitator/ \newline Group Member\\ 
        \hline
        Buchem \cite{Buchem2023Scaling-UpNAO} & 3-5 & None & Established & Team & Static & Decision & Humanoid (Nao) & Facilitator\\ 
        \hline
        Chandra et al \cite{Chandra2016ChildrensRobot, Chandra2015CanActivity} & 2 & Children & Established & Peers & Static & Learning & Humanoid (Nao) & Facilitator\\ 
        \hline
        Charisi et al \cite{Charisi2021TheDynamics} & 2 & Children & Forming & Peers & Static & Intellective & Tabletop (Haru) & Group Member\\ 
        \hline
        Chen et al \cite{Chen2022DesigningInterviews} & 2 & Parent/Child & Established & Family & Static & Leisure/ \newline Creativity & Tabletop \newline (Jibo) & Facilitator\\ 
        \hline
        Chew \& Nakamura \cite{Chew2023WhoInteractions} & 3-10 & Teenagers & Established & Peers & Static & Leisure/ \newline Cognitive Conflict & Tabletop (Haru) & Facilitator\\ 
        \hline
        Cumbal et al \cite{Cumbal2022ShapingBackchannels, Parreira2022, Gillet2022LearningInteractions, Gillet2021} & 2 & Language Skill Variation & Forming & Strangers & Static & Creativity & Tabletop (Furhat) & Facilitator\\ 
        \hline
        de Rooj et al \cite{deRooij2023Co-DesigningDynamics} & 3-5 & None & Established & Team & Static & Creativity \newline Decision & Humanoid (Nao) & Facilitator\\ 
        \hline
        Dietrich \& Weisswange \cite{Dietrich2022} & 2 & None & Established & Mixed & Casual & Sharing & & Facilitator\\ 
        \bottomrule
    \end{tabular}
\end{table}
\end{landscape}
\begin{landscape}
\begin{table}[!htbp]
    \caption{Overview of the input factors of all concepts included in this review (Part II)}
    \label{tab:allpapers_inputs2}
    \centering
    \begin{tabular} {|p{0.1\columnwidth}|p{0.04\columnwidth}|p{0.15\columnwidth}|p{0.08\columnwidth}|p{0.09\columnwidth}|p{0.07\columnwidth}|p{0.12\columnwidth}|p{0.1\columnwidth}|p{0.09\columnwidth}|}
        \toprule
        Reference      & Group \newline Size & Member Specs  & Group \newline History & Relationships & Setting   & Task & Embodiment   & Role\\
        \hline \midrule
        D\"{o}rrenb\"{a}cher et al \cite{Dorrenbacher2023} & & & & & & & & \\
        PARFAR & 2 & None & Established & Couple \newline Family & Casual & Cognitive Conflict & & Facilitator\\ 
        SEYNO & Crowd & None & Random & Strangers & Public & Co-existence & & Facilitator\\ 
        SEATY & Crowd & None & Random & Strangers & Public & Co-existence & & Facilitator\\
        \hline
        Druckman et al \cite{Druckman2021WhoHumans} & 2 & None & Established & Rivals & Static & Mixed Motive & Tabletop \newline (Telenoid) & Facilitator\\ 
        \hline
        Edwards et al \cite{Edwards2018ALearning} & 5-10 & Teenagers & Established & Peers & Static & Learning & Tabletop (Sota) & Facilitator\\ 
        \hline
        Erel et al \cite{Erel2021, Rifinski2021Human-human-robotInteraction} & 2 & None & Established & Friends & Static & Sharing & Actuated \newline Object (Lamp) & Bystander\\ 
        \hline
        Fan et al \cite{Fan2021FieldAdults, Fan2022SAR-Connect:Adults, Fan2017AModels, Fan2016AImpaired} & 2 & Older Adults & Forming & Peers & Static & Performance \newline Leisure & Tabletop \newline (static Nao) & Facilitator\\ 
        \hline
        Fu, Carolyn et al  \cite{Fu2017TurnTurn-Taking} & 3-5 & None & Established & Team & Static & Decision & Actuated \newline Object (table) & Facilitator\\ 
        \hline
        Fu, Changzeng et al  \cite{Fu2021UsingConversations} & 3-7 & None & Established & Friends & Static & Networking & Humanoid \newline (ERICA) & Facilitator\\ 
        \hline
        Fu, Di et al \cite{Fu2023TheCollaboration} & 2 & None & Established & Team & Static & Performance & Tabletop \newline (iCup Head) & Facilitator\\ 
        \hline
        Fucinato et al \cite{Fucinato2023CharismaticCreativity} & 3-4 & None & Established & Team & Virtual & Creativity & Humanoid (EZ) & Facilitator\\ 
        \hline
        Fukuda et al \cite{Fukuda2016AnalysisMediator} & 3-10 & None & Random & Strangers & Public & Learning & Humanoid \newline (Robovie) & Facilitator\\ 
        \hline
        Gillet, van den Bos et al \cite{Gillet2020, Gillet2020AChildren} & 3 & Children & Growing & Peers & Static & Creativity & Mobile \newline (Cozmo) & Facilitator\\ 
        \hline
        Gillet, Winkle et al \cite{Gillet2022Ice-BreakersTeenagers, Gillet2022AInteractions} & 3-5 & Teenagers & Established & Peers & Static & Creativity \newline Decision \newline Planning \newline Intellective & Humanoid \newline (Nao) & Facilitator\\ 
        \hline
        Gordon et al \cite{Gordon2022InvestigationAgent, Mizrahi2022Long-TermDiscussions., Mizrahi2022VRobotator:K-12} & 3-5 & Teenagers & Forming & Peers & Virtual & Learning \newline Cognitive Conflict & Virtual & Facilitator\\ 
        \bottomrule
    \end{tabular}
\end{table}
\end{landscape}
\begin{landscape}
\begin{table}[!htbp]
    \caption{Overview of the input factors of all concepts included in this review (Part III)}
    \label{tab:allpapers_inputs3}
    \centering
    \begin{tabular} {|p{0.1\columnwidth}|p{0.04\columnwidth}|p{0.15\columnwidth}|p{0.08\columnwidth}|p{0.09\columnwidth}|p{0.07\columnwidth}|p{0.12\columnwidth}|p{0.1\columnwidth}|p{0.09\columnwidth}|}
        \toprule
        Reference      & Group \newline Size & Member Specs  & Group \newline History & Relationships & Setting   & Task & Embodiment   & Role\\
        \hline \midrule
        Hasegawa \& Nakauchi \cite{Hasegawa2014FacilitationExaggeration, Hasegawa2014TelepresenceTeleconferences} & 3-10 & None & Established & Team & Virtual & Decision & Humanoid \newline (custom telepresence robot) & Facilitator\\ 
        \hline
        Hayamizu et al \cite{Hayamizu2013AnControl, Sano2009AControl} & 2 & None & Forming & Strangers & Static & Leisure & Tabletop \newline (custom) & Facilitator\\ 
        \hline
        Hitron et al \cite{Hitron2022AIRobots} & 2 & None & Random & Strangers & Static & Cognitive Conflict & Actuated \newline Object (Lamp) & Facilitator\\ 
        \hline
        Hoffman et al \cite{Hoffman2015, Zuckerman2015EmpathyCompanions} & 2 & None & Established & Couples & Static & Leisure \newline Cognitive Conflict & Actuated \newline Object (Lamp) & Bystander\\ 
        \hline
        Ikari et al \cite{Ikari2020Multiple-RobotDiscussion} & 4 & None & Forming & Peers & Static & Cognitive Conflict & Tabletop \newline (CommU) & Facilitator\\ 
        \hline
        Inoue et al \cite{Inoue2021AParticipants} & 2 & Multi-cultured & Forming & Strangers & Virtual & Leisure & Unspecific \newline (VEA\footnote{Virtual Embodied Agent}) & Group Member\\ 
        \hline
        Isbister et al \cite{Isbister2000HelperSpace} & 2 & None & Mixed &  & Casual & Sharing & Humanoid \newline (ERICA) & Group Member\\ 
        \hline
        Jeong et al \cite{Jeong2018} & 3 & Young Adults & Established & Friends & Virtual & Networking & Tabletop \newline (custom) & None\\ 
        \hline
        Jung, Martelaro, Hinds \cite{Jung2015UsingViolations} & 2 +1 & None & Established & Team & Static & Intellective & Mobile \newline (Arm + Base) & Group Member\\ 
        \hline
        Jung, DiFranzo et al \cite{Jung2020Robot-assistedGroups} & 2 & None & Established & Team & Static & Performance & Mobile \newline (Arm) & Facilitator\\ 
        \hline
        Kanda et al \cite{Kanda2012ChildrenRobot} & 2-5 & Children & Established & Peers & Dedicated Space & Intellective \newline Planning & Humanoid \newline (Robovie) & Facilitator\\ 
        \hline
        Kang et al \cite{Kang2023TheGroup} & 2 & None & Random & Strangers & Static & Mixed Motive & Actuated \newline Object (Table) & None\\ 
        \hline
        Karatas et al \cite{Karatas2020UtilizationArt} & Crowd +1 & Performer + Audience & Random & Authority & Dedicated Space & Leisure & Tabletop \newline (BB9e) & Facilitator\\ 
        \hline
        Kim et al \cite{Kim2023ChildrobotBehaviors, Kim2018DesigningOpportunities} & 2 & Children & Forming & Peers & Static & Learning & Tabletop \newline (Skusie) & Facilitator\\ 
        \hline
        Kobuki et al \cite{Kobuki2023RoboticStudy, Seaborn2023VoiceConversation} & 3-10 & None & Forming & Strangers & Virtual \newline Static & Leisure/ \newline Sharing & Tabletop \newline (custom) & Facilitator\\ 
        \bottomrule
    \end{tabular}
\end{table}
\end{landscape}
\begin{landscape}
\begin{table}[!htbp]
    \caption{Overview of the input factors of all concepts included in this review (Part IV)}
    \label{tab:allpapers_inputs4}
    \centering
    \begin{tabular} {|p{0.1\columnwidth}|p{0.04\columnwidth}|p{0.15\columnwidth}|p{0.08\columnwidth}|p{0.09\columnwidth}|p{0.07\columnwidth}|p{0.12\columnwidth}|p{0.1\columnwidth}|p{0.09\columnwidth}|}
        \toprule
        Reference      & Group \newline Size & Member Specs  & Group \newline History & Relationships & Setting   & Task & Embodiment   & Role\\
        \hline \midrule
        Kochigami et al \cite{Kochigami2018DoesCommunication} & 3-10 & None & Forming & Strangers & Casual & Leisure \newline Networking & Humanoid \newline (Pepper \& Nao) & Facilitator\\ 
        \hline
        Leite et al \cite{Leite2016AutonomousInteraction} & 3 & Children & Established & Peers & Static & Creativity & Tabletop \newline (Keepon (x2)) & Facilitator\\ 
        \hline
        Li et al \cite{Li2023ImprovingGroups} & 3-5 & Language Skill Variation & Established & Team & Virtual & Decision & Unspecific \newline (VEA) & Facilitator\\ 
        \hline
        Lopes et al \cite{Lopes2021SocialTeams} & 3 & None & Established & Team & Static & Planning & Tabletop \newline (EMYS) & Group Member\\ 
        \hline
        Matsuyama et al \cite{Matsuyama2010PsychologicalGame, Fujie2009ConversationCommunication, Matsuyama2008DesigningCommunication} & 3-10 +1 & None & Forming & Peers & Static & Leisure & Humanoid \newline (SCHEMA) & Group Member\\ 
        \hline
        Matsuyama et al \cite{Matsuyama2015Four-participantParticipant, Matsuyama2015TowardsFacilitation} & 3 & None & Growing & Mixed & Casual & Leisure & Humanoid \newline (SCHEMA) & Group Member\\ 
        \hline
        Mitnik et al \cite{Mitnik2008AnMediator} & 3 & Children/ \newline Teenagers & Established & Peers & Dedicated Space & Learning & Mobile \newline (custom) & Facilitator\\ 
        \hline
        Moharana et al \cite{Moharana2019RobotsCaregivers} & 2 & Care-giver/ \newline Care-receiver \newline Dementia & Established & Authority & Casual & Care-taking & Unspecific & Facilitator\\ 
        \hline
        Nagao \& Takeuchi \cite{Nagao1994SocialAgents} & 2 & None & Forming & Team & Casual & Sharing & Unspecific & Facilitator\\ 
        \hline
        Nakanishi et al \cite{Nakanishi2003CanCommunities} & 2 & None & Forming & Strangers & Virtual & Leisure & Unspecific \newline (VEA) & Group Member\\ 
        \hline
        Neto et al \cite{Neto2023TheChildren, Neto2021FosteringRobots} & 3 & Children \newline Visual Impairments & Growing & Peers & Static & Decision & Mobile \newline (Dash) & Facilitator\\ 
        \hline
        Noguchi et al \cite{Noguchi2023HowRecipients} & 2 & Older Adults/ \newline Families & Established & Family & Virtual & Sharing & Tabletop \newline (custom) & Facilitator\\ 
        \hline
        Ohshima et al \cite{Ohshima2017Neut:Conversations} & 3 & None & Forming & Strangers & Static & Leisure & Tabletop \newline (custom) & Facilitator\\ 
        \hline
        Ono et al \cite{Ono1999} & 2 & None & Forming & Peers & Casual & Co-existing & Mobile \newline (Pioneer) & Facilitator\\ 
        \hline
        Pliasa \& Fachantidis \cite{Pliasa2019CanDevelopment} & 2 & Children \newline Autism & Forming & Peers & Static & Leisure & Object \newline (Pillow) & Facilitator\\ 
        \bottomrule
    \end{tabular}
\end{table}
\end{landscape}
\begin{landscape}
\begin{table}[!htbp]
    \caption{Overview of the input factors of all concepts included in this review (Part V)}
    \label{tab:allpapers_inputs5}
    \centering
    \begin{tabular} {|p{0.1\columnwidth}|p{0.04\columnwidth}|p{0.15\columnwidth}|p{0.08\columnwidth}|p{0.09\columnwidth}|p{0.07\columnwidth}|p{0.12\columnwidth}|p{0.1\columnwidth}|p{0.09\columnwidth}|}
        \toprule
        Reference      & Group \newline Size & Member Specs  & Group \newline History & Relationships & Setting   & Task & Embodiment   & Role\\
        \hline \midrule
        Rosenberg-Kima et al \cite{Rosenberg-Kima2020Robot-supportedFacilitation} & 3-5 & None & Established & Team & Static & Creativity/ \newline Decision & Humanoid \newline (Nao) & Facilitator\\ 
        \hline
        Sadka, Jacobi et al \cite{Sadka2022ByOpening-encounters} & 2 & None & Random & Strangers & Casual & Co-existing & Actuated \newline Object (bar stool) & None\\ 
        \hline
        Sadka, Parush et al \cite{Sadka2023AllConflict} & 2 & None & Established & Couple & Static & Sharing & Actuated \newline Object (bar stool) & None\\ 
        \hline
        Scasselatti et al \cite{Scassellati2018ImprovingRobot} & 2 &  Inter-generational \newline Children with Autism \newline Care-taker/receiver & Established & Authority & Static & Intellective \newline Planning \newline Leisure & Tabletop \newline (Jibo) & Facilitator\\ 
        \hline
        Sebo et al \cite{Sebo2020TheBehavior, StrohkorbSebo2020} & 3 & None & Growing & Team & Static & Decision & Tabletop \newline (Jibo) & Group Member\\ 
        \hline
        Shamekhi et al \cite{Shamekhi2018} & 2 & None & Established & Team & Static & Decision & Mobile \newline (BEAM) & Facilitator\\ 
        \hline
        Shamekhi \& Bickmore \cite{Shamekhi2019} & 2 & None & Established & Team & Static & Decision & Tabletop \newline (Furhat) & Facilitator\\ 
        \hline
        Shen, Tennent et al \cite{Shen2017RobotConformity} & 3-5 & None & Established & Team & Static & Decision &  & \\ 
        \hline
        Shen, Slovak et al \cite{Shen2018} & 2 & Young Children & Established/ Forming & Peers/Friends/ Family & Dedicated Space & Leisure & Tabletop \newline (Keepon) & Facilitator\\ 
        \hline
        Shim et al \cite{Shim2017AnEvaluation, Arkin2014MoralCaregiving, Pettinati2015TowardsRelationships, Shim2015AnRelationships} & 2 & Inter-generational \newline Parkinson's Disease \newline Care-taker/receiver & Established & Authority & Static & Sharing & Humanoid \newline (Nao) & Facilitator\\ 
        \hline
        Shimoda et al \cite{Shimoda2022ApplicationRehabilitation} & 3-10 +1 & Novice Mediator \newline Aphasia \newline Care-taker/receiver & Established & Peers \newline Authority & Static & Learning &  & Facilitator\\ 
        \hline
        Shin et al \cite{Shin2021} & 2 & None & Forming & Strangers & Virtual & Leisure &  & Facilitator\\ 
        \hline
        Short \& Matari\'{c} \cite{Short2017} & 3 & None & Established & Team & Static & Performance & Tabletop \newline (SPRITE) & Group Member\\ 
        \hline
        Sinnema \& Alimardani \cite{Sinnema2019TheInteraction} & 2-4 & Older Adults & Forming & Peers & Dedicated Space & Intellective & Humanoid \newline (Nao) & Facilitator\\ 
        \bottomrule
\end{tabular}
\end{table}
\end{landscape}
\begin{landscape}
\begin{table}[!htbp]
    \caption{Overview of the input factors of all concepts included in this review (Part VI)}
    \label{tab:allpapers_inputs6}
    \centering
    \begin{tabular} {|p{0.1\columnwidth}|p{0.04\columnwidth}|p{0.15\columnwidth}|p{0.08\columnwidth}|p{0.09\columnwidth}|p{0.07\columnwidth}|p{0.12\columnwidth}|p{0.1\columnwidth}|p{0.09\columnwidth}|}
        \toprule
        Reference      & Group \newline Size & Member Specs  & Group \newline History & Relationships & Setting   & Task & Embodiment   & Role\\
        \hline \midrule
        Stoican et al \cite{Stoican2022LearningCollaboration} & 2 & None & Forming & Team & Dedicated Space & Performance & Mobile \newline (arm) & Group Member\\ 
        \hline
        Stoll et al \cite{Stoll2018KeepingFormat} & 2 & None & Established & Peers & Casual & Mixed Motive\ \newline Cognitive Conflict & Unspecific \newline (VEA) & Bystander\\ 
        \hline
        Strohkorb et al \cite{Strohkorb2016ImprovingRobot} & 2 & Children & Forming & Peers & Static & Intellective & Tabletop \newline (Keepon) & Group Member\\ 
        \hline
        Tahir et al \cite{Tahir2020AConversations, Tahir2014PerceptionDialogs} & 2 & None & Random & Strangers & Static & Leisure & Humanoid \newline (Nao) & Facilitator\\ 
        \hline
        Takano et al \cite{Takano2009PsychologicalNeeds} & 2 & Doctor/Patient & Random & Authority & Static & Sharing & Humanoid \newline (ReplieeQ2) & Bystander\\ 
        \hline
        Takeuchi \& You \cite{Takeuchi2014Whirlstools:Affordance} & 2-5 & None & Random & Strangers & Public & Co-existing & Actuated \newline Object (public seating) & None\\ 
        \hline
        Tennent et al \cite{Tennent2019} & 3-5 & None & Established & Team & Static & Creativity \newline Decision & Actuated \newline Object (microphone) & Facilitator\\ 
        \hline
        Traeger et al \cite{Traeger2020, StrohkorbSebo2018TheTeams} & 3 & None & Established & Team & Static & Planning & Humanoid \newline (Nao) & Group Member\\ 
        \hline
        Uchida et al \cite{Uchida2020ImprovingExperience} & 5-10 & None & Forming & Peers & Static & Leisure/ \newline Sharing & Humanoid \newline (Pepper) & Facilitator\\ 
        \hline
        Utami \& Bickmore \cite{Utami2019CollaborativeRobot, Utami2017} & 2 & None & Established & Couple & Static & Sharing & Tabletop \newline (Furhat) & Facilitator\\ 
        \hline
        Wang et al \cite{Wang2023ExploringConflict} & 2 & None & Established & Mixed & Virtual & Leisure &  & Facilitator\\ 
        \hline
        Xu et al \cite{Xu2014} & 2 & None & Forming & Strangers & Public & Co-existing & Unspecific \newline (VEA) & Facilitator\\ 
        \hline
        Yamazaki et al \cite{Yamazaki2012AVisitor} & 2-10 & Sub-groups & Forming & Mixed & Public & Learning & Humanoid \newline (custom) & Facilitator\\ 
        \hline
        Zhang et al \cite{Zhang2023} & 2 & None & Random & Strangers & Casual & Co-existing & Humanoid \newline (Nao) & Facilitator\\ 
        \hline
        Zheng et al \cite{Zheng2005DesigningGuide} & 3-10 \newline varying & None & Forming & Strangers & Virtual & Leisure & Unspecific \newline (VAE) & Facilitator\\ 
        \bottomrule
\end{tabular}
\end{table}
\end{landscape}

\begin{landscape}
\begin{table}[!htbp]
    \caption{Overview of the mediation process and output factors of all concepts included in this review (Part I)}
    \label{tab:allpapers_dynamics}
    \centering
    \begin{tabular} {|p{0.1\columnwidth}|p{0.21\columnwidth}|p{0.08\columnwidth}|p{0.11\columnwidth}|p{0.06\columnwidth}|p{0.12\columnwidth}|p{0.14\columnwidth}|}
        \toprule
        Reference     & Intervention Type & Intervention \newline Modality & Intervention \newline Trigger & Mediation \newline Style & Target \newline Outcome & Target \newline Proxy Measure  \\
        \hline \midrule
        Ablett et al \cite{Ablett2007BuildBot:Teams} & Support problem identification \newline Evaluate performance & Actions \newline Utterances & Task State & Explicit & Enjoyment \newline Social Status & Engagement \\ 
        \hline
        Al Moubayed et al \cite{AlMoubayed2013TutoringTutor} & Structure activities \newline Participation management \newline Facilitate decision making \newline Evaluate performance \newline Provide information & Speech & Task State & Explicit & Solution Quality & Participation Balance\\ 
        \hline
        Alves-Oliveira et al \cite{Alves-Oliveira2019EmpathicStudy} & Structure activities \newline Summarize state \newline Participation management \newline Provide information \newline Keeping focus & Speech & Engagement & Explicit & Knowledge & Interaction Qualities\\ 
        \hline
        Aylett et al \cite{Aylett2023} & Promote mutual understanding \newline Structure interactions & Speech & Task State & Explicit & Individual Skill & Interaction Quantity\\ 
        \hline
        Baghaei Ravari et al \cite{BaghaeiRavari2021EffectsLearning} & Participation management \newline Manage interpersonal processes & Speech & Conversation Flow & Implicit & Individual Skill \newline Enjoyment & Interaction Quantity \newline Participation Balance \newline Engagement\\ 
        \hline
        Basu et al \cite{Basu2001TowardsSettings} & Participation Management & Speech \newline Other & Participation Statistics & Implicit & & Participation Balance \\ 
        \hline
        Birmingham et al \cite{Birmingham2020} & Structure activities \newline Promote mutual understanding \newline Show own state \newline Create open atmosphere \newline Participation management & Speech \newline Gaze & Task State & Explicit & Satisfaction \newline Trust \newline Cohesion \newline Knowledge & \\ 
        \hline
        Buchem \cite{Buchem2023Scaling-UpNAO} & Structure activities \newline Evaluate performance \newline Facilitate decision making & Speech & Pre-Structure & Explicit & Individual Skill \newline Satisfaction \newline Enjoyment \newline Motivation & \\ 
        \hline
        Chandra et al \cite{Chandra2016ChildrensRobot, Chandra2015CanActivity} & Assign roles \newline Support problem identification \newline Structure activities \newline Solicit feedback \newline Welcoming/Closing & Speech \newline Gaze & Task State & Explicit & Individual Skill & Interaction Quantity \\ 
        \bottomrule
    \end{tabular}
\end{table}
\end{landscape}
\begin{landscape}
\begin{table}[!htbp]
    \caption{Overview of the mediation process and output factors of all concepts included in this review (Part II)}
    \label{tab:allpapers_dynamics2}
    \centering
    \begin{tabular} {|p{0.1\columnwidth}|p{0.21\columnwidth}|p{0.08\columnwidth}|p{0.11\columnwidth}|p{0.06\columnwidth}|p{0.12\columnwidth}|p{0.14\columnwidth}|}
        \toprule
        Reference     & Intervention Type & Intervention \newline Modality & Intervention \newline Trigger & Mediation \newline Style & Target \newline Outcome & Target \newline Proxy Measure  \\
        \hline \midrule
        Charisi et al \cite{Charisi2021TheDynamics} & Provide information \newline Manage interpersonal processes & Speech & Task State \newline Participation Statistics \newline Explicit Request & Explicit & Solution Quality & Interaction Quantity \\ 
        \hline
        Chen et al \cite{Chen2022DesigningInterviews} & Manage interpersonal processes \newline Promote mediating actions \newline Encourage idea generation \newline Provide new task & Speech & Conversation Flow & Explicit & Satisfaction \newline Enjoyment \newline Individual Skill \newline Relationships & Interaction Quantity \newline Interaction Qualities \newline Engagement \newline Affect Display Quality\\ 
        \hline
        Chew \& Nakamura \cite{Chew2023WhoInteractions} & Support problem identification \newline Provide information \newline Encourage idea generation & Speech & & Explicit & & Participation Balance \newline Engagement \\ 
        \hline
        Cumbal et al \cite{Cumbal2022ShapingBackchannels, Parreira2022, Gillet2022LearningInteractions, Gillet2021} & Active presence \newline Participation management & Speech \newline Gaze \newline Utterances & Participation Statistics & Implicit & & Participation Balance \\ 
        \hline
        de Rooj et al \cite{deRooij2023Co-DesigningDynamics} & Structure activities \newline Show own state & Speech \newline Gaze \newline Gesture & Task State & Explicit & & Perceived Cooperation \newline Perceived Conflicts \newline Perceived Contribution\\ 
        \hline
        Dietrich \& Weisswange \cite{Dietrich2022} & Handle misunderstandings \newline Promote mutual understanding \newline Transfer messages & Speech & Common Interest & Explicit & Mutual Understanding & \# Helping \newline Conversation Flow\\ 
        \hline
        D\"{o}rrenb\"{a}cher et al \cite{Dorrenbacher2023} & & & & & & \\
        PARFAR & Manage interpersonal processes \newline Manage conflict & Speech & Conflict & Explicit & Individual Skill &\\ 
        SEYNO & Manage interpersonal processes \newline Manage conflicts & Speech & Conflict & Explicit & & \# Conflicts\\ 
        SEATY & Manage interpersonal processes & Action & Participant Intention & Explicit & & \# Helping\\
        \hline
        Druckman et al \cite{Druckman2021WhoHumans} & Manage conflict \newline Structure activities \newline Provide information & Speech & Pre-Structure & Explicit & Solution Quality \newline Satisfaction & Consensus \newline Mediation Satisfaction\\ 
        \bottomrule
    \end{tabular}
\end{table}
\end{landscape}
\begin{landscape}
\begin{table}[!htbp]
    \caption{Overview of the mediation process and output factors of all concepts included in this review (Part III)}
    \label{tab:allpapers_dynamics3}
    \centering
    \begin{tabular} {|p{0.1\columnwidth}|p{0.21\columnwidth}|p{0.08\columnwidth}|p{0.11\columnwidth}|p{0.06\columnwidth}|p{0.12\columnwidth}|p{0.14\columnwidth}|}
        \toprule
        Reference     & Intervention Type & Intervention \newline Modality & Intervention \newline Trigger & Mediation \newline Style & Target \newline Outcome & Target \newline Proxy Measure  \\
        \hline \midrule             
        Edwards et al \cite{Edwards2018ALearning} & Moderation \newline Encourage idea generation \newline Promote mutual understanding & Speech \newline Action & Conversation Flow & Explicit & Individual Skill \newline Motivation & Engagement \newline Mediation Satisfaction\\ 
        \hline
        Erel et al \cite{Erel2021, Rifinski2021Human-human-robotInteraction} & Active presence \newline Create open atmosphere & Gesture & Pre-Structure & Un-directed & Well-being \newline Satisfaction & Interaction Quality\\ 
        \hline
        Fan et al \cite{Fan2021FieldAdults, Fan2022SAR-Connect:Adults, Fan2017AModels, Fan2016AImpaired} & Welcoming/Closing \newline Provide information \newline Structure interactions & Speech \newline Gaze \newline Gesture & Performance & Explicit & & Interaction Quantity \newline Engagement\\ 
        \hline
        Fu, Carolyn et al  \cite{Fu2017TurnTurn-Taking} & Participation Management & Other & Participation Statistics & Implicit & & Participation Balance\\ 
        \hline
        Fu, Changzeng et al  \cite{Fu2021UsingConversations} & Encourage group identity \newline Promote mutual understanding & Speech & Common Interests & Explicit & Cohesion \newline Relationships &\\ 
        \hline
        Fu, Di et al \cite{Fu2023TheCollaboration} & Show own state \newline Structure activities \newline Motivating & Gaze \newline Facial Expression & Pre-Structure & Accidental & Solution Speed & \\ 
        \hline
        Fucinato et al \cite{Fucinato2023CharismaticCreativity} & Structure activities \newline Summarize state \newline Evaluate performance \newline Encourage idea generation & Speech & Pre-Structure & Un-directed & Solution Quality \newline Solution Quantity \newline Satisfaction \newline Psychological safety &\\ 
        \hline
        Fukuda et al \cite{Fukuda2016AnalysisMediator} & Provide new task & Speech & Pre-Structure & Explicit & & Interaction Quantity \\ 
        \hline
        Gillet, van den Bos et al \cite{Gillet2020, Gillet2020AChildren} & Structure interactions \newline Provide information \newline Active presence & Action \newline Gesture & Participation Statistics & Explicit & Cohesion & Participation Balance \\ 
        \hline
        Gillet, Winkle et al \cite{Gillet2022Ice-BreakersTeenagers, Gillet2022AInteractions} & Create open atmosphere \newline Participation management \newline Provide information \newline Structure interactions \newline Support multiple perspectives \newline Keeping focus & Speech \newline Gesture \newline Gaze & Explicit & & Enjoyment/ \newline Satisfaction & Participation Balance \\ 
        \bottomrule
    \end{tabular}
\end{table}
\end{landscape}
\begin{landscape}
\begin{table}[!htbp]
    \caption{Overview of the mediation process and output factors of all concepts included in this review (Part IV)}
    \label{tab:allpapers_dynamics4}
    \centering
    \begin{tabular} {|p{0.1\columnwidth}|p{0.21\columnwidth}|p{0.08\columnwidth}|p{0.11\columnwidth}|p{0.06\columnwidth}|p{0.12\columnwidth}|p{0.14\columnwidth}|}
        \toprule
        Reference     & Intervention Type & Intervention \newline Modality & Intervention \newline Trigger & Mediation \newline Style & Target \newline Outcome & Target \newline Proxy Measure  \\
        \hline \midrule
        Gordon et al \cite{Gordon2022InvestigationAgent, Mizrahi2022Long-TermDiscussions., Mizrahi2022VRobotator:K-12} & Participation management \newline Encourage idea generation \newline Manage conflict \newline Create open atmosphere \newline Evaluate performance & Speech & Conversation Flow \newline Participation Statistics \newline Contribution Semantics \newline Conflict & Explicit & Individual Skill \newline Team Confidence \newline Cohesion & Engagement \newline Affect Display Quality \newline Interaction Quantity \newline Participation Balance \newline Perceived Usefulness\\ 
        \hline
        Hasegawa \& Nakauchi \cite{Hasegawa2014FacilitationExaggeration, Hasegawa2014TelepresenceTeleconferences} & Participation management & Gesture \newline Gaze & Participant Intention & Implicit & &\\ 
        \hline
        Hayamizu et al \cite{Hayamizu2013AnControl, Sano2009AControl} & Manage relationships \newline Structure interactions \newline Active presence \newline Spark conversational topic & Speech \newline Gaze \newline Gesture & Conversation Flow & Explicit \newline Implicit & Satisfaction \newline Relationships &Conversation Flow \\ 
        \hline
        Hitron et al \cite{Hitron2022AIRobots} & Participation management \newline Evaluate performance & Action & Participation Statistics & Explicit & Satisfaction & Perceived Fairness \newline Mediation Satisfaction\\ 
        \hline
        Hoffman et al \cite{Hoffman2015, Zuckerman2015EmpathyCompanions} & Active presence \newline Mirror atmosphere & Gesture & Conversation Flow/ \newline Conflict & Un-directed & Relationships & Interaction Quality \newline Interaction Quantity\\ 
        \hline
        Ikari et al \cite{Ikari2020Multiple-RobotDiscussion} & Facilitate decision making \newline Support multiple perspectives & Speech & Pre-Structure & Explicit & Satisfaction \newline Cohesion \newline Team Confidence & Perceived Contribution\\ 
        \hline
        Inoue et al \cite{Inoue2021AParticipants} & Promote mutual understanding \newline Participation management & Speech & Participation Statistics \newline Contribution Semantics & Explicit & & Perceived Contribution \newline Conversation Flow\\ 
        \hline
        Isbister et al \cite{Isbister2000HelperSpace} & Spark conversational topic \newline Structure Interactions & Speech & Conversation Flow & Explicit & Relationships \newline Satisfaction &\\ 
        \hline
        Jeong et al \cite{Jeong2018} & Promote mutual understanding \newline Team building & Speech \newline Other & Participant State & Un-directed & Relationships \newline Social Status & Interaction Quantity\\ 
        \hline
        Jung, Martelaro, Hinds \cite{Jung2015UsingViolations} & Manage interpersonal processes \newline Manage conflicts & Speech & Conflict & Explicit & Solution Quality \newline Cohesion & Quality of Affect Display Quality\\ 
        \bottomrule
    \end{tabular}
\end{table}
\end{landscape}
\begin{landscape}
\begin{table}[!htbp]
    \caption{Overview of the mediation process and output factors of all concepts included in this review (Part V)}
    \label{tab:allpapers_dynamics5}
    \centering
    \begin{tabular} {|p{0.1\columnwidth}|p{0.21\columnwidth}|p{0.08\columnwidth}|p{0.11\columnwidth}|p{0.06\columnwidth}|p{0.12\columnwidth}|p{0.14\columnwidth}|}
        \toprule
        Reference     & Intervention Type & Intervention \newline Modality & Intervention \newline Trigger & Mediation \newline Style & Target \newline Outcome & Target \newline Proxy Measure  \\
        \hline \midrule
        Jung, DiFranzo et al \cite{Jung2020Robot-assistedGroups} & Provide information & Action & Pre-Structure & Accidental & Solution Quality \newline Satisfaction \newline Cohesion &\\ 
        \hline
        Kanda et al \cite{Kanda2012ChildrenRobot} & Manage conflict \newline Manage interpersonal processes \newline Evaluate performance \newline Structure activities \newline Structure interactions \newline Motivating & Speech \newline Action & Conflict & Explicit & Individual Skill &\\ 
        \hline
        Kang et al \cite{Kang2023TheGroup} & Manage interpersonal processes & Action & Participant Intention & Implicit & Solution Quality &\\ 
        \hline
        Karatas et al \cite{Karatas2020UtilizationArt} & Structure interactions & Gaze & Task State & Implicit & & Engagement \\ 
        \hline
        Kim et al \cite{Kim2023ChildrobotBehaviors, Kim2018DesigningOpportunities} & Structure interactions \newline Evaluate performance \newline Promote mutual understanding &  Speech \newline Gesture \newline Action & Pre-Structure & Explicit & Relationships & Coordination \newline Affect Display Quantity \newline Interaction Quality \newline Consensus \newline \# Helping\\ 
        \hline
        Kobuki et al \cite{Kobuki2023RoboticStudy, Seaborn2023VoiceConversation} & Active presence \newline Structure activities & Utterances & Pre-Structure & Explicit \newline Un-directed & Enjoyment \newline Individual Skill &\\ 
        \hline
        Kochigami et al \cite{Kochigami2018DoesCommunication} & Manage relationships & Speech & Group Structure Change & Explicit & Mood & Interaction Quantity \\ 
        \hline
        Leite et al \cite{Leite2016AutonomousInteraction} & Keeping focus \newline Show own state & Speech \newline Gaze & Engagement & Explicit & Knowledge & Engagement \newline Affect Display Quantity \newline Affect Display Quality\\ 
        \hline
        Li et al \cite{Li2023ImprovingGroups} & Participation Management & Speech \newline Gesture & Participation Statistics \newline Explicit Request & Explicit & & Perceived Cooperation \newline Participation Balance\\ 
        \bottomrule
    \end{tabular}
\end{table}
\end{landscape}
\begin{landscape}
\begin{table}[!htbp]
    \caption{Overview of the mediation process and output factors of all concepts included in this review (Part VI)}
    \label{tab:allpapers_dynamics6}
    \centering
    \begin{tabular} {|p{0.1\columnwidth}|p{0.23\columnwidth}|p{0.08\columnwidth}|p{0.12\columnwidth}|p{0.06\columnwidth}|p{0.12\columnwidth}|p{0.12\columnwidth}|}
        \toprule
        Reference     & Intervention Type & Intervention \newline Modality & Intervention \newline Trigger & Mediation \newline Style & Target Outcome & Target Proxy \newline Measure  \\
        \hline \midrule
        Lopes et al \cite{Lopes2021SocialTeams} & Motivating \newline Evaluate performance & Speech & Task State & Explicit & Solution Quality \newline Satisfaction & Engagement \newline Role Ambiguity\\ 
        \hline
        Matsuyama et al \cite{Matsuyama2010PsychologicalGame, Fujie2009ConversationCommunication, Matsuyama2008DesigningCommunication} &  Participation management \newline Inclusion \newline Spark conversational topic & Speech & Participation Statistics/ \newline Conversation Flow \newline Participant Intention & Implicit & Entitativity & Engagement \newline Participation Balance\\ 
        \hline
        Matsuyama et al \cite{Matsuyama2015Four-participantParticipant, Matsuyama2015TowardsFacilitation} & Create open atmosphere \newline Provide information \newline Asking for the group & Speech & Engagement & Implicit & & Affect Display Quality\\ 
        \hline
        Mitnik et al \cite{Mitnik2008AnMediator} & Assign roles \newline Structure activities \newline Evaluate performance & Speech \newline Action & Task State & Explicit & Motivation \newline Knowledge & Interaction Quantity\\ 
        \hline
        Moharana et al \cite{Moharana2019RobotsCaregivers} & Manage interpersonal processes \newline Provide new task & Other & Pre-Structure & Explicit & Enjoyment \newline Mood &\\ 
        \hline
        Nagao \& Takeuchi \cite{Nagao1994SocialAgents} & Handle misunderstandings & Speech & Participant Intention & Explicit & Mutual Understanding & Conversation Flow\\ 
        \hline
        Nakanishi et al \cite{Nakanishi2003CanCommunities} & Promote Mutual Understanding & Speech & Pre-Structure & Implicit & Relationships &\\ 
        \hline
        Neto et al \cite{Neto2023TheChildren, Neto2021FosteringRobots} & Active presence \newline Participation management & Speech \newline Gaze \newline Action \newline Other & Participation Statistics & Implicit & Solution Speed \newline In-group Identification & Participation Balance \newline Contribution Balance \newline Socialness of behavior \newline Engagement\\ 
        \hline
        Noguchi et al \cite{Noguchi2023HowRecipients} & Transfer messages & Speech & Explicit Request & Implicit & Relationships \newline Mutual Understanding &\\ 
        \hline
        Ohshima et al \cite{Ohshima2017Neut:Conversations} & Kick-start \newline Participation management & Speech \newline Gaze & Conversation Flow & Explicit & & Conversation Flow\\ 
        \hline
        Ono et al \cite{Ono1999} & Manage relationships & Action & Common Interests & Implicit & Relationships & Interaction Quantity\\ 
        \bottomrule
\end{tabular}
\end{table}
\end{landscape}
\begin{landscape}
\begin{table}[!htbp]
    \caption{Overview of the mediation process and output factors of all concepts included in this review (Part VII)}
    \label{tab:allpapers_dynamics7}
    \centering
    \begin{tabular} {|p{0.1\columnwidth}|p{0.23\columnwidth}|p{0.08\columnwidth}|p{0.12\columnwidth}|p{0.06\columnwidth}|p{0.12\columnwidth}|p{0.12\columnwidth}|}
        \toprule
        Reference     & Intervention Type & Intervention \newline Modality & Intervention \newline Trigger & Mediation \newline Style & Target Outcome & Target Proxy \newline Measure  \\
        \hline \midrule
        Pliasa \& Fachantidis \cite{Pliasa2019CanDevelopment} & Provide new task \newline Structure activities \newline Promote mutual understanding  & Speech & Pre-Structure & Explicit & Individual Skill & Engagement\\ 
        \hline
        Rosenberg-Kima et al \cite{Rosenberg-Kima2020Robot-supportedFacilitation} & Structure activities \newline Structure interactions & Speech \newline Other & Task State & Explicit & Satisfaction &\\ 
        \hline
        Sadka, Jacobi et al \cite{Sadka2022ByOpening-encounters} & Enable interactions & Action & Pre-Structure & Implicit & & Interaction Quantity\\ 
        \hline
        Sadka, Parush et al \cite{Sadka2023AllConflict} & Enable interactions & Action & Pre-Structure & Implicit & & Interaction Quantity \newline Interaction Quality \newline Perceived Conflicts\\ 
        \hline
        Scasselatti et al \cite{Scassellati2018ImprovingRobot} & Manage interpersonal processes \newline Moderation \newline Provide new task \newline Active presence & Gaze & Individual Skill & Implicit & Individual Skill & Interaction Quantity \newline Interaction Quality \newline Engagement\\ 
        \hline
        Sebo et al \cite{Sebo2020TheBehavior, StrohkorbSebo2020} & Active presence \newline Provide information & Speech & Task State \newline Explicit Request & Implicit & Team Climate \newline In-group Identification & Participation Balance\\ 
        \hline
        Shamekhi et al \cite{Shamekhi2018} & Structure activities \newline Facilitate decision making \newline Create open atmosphere \newline Active presence & Speech & Task State & Explicit & Satisfaction \newline Cohesion & Consensus\\ 
        \hline
        Shamekhi \& Bickmore \cite{Shamekhi2019} & Moderation \newline Structure activities \newline Support problem identification \newline Participation management \newline Active presence \newline Evaluate performance \newline Create open atmosphere & Speech & Task State & Explicit & Satisfaction \newline Solution Speed \newline Cohesion \newline Team Confidence & Participation Balance \newline Perceived Conflicts \newline Mediation Satisfaction \newline Adherence to Structure\\ 
        \bottomrule
\end{tabular}
\end{table}
\end{landscape}
\begin{landscape}
\begin{table}[!htbp]
    \caption{Overview of the mediation process and output factors of all concepts included in this review (Part VIII)}
    \label{tab:allpapers_dynamics8}
    \centering
    \begin{tabular} {|p{0.1\columnwidth}|p{0.23\columnwidth}|p{0.08\columnwidth}|p{0.12\columnwidth}|p{0.06\columnwidth}|p{0.12\columnwidth}|p{0.12\columnwidth}|}
        \toprule
        Reference     & Intervention Type & Intervention \newline Modality & Intervention \newline Trigger & Mediation \newline Style & Target Outcome & Target Proxy \newline Measure  \\
        \hline \midrule
        Shen, Tennent et al \cite{Shen2017RobotConformity} & Active presence & Facial Expression & Un-directed & & & Consensus\\ 
        \hline
        Shen, Slovak et al \cite{Shen2018} & Manage interpersonal processes \newline Support problem identification \newline Manage conflict \newline Promote mediating actions & Speech & Conflict & Explicit & Individual Skill & Constructiveness of Behavior \newline Socialness of behavior\\ 
        \hline
        Shim et al \cite{Shim2017AnEvaluation, Arkin2014MoralCaregiving, Pettinati2015TowardsRelationships, Shim2015AnRelationships} & Structure activities \newline Promote mutual understanding \newline Manage interpersonal processes \newline Handle misunderstandings & Speech & Conflict & Explicit & Relationships  \newline Affective State & \# Conflicts\\ 
        \hline
        Shimoda et al \cite{Shimoda2022ApplicationRehabilitation} & Structure activities \newline Create open atmosphere & Speech & Explicit Request & Explicit & Individual Skill & Engagement\\ 
        \hline
        Shin et al \cite{Shin2021} & Encourage group identity \newline Manage relationships \newline Spark conversational topic & Speech & Pre-Structure & Explicit & Relationships & Conversation Flow \newline Interaction Quality\\ 
        \hline
        Short \& Matari\'{c} \cite{Short2017} & Provide information & Speech \newline Action & Participation Statistics & Implicit & Solution Quality \newline Cohesion & Contribution Balance\\ 
        \hline
        Sinnema \& Alimardani \cite{Sinnema2019TheInteraction} & Manage interpersonal processes & Speech & Pre-Structure & Explicit & Social Status \newline Mood \newline Cohesion & Interaction Quantity\\ 
        \hline
        Stoican et al \cite{Stoican2022LearningCollaboration} & Provide information & Action & Individual Skill/ \newline Performance & Implicit & Well-being \newline Trust &\\ 
        \hline
        Stoll et al \cite{Stoll2018KeepingFormat} & Create open atmosphere & Speech & Conflict & Implicit & & Intensity of Conflicts\\ 
        \hline
        Strohkorb et al \cite{Strohkorb2016ImprovingRobot} & Promote mutual understanding \newline Provide information \newline Solicit feedback & Speech \newline Gaze & Pre-Structure & Implicit & Solution Quality \newline Cohesion \newline Team Confidence & Task Cohesion\\ 
        \hline
        Tahir et al \cite{Tahir2020AConversations, Tahir2014PerceptionDialogs} & Manage interpersonal processes & Speech \newline Gesture & Conversation Flow & Explicit & & Conversation Flow\\ 
        \hline
        Takano et al \cite{Takano2009PsychologicalNeeds} & Active presence \newline Mirror atmosphere & Facial Expression & & Un-directed & Satisfaction \newline Mood \newline Mutual Understanding &\\ 
        \bottomrule
\end{tabular}
\end{table}
\end{landscape}
\begin{landscape}
\begin{table}[!htbp]
    \caption{Overview of the mediation process and output factors of all concepts included in this review (Part IX)}
    \label{tab:allpapers_dynamics9}
    \centering
    \begin{tabular} {|p{0.1\columnwidth}|p{0.23\columnwidth}|p{0.08\columnwidth}|p{0.12\columnwidth}|p{0.06\columnwidth}|p{0.12\columnwidth}|p{0.12\columnwidth}|}
        \toprule
        Reference     & Intervention Type & Intervention \newline Modality & Intervention \newline Trigger & Mediation \newline Style & Target Outcome & Target Proxy \newline Measure  \\
        \hline \midrule
        Takeuchi \& You \cite{Takeuchi2014Whirlstools:Affordance} & Enable interactions & Action & Implicit & & & Interaction Quantity \newline Interaction Quality\\ 
        \hline
        Tennent et al \cite{Tennent2019} & Participation management & Gaze \newline Action & Participation Statistics & Implicit & Solution Quality & Participation Balance\\ 
        \hline
        Traeger et al \cite{Traeger2020, StrohkorbSebo2018TheTeams} & Show own state \newline Evaluate performance \newline Create open atmosphere \newline Provide information & Speech & Pre-Structure & Un-directed & Satisfaction \newline Team Climate & Interaction Quality \newline Participation Balance\\ 
        \hline
        Uchida et al \cite{Uchida2020ImprovingExperience} & Manage relationships \newline Encourage group identity \newline Promote mutual understanding & Speech & Common Interests & Explicit & Cohesion &\\ 
        \hline
        Utami \& Bickmore \cite{Utami2019CollaborativeRobot, Utami2017} & Active presence \newline Explain tools \newline Evaluate performance & Speech \newline Gaze & Pre-Structure & Explicit & Relationships \newline Enjoyment & Coordination \newline Affect Display Quantity\\ 
        \hline
        Wang et al \cite{Wang2023ExploringConflict} & Mirror atmosphere \newline Aid in emotion support & Social Touch & Conflict & Implicit & Mood & Intensity of Conflicts\\ 
        \hline
        Xu et al \cite{Xu2014} & Manage relationships \newline Create open atmosphere & Speech & Common Interests & Explicit & & Conversation Flow \\ 
        \hline
        Yamazaki et al \cite{Yamazaki2012AVisitor} & Provide new task \newline Participation management & Speech \newline Gaze \newline Gesture & Task State & Explicit & Enjoyment & Interaction Quantity\\ 
        \hline
        Zhang et al \cite{Zhang2023} & Spark conversational topic \newline Promote mutual understanding & Speech & Pre-Structure & Explicit & Well-being \newline Satisfaction \newline Mood \newline Relationships & Interaction Quantity \newline Conversation Flow\\ 
        \hline
        Zheng et al \cite{Zheng2005DesigningGuide} & Participation management \newline Welcoming/Closing \newline Moderation \newline Encourage idea generation & Speech & Group Structure Changes & Explicit & & Participation Balance\\ 
        \bottomrule
\end{tabular}
\end{table}
\end{landscape}

\begin{landscape}
\begin{table}[!htbp]
    \caption{Overview of the experiment parameters of all concepts included in this review (Part I). The Measures column reports if results were produced through a questionnaire ("Q"), video/audio coding of the activity ("C"), testing participants before/after the experiment ("T"), qualitative interviews ("I"), an objective measure during the activity ("O") or complex measures produced by machine learning algorithms ("A"). For the first three types, bold font marks a validated measure. Measures showing significant effects are green, mixed or partial positive effects are cyan.}
    \label{tab:allpapers_experiments}
    \centering
    \begin{tabular} {|p{0.1\columnwidth}|p{0.11\columnwidth}|p{0.1\columnwidth}|p{0.08\columnwidth}|p{0.11\columnwidth}|p{0.18\columnwidth}|p{0.08\columnwidth}|p{0.12\columnwidth}|}
        \toprule
        Reference     & Experiment Type & Robot Control & \# Groups \newline (Participants) & Experiment \newline Conditions & Measures & Participant \newline Specs & Relation/ \newline Reputation \\
        \hline \midrule
        Ablett et al \cite{Ablett2007BuildBot:Teams} & Proof of Concept & Autonomous & 2 (14) & No Embodiment \newline \textit{Intra-subject} & Custom (Q) & Student Peer Groups & None \\ 
        \hline
        Al Moubayed et al \cite{AlMoubayed2013TutoringTutor} & Proof of Concept &  &  &  & Tutor Assessment Questionnaire (\textbf{Q}, based on \cite{Laugwitz2008ConstructionQuestionnaire}) &  & None/\newline Task Expert \\ 
        \hline
        Alves-Oliveira et al \cite{Alves-Oliveira2019EmpathicStudy} & User study (Lab) & Autonomous & 28 (63) \newline 9 (18) & No Mediator \newline Ablated Robot \newline \textit{Inter-subject} & Factual Knowledge (\textbf{T}) \newline Multiple Perspectives (\textbf{T}) \newline Personal Values (\textcolor{cyan}{C}) & Teenagers \newline Peers & Neutral/\newline Task Expert\\  
        \hline
        Aylett et al \cite{Aylett2023} & Design &  &  &  &  &  & None \\
        \hline
        Baghaei Ravari et al \cite{BaghaeiRavari2021EffectsLearning} & User study (Lab) & Autonomous & 34 (68) & Ablated Robot \newline \textit{Inter-subject} & Talking Time (\textcolor{lime}{O}) \newline Participation Balance (O) \newline Task Engagement (\textcolor{lime}{O}) \newline Knowledge (T) \newline Intrinsic Motivation Inventory (\textbf{Q}, \cite{Deci1994FacilitatingPerspective}) \newline Situational Motivation Scale (\textbf{Q}, \cite{Guay2000OnSIMS}) \newline Enjoyment (Q) \newline Pick-a-Mood (\textcolor{cyan}{\textbf{Q}}, \cite{Desmet2016Mood})\newline Mediator Quality (Q) & Adults \newline Strangers & \\
        \hline
        Basu et al \cite{Basu2001TowardsSettings} & Design &  &  &  & Influence Graphs (A) &  & \\ 
        \hline
        Birmingham et al \cite{Birmingham2020} & User study (Lab) & Wizard & 27 (81) & None & Dyadic Trust Scale (\textcolor{lime}{\textbf{Q}}, \cite{Larzelere1980TheRelationships}) \newline Speciﬁc Interpersonal Trust Scale (\textcolor{lime}{\textbf{Q}}, \cite{Johnson-George1982MeasurementOther}) \newline Trust Antecedents (Q) & Students \newline Strangers \newline Stressed & Group-aligned/\newline None\\
        \hline
        Buchem \cite{Buchem2023Scaling-UpNAO} & User study (Lab) & Autonomous & 10 (46) & None & Shortened Learning Experience Scale (\textbf{Q}, based on \cite{Fokides2018DevelopmentResults.}) & Students \newline Peers & Neutral/\newline Mediation Expert\\
        \hline
        Chandra et al \cite{Chandra2016ChildrensRobot, Chandra2015CanActivity} & User study (Lab) & Wizard & 20 (40) & Intervention Type \newline \textit{Inter-subject} & Verbal Behavior (\textbf{C}) \newline Writing Skill (T) & Children \newline Peers & None \\
        \bottomrule
    \end{tabular}
\end{table}
\end{landscape}
\begin{landscape}
\begin{table}[!htbp]
    \caption{Overview of the experiment parameters of all concepts included in this review (Part II). The Measures column reports if results were produced through a questionnaire ("Q"), video/audio coding of the activity ("C"), testing participants before/after the experiment ("T"), qualitative interviews ("I"), an objective measure during/after the activity ("O") or complex measures produced by machine learning algorithms ("A"). For the first three types, bold font marks a validated measure. Measures showing significant effects are green, mixed or partial positive effects are cyan.}
    \label{tab:allpapers_experiments2}
    \centering
    \begin{tabular} {|p{0.1\columnwidth}|p{0.11\columnwidth}|p{0.1\columnwidth}|p{0.08\columnwidth}|p{0.11\columnwidth}|p{0.18\columnwidth}|p{0.08\columnwidth}|p{0.12\columnwidth}|}
        \toprule
        Reference     & Experiment Type & Robot Control & \# Groups \newline (Participants) & Experiment \newline Conditions & Measures & Participant \newline Specs & Relation/ \newline Reputation \\
        \hline \midrule
        Charisi et al \cite{Charisi2021TheDynamics} & User study (Lab) & Autonomous & 42 (84) & Ablated Robot \newline \textit{Inter-subject} & Task Performance (\textcolor{lime}{O}) \newline Social Interaction Count (C) \newline Participation Balance (C) & Children \newline Strangers & Group-aligned/\newline Task Expert\\
        \hline
        Chen et al \cite{Chen2022DesigningInterviews} & Field Study & Wizard & 12 (24) & No Mediator \newline Ablated Robot \newline \textit{Intra-subject} & Experience (I) & Parent \& \newline Child & None \\ 
        \hline
        Chew \& Nakamura \cite{Chew2023WhoInteractions} & Design &  &  &  & Participation Balance (O) &  & None \\
        \hline
        Cumbal et al \cite{Cumbal2022ShapingBackchannels, Parreira2022, Gillet2022LearningInteractions, Gillet2021} & User study (Lab) & Autonomous & 20 (40) & Ablated Robot \newline \textit{Inter-subject} & Participation Balance (O) & Adults \newline Strangers \newline Language Skill Difference & \\ 
        \hline
        de Rooj et al \cite{deRooij2023Co-DesigningDynamics} & User study (Lab) & Wizard & 29 (110) & Intervention Type \newline \textit{Inter-subject} & Perceived Cooperation, Conflict \& Contribution (Q) & Students \newline Strangers & None\\
        \hline
        Dietrich \& Weisswange \cite{Dietrich2022} & Design &  &  &  &  &  & Individual-aligned/\newline None\\
        \hline
        D\"{o}rrenb\"{a}cher et al \cite{Dorrenbacher2023} &  &  &  &  &  &  & \\
        PARFAR &  &  &  &  &  &  & Group-aligned/\newline Mediation Expert\\
        SEYNO &  &  &  &  &  &  & Other-aligned/\newline Mediation Expert\\
        SEATY &  &  &  &  &  &  & Other-aligned/\newline Mediation Expert\\
        \hline
        Druckman et al \cite{Druckman2021WhoHumans} & User study (Lab) & Wizard & 143 (286) & No Mediator \newline Human Mediator \newline No Embodiment \newline \textit{Mixed} & Task Success (O) \newline Solution Quality (O)  \newline Perceived Mediation Quality (Q) & Students \newline Strangers & Neutral/\newline Mediation Expert\\
        \bottomrule
    \end{tabular}
\end{table}
\end{landscape}
\begin{landscape}
\begin{table}[!htbp]
    \caption{Overview of the experiment parameters of all concepts included in this review (Part III). The Measures column reports if results were produced through a questionnaire ("Q"), video/audio coding of the activity ("C"), testing participants before/after the experiment ("T"), qualitative interviews ("I"), an objective measure during the activity ("O") or complex measures produced by machine learning algorithms ("A"). For the first three types, bold font marks a validated measure. Measures showing significant effects are green, mixed or partial positive effects are cyan.}
    \label{tab:allpapers_experiments3}
    \centering
    \begin{tabular} {|p{0.1\columnwidth}|p{0.11\columnwidth}|p{0.1\columnwidth}|p{0.08\columnwidth}|p{0.11\columnwidth}|p{0.18\columnwidth}|p{0.08\columnwidth}|p{0.12\columnwidth}|}
        \toprule
        Reference     & Experiment Type & Robot Control & \# Groups \newline (Participants) & Experiment \newline Conditions & Measures & Participant \newline Specs & Relation/ \newline Reputation \\
        \hline \midrule
        Edwards et al \cite{Edwards2018ALearning} & Field Study & Autonomous & 2 (14) & No Embodiment \newline \textit{Inter-subject} & Teacher Quality Index (Q) & Teenagers \newline Peers & None\\
        \hline
        Erel et al \cite{Erel2021, Rifinski2021Human-human-robotInteraction} & User study (Lab) & Wizard & 32 (64) & Ablated Robot \newline \textit{Inter-subject} & Perceived Interaction Quality (\textcolor{lime}{\textbf{Q}}, based on \cite{Collins2000ARelationships}) \newline Interpersonal Coordination (\textcolor{lime}{\textbf{C}}, based on \cite{Bernieri1991InterpersonalSynchrony.}) \newline Non-verbal Immediacy (\textcolor{lime}{\textbf{C}}) \newline Verbal Empathy (\textcolor{lime}{\textbf{C}}) & Students \newline Friends & \\
        \hline
        Fan et al \cite{Fan2021FieldAdults, Fan2022SAR-Connect:Adults, Fan2017AModels, Fan2016AImpaired} & Field Study & Autonomous & 7 (14) & None & Effort (O) \newline Interpersonal Gaze (O) \newline Stress (O/C) \newline Interpersonal Communication (O) & Older Adults \newline Peers & None\\
        \hline
        Fu, Carolyn et al  \cite{Fu2017TurnTurn-Taking} & Proof of Concept & Autonomous &  &  &  &  & \\
        \hline
        Fu, Changzeng et al  \cite{Fu2021UsingConversations} & User study (Lab) & Autonomous & 6 (36) & Ablated Robot \newline \textit{Inter-subject} & Inclusion of Other in the Self (\textbf{Q}, \cite{Aron1992InclusionCloseness}) \newline Familiarity of Group Members (Q, based on \cite{Uchida2020ImprovingExperience}) & Adults \newline Friends/Peers & None\\
        \hline
        Fu, Di et al \cite{Fu2023TheCollaboration} & User study (Lab) & Autonomous & 25 (50) & Ablated Robot \newline \textit{Intra-subject} & Task Completion Time (\textcolor{lime}{O}) & Adults \newline Strangers & \\
        \hline
        Fucinato et al \cite{Fucinato2023CharismaticCreativity} & User study (Lab) & Wizard & 30 (100) & Intervention Type \newline \textit{Inter-subject} & Perceived Team Performance (Q) \newline Collective Engagement (\textbf{Q}, based on \cite{Salanova2003}) \newline Creativity Performance (\textcolor{lime}{\textbf{O}}, \cite{Guilford1967TheIntelligence}) \newline Group Talking Activity (O) & Students \newline Strangers & \\
        \hline
        Fukuda et al \cite{Fukuda2016AnalysisMediator} & User study (Lab) & Autonomous & 40 (120) & Human Mediator \newline \textit{Inter-subject} & Interaction Quantity (C) & Students \newline Strangers & None\\
        \bottomrule
    \end{tabular}
\end{table}
\end{landscape}
\begin{landscape}
\begin{table}[!htbp]
    \caption{Overview of the experiment parameters of all concepts included in this review (Part IV). The Measures column reports if results were produced through a questionnaire ("Q"), video/audio coding of the activity ("C"), testing participants before/after the experiment ("T"), qualitative interviews ("I"), an objective measure during the activity ("O") or complex measures produced by machine learning algorithms ("A"). For the first three types, bold font marks a validated measure. Measures showing significant effects are green, mixed or partial positive effects are cyan.}
    \label{tab:allpapers_experiments4}
    \centering
    \begin{tabular} {|p{0.1\columnwidth}|p{0.11\columnwidth}|p{0.1\columnwidth}|p{0.08\columnwidth}|p{0.11\columnwidth}|p{0.18\columnwidth}|p{0.08\columnwidth}|p{0.12\columnwidth}|}
        \toprule
        Reference     & Experiment Type & Robot Control & \# Groups \newline (Participants) & Experiment \newline Conditions & Measures & Participant \newline Specs & Relation/ \newline Reputation \\
        \hline \midrule
        Gillet, van den Bos et al \cite{Gillet2020, Gillet2020AChildren} & User study (Lab) & Wizard & 13 (39) & Ablated Robot \newline \textit{Inter-subject} & Participation Balance (O) Inclusion (T, using dictator game based on \cite{Fehr2008EgalitarianismChildren}) & Children \newline Peers \& New Member & None/\newline Task Expert\\
        \hline
        Gillet, Winkle et al \cite{Gillet2022Ice-BreakersTeenagers, Gillet2022AInteractions} & Design & Wizard & 3 (16) &  &  &  & None\\
        \hline
        Gordon et al \cite{Gordon2022InvestigationAgent, Mizrahi2022Long-TermDiscussions., Mizrahi2022VRobotator:K-12} & User study (Lab) & Autonomous & 11 (50) & No Mediator \newline \textit{Intra-subject} & Happiness Expressions (\textcolor{lime}{A}) \newline Knowledge Gain (T) \newline Accumulated Group Measures (\textcolor{cyan}{--}) \newline Custom (Q) & Teenagers \newline Peers & None\\
        \hline
        Hasegawa \& Nakauchi \cite{Hasegawa2014FacilitationExaggeration, Hasegawa2014TelepresenceTeleconferences} & Design &  &  &  &  &  & \\
        \hline
        Hayamizu et al \cite{Hayamizu2013AnControl, Sano2009AControl} & User study (Lab) & Autonomous & 10 (20) \newline 19 (38) & No Mediator \newline \textit{Intra-subject?} & Custom (Q) & Students/ Older Adults \newline Strangers & None/\newline Mediation Expert\\
        \hline
        Hitron et al \cite{Hitron2022AIRobots} & User study (Lab) & Autonomous & 10 (20) & None & Custom (I) & Students \newline Strangers & Neutral/\newline Mediation Expert\\
        \hline
        Hoffman et al \cite{Hoffman2015, Zuckerman2015EmpathyCompanions} & User study (Lab) & Autonomous & 30 (60) & Ablated Robot \newline \textit{Inter-subject} & Interaction Qualities (C) & Adults \newline Couples & \\
        \hline
        Ikari et al \cite{Ikari2020Multiple-RobotDiscussion} & User study (Lab) & Autonomous & 16 (64) & No Mediator \newline \textit{Inter-subject} & Inclusion of Other in the Self (\textbf{Q}, \cite{Aron1992InclusionCloseness}) \newline Custom (Q) & Students \newline Strangers & None\\
        \hline
        Inoue et al \cite{Inoue2021AParticipants} & Proof of Concept & Autonomous &  &  &  &  & None\\
        \hline
        Isbister et al \cite{Isbister2000HelperSpace} & User study (Lab) & Autonomous & 45 (90) & Intervention Type \newline No Mediator \newline \textit{Inter-subject} & Custom (Q) & Students \newline Strangers \newline American \& Japanese & None\\
        \bottomrule
    \end{tabular}
\end{table}
\end{landscape}
\begin{landscape}
\begin{table}[!htbp]
    \caption{Overview of the experiment parameters of all concepts included in this review (Part V). The Measures column reports if results were produced through a questionnaire ("Q"), video/audio coding of the activity ("C"), testing participants before/after the experiment ("T"), qualitative interviews ("I"), an objective measure during the activity ("O") or complex measures produced by machine learning algorithms ("A"). For the first three types, bold font marks a validated measure. Measures showing significant effects are green, mixed or partial positive effects are cyan.}
    \label{tab:allpapers_experiments5}
    \centering
    \begin{tabular} {|p{0.1\columnwidth}|p{0.11\columnwidth}|p{0.1\columnwidth}|p{0.08\columnwidth}|p{0.11\columnwidth}|p{0.18\columnwidth}|p{0.08\columnwidth}|p{0.12\columnwidth}|}
        \toprule
        Reference     & Experiment Type & Robot Control & \# Groups \newline (Participants) & Experiment \newline Conditions & Measures & Participant \newline Specs & Relation/ \newline Reputation \\
        \hline \midrule
        Jeong et al \cite{Jeong2018} & Field Study & Autonomous & 4 (12) & None & Number of Interactions (Q) \newline Custom (Q \& I) & Adults \newline Friends & \\
        \hline
        Jung, Martelaro, Hinds \cite{Jung2015UsingViolations} & User study (Lab) & Wizard & 57 (114) & Ablated Robot \newline \textit{Inter-subject} & Performance (O) \newline Affect (\textcolor{cyan}{\textbf{Q}}, based on the Self-Assessment Manikin \cite{Bradley1994MeasuringDifferential}) \newline Perceived Conflict (\textbf{Q}, based on ]\cite{Jehn1995AConflict}) \newline Perceived Confederate Contribution (Q) & Students \newline Strangers & Group-aligned/\newline Task Expert\\
        \hline
        Jung, DiFranzo et al \cite{Jung2020Robot-assistedGroups} & User study (Lab) & Wizard & 62 (124) & Ablated Robot \newline \textit{Inter-subject} & Performance (O) \newline Relationship Satisfaction (Q, based on Subjective Value Inventory \cite{Curhan2006WhatNegotiation}) & Students \newline Strangers & \\
        \hline
        Kanda et al \cite{Kanda2012ChildrenRobot} & Field Study & Wizard & 8 (31) & Ablated Robot \newline \textit{Inter-subject} & Skill \& Knowledge (T) \newline Engagement (C) & Children \newline Strangers & None/\newline Task Expert\\
        \hline
        Kang et al \cite{Kang2023TheGroup} & User study (Lab) & Wizard & 23 (46) & No Mediator \newline \textit{Inter-subject} & Perceived Mediator Usefulness (\textbf{Q}, based on \cite{Heerink2009MeasuringToolkit}) \newline Social Awareness ("Nunchi) (\textcolor{lime}{Q}) \newline Work Load (\textbf{Q} \newline NASA-TLX \cite{Hart2006Nasa-TaskLater}) & Adults \newline Strangers/ Peers & \\
        \hline
        Karatas et al \cite{Karatas2020UtilizationArt} & Proof of Concept & Autonomous & 2 (3) & No Mediator \newline \textit{Intra-subject} & Engagement (Q) & Adults \newline Performer \& Audience & \\
        \hline
        Kim et al \cite{Kim2023ChildrobotBehaviors, Kim2018DesigningOpportunities} & User study (Lab) & Wizard & 5 (10) & Intervention Type \newline \textit{Intra-subject} & Friendship Behaviors (\textcolor{lime}{\textbf{C}}) & Children \newline Peers \newline Inter-cultural & Group-aligned/\newline Novice\\
        \bottomrule
\end{tabular}
\end{table}
\end{landscape}
\begin{landscape}
\begin{table}[!htbp]
    \caption{Overview of the experiment parameters of all concepts included in this review (Part VI). The Measures column reports if results were produced through a questionnaire ("Q"), video/audio coding of the activity ("C"), testing participants before/after the experiment ("T"), qualitative interviews ("I"), an objective measure during the activity ("O") or complex measures produced by machine learning algorithms ("A"). For the first three types, bold font marks a validated measure. Measures showing significant effects are green, mixed or partial positive effects are cyan.}
    \label{tab:allpapers_experiments6}
    \centering
    \begin{tabular} {|p{0.1\columnwidth}|p{0.11\columnwidth}|p{0.1\columnwidth}|p{0.08\columnwidth}|p{0.11\columnwidth}|p{0.18\columnwidth}|p{0.08\columnwidth}|p{0.12\columnwidth}|}
        \toprule
        Reference     & Experiment Type & Robot Control & \# Groups \newline (Participants) & Experiment \newline Conditions & Measures & Participant \newline Specs & Relation/ \newline Reputation \\
        \hline \midrule
        Kobuki et al \cite{Kobuki2023RoboticStudy, Seaborn2023VoiceConversation} & User study (Lab) & Autonomous & 9 (18) \newline 12 (24) & Ablated Robot \newline \textit{Inter-subject} & System Usability Scale (\textbf{Q}, \cite{Brooke1996SUS:Scale}) & Students/ \newline Older Adults \newline Strangers & None\\
        \hline
        Kochigami et al \cite{Kochigami2018DoesCommunication} & Proof of Concept & Wizard & 1 (14) & None & Group Interactions (C) \newline Perceived Interaction (Q) & Adults/ Children \newline Unfamiliar Families & None\\
        \hline
        Leite et al \cite{Leite2016AutonomousInteraction} & Field Study & Autonomous & 24 (72) & Intervention Type \newline Ablated Robot \newline \textit{Inter-subject} & Activity Recall (T/I) \newline Emotional Understanding Score (T/I) \newline Valence \& Arousal (A) & Children \newline Peers & None\\
        \hline
        Li et al \cite{Li2023ImprovingGroups} & User study (Lab) & Autonomous & 16 (48) & No Mediator \newline \textit{Intra-subject} & Participation Balance (\textcolor{cyan}{O}) \newline Perceived Collaboration (\textbf{Q}, \cite{Chen2018BeyondCommunication}) \newline Relational Affect (\textbf{Q}) \newline Work Load (\textbf{Q}, NASA-TLX \cite{Hart2006Nasa-TaskLater}) \newline Experience (I) & Adults \newline Strangers \newline Language Skill Difference & None\\
        \hline
        Lopes et al \cite{Lopes2021SocialTeams} & User study (Lab) & Wizard & 36 (108) & Intervention Type \newline \textit{Inter-subject} & Productivity (\textcolor{lime}{O}) \newline Group Engagement (\textcolor{lime}{\textbf{Q}}, \cite{Hoffman2004CollaborationTeams} \newline Role Ambiguity (Q, \cite{Homans1958SocialExchange}) \newline Mediator Trust (Q, Human-Robot Trust Scale \cite{Schaefer2016MeasuringScale-HRI}) & Adults \newline Teams & Group-aligned/\newline Mediation Expert\\
        \hline
        Matsuyama et al \cite{Matsuyama2010PsychologicalGame, Fujie2009ConversationCommunication, Matsuyama2008DesigningCommunication} & User study (Lab) & Autonomous & 10 (30) & No Mediator \newline \textit{Inter-subject} & Custom (Q) \newline Smiling Time (\textcolor{lime}{C}) & Students \newline Strangers \newline Confederate Moderator & \\
        \hline
        Matsuyama et al \cite{Matsuyama2015Four-participantParticipant, Matsuyama2015TowardsFacilitation} & Proof of Concept & Autonomous &  &  & Custom Entitativity (Q) &  & \\
        \bottomrule
\end{tabular}
\end{table}
\end{landscape}
\begin{landscape}
\begin{table}[!htbp]
    \caption{Overview of the experiment parameters of all concepts included in this review (Part VII). The Measures column reports if results were produced through a questionnaire ("Q"), video/audio coding of the activity ("C"), testing participants before/after the experiment ("T"), qualitative interviews ("I"), an objective measure during the activity ("O") or complex measures produced by machine learning algorithms ("A"). For the first three types, bold font marks a validated measure. Measures showing significant effects are green, mixed or partial positive effects are cyan.}
    \label{tab:allpapers_experiments7}
    \centering
    \begin{tabular} {|p{0.1\columnwidth}|p{0.11\columnwidth}|p{0.1\columnwidth}|p{0.08\columnwidth}|p{0.11\columnwidth}|p{0.18\columnwidth}|p{0.08\columnwidth}|p{0.12\columnwidth}|}
        \toprule
        Reference     & Experiment Type & Robot Control & \# Groups \newline (Participants) & Experiment \newline Conditions & Measures & Participant \newline Specs & Relation/ \newline Reputation \\
        \hline \midrule
        Mitnik et al \cite{Mitnik2008AnMediator} & Field Study & Autonomous & 27 (70) & No Mediator \newline \textit{Inter-subject} & Knowledge (\textcolor{lime}{T}) \newline Collaboration (C \& Q) \newline Sociograms (Q) \newline Motivation (Q) & Teenagers \newline Peers & None\\
        \hline
        Moharana et al \cite{Moharana2019RobotsCaregivers} & Design &  &  &  &  &  & None\\
        \hline
        Nagao \& Takeuchi \cite{Nagao1994SocialAgents} & Design &  &  &  &  &  & None\\
        \hline
        Nakanishi et al \cite{Nakanishi2003CanCommunities} & User study (Lab) & Autonomous & ? (185) & Intervention Type \newline \textit{Intra-subject} & Interpersonal Similarity (\textcolor{lime}{\textbf{Q}}) \newline Interpersonal Attraction (\textcolor{lime}{\textbf{Q}}) & Students \newline Strangers & \\
        \hline
        Neto et al \cite{Neto2023TheChildren, Neto2021FosteringRobots} & User study (Lab) & Autonomous & 26 (78) & No Mediator \newline Intervention Type \newline \textit{Mixed} & Individual Contribution (C) \newline Individual Creativity (C) \newline Participation Balance (\textcolor{cyan}{C}) \newline Praising \& Being Praised (C, following \cite{Gottman1975SocialChildren}) \newline Engagement (C, following \cite{Gottman1975SocialChildren}) \newline Gaze at Group (C) \newline Solution Speed (O) \newline Inclusion of Other in the Self (Q, based on \cite{Aron1992InclusionCloseness}) \newline Perceived Inclusion (\textcolor{cyan}{Q}) \newline Perceived Mediator Utility (I) \newline Perceived Mediator Fairness (I) & Children \newline Peers \newline With \& Without Visual Impairments & \\
        \hline
        Noguchi et al \cite{Noguchi2023HowRecipients} & Proof of Concept & Autonomous & 36 (36) & Ablated Robot \newline \textit{Inter-subject} & Fear of Negative Evaluation (\textbf{Q}, \cite{Sasagawa2004DevelopmentTheory.}, based on \cite{Watson1969MeasurementAnxiety}) \newline State-Trait Anxiety Inventory (\textbf{Q}, \cite{Spielberger1971TheInventory}) & Older Adults & \\
        \bottomrule
\end{tabular}
\end{table}
\end{landscape}
\begin{landscape}
\begin{table}[!htbp]
    \caption{Overview of the experiment parameters of all concepts included in this review (Part VIII). The Measures column reports if results were produced through a questionnaire ("Q"), video/audio coding of the activity ("C"), testing participants before/after the experiment ("T"), qualitative interviews ("I"), an objective measure during the activity ("O") or complex measures produced by machine learning algorithms ("A"). For the first three types, bold font marks a validated measure. Measures showing significant effects are green, mixed or partial positive effects are cyan.}
    \label{tab:allpapers_experiments8}
    \centering
    \begin{tabular} {|p{0.1\columnwidth}|p{0.11\columnwidth}|p{0.1\columnwidth}|p{0.08\columnwidth}|p{0.11\columnwidth}|p{0.18\columnwidth}|p{0.08\columnwidth}|p{0.12\columnwidth}|}
        \toprule
        Reference     & Experiment Type & Robot Control & \# Groups \newline (Participants) & Experiment \newline Conditions & Measures & Participant \newline Specs & Relation/ \newline Reputation \\
        \hline \midrule
        Ohshima et al \cite{Ohshima2017Neut:Conversations} & User study (Lab) & Autonomous & 8 (24) & Intervention Type \newline \textit{Intra-subject} & Pressure to Speak (\textcolor{cyan}{Q}) \newline Turn-taking Pauses (C) & Students \newline Strangers & None\\
        \hline
        Ono et al \cite{Ono1999} & Proof of Concept & Autonomous &  &  &  &  & \\
        \hline
        Pliasa \& Fachantidis \cite{Pliasa2019CanDevelopment} & User study (Lab) & Wizard & 6 (12) & Human Mediator \newline \textit{Intra-subject} & Social Behavior Quality (\textcolor{lime}{C}) & Children \newline Strangers \newline With \& Without Autism & None\\
        \hline
        Rosenberg-Kima et al \cite{Rosenberg-Kima2020Robot-supportedFacilitation} & User study (Lab) & Autonomous & 9 (36) & No Embodiment \newline No Mediator \newline \textit{Intra-subject} & Attitude Toward the Group (\textbf{Q}) \newline Perceived Mediator Utility (I) & Students \newline Teams & None\\
        \hline
        Sadka, Jacobi et al \cite{Sadka2022ByOpening-encounters} & User study (Lab) & Wizard & 9 (18) & No Mediator \newline \textit{Inter-subject} & \# Interactions (C, based on \cite{Burgoon1984NonverbalReticence}) \newline Experience (I) \newline Relational Affect (I) & Students \newline Strangers & \\
        \hline
        Sadka, Parush et al \cite{Sadka2023AllConflict} & User study (Lab) & Wizard & 10 (20) & No Mediator \newline \textit{Inter-subject} & Perceived Conflict Intensity (\textcolor{lime}{Q}) \newline Intimate Behaviors (\textcolor{lime}{C}) \newline Experience (I) & Adults \newline Couples & \\
        \hline
        Scasselatti et al \cite{Scassellati2018ImprovingRobot} & Field Study & Autonomous & 12 (24) & No Mediator \textit{Intra-subject} & Performance/Social Skills (O) \newline Joint Attention Assessment (\textcolor{cyan}{\textbf{T}}, \cite{Bean2012AssessmentAdolescents}) \newline Perceived Communicative Behavior (\textcolor{lime}{Q}) & Children \& Adults \newline Authority \newline Children with Autism & \\
        \bottomrule
\end{tabular}
\end{table}
\end{landscape}

\begin{landscape}
\begin{table}[!htbp]
    \caption{Overview of the experiment parameters of all concepts included in this review (Part IX). The Measures column reports if results were produced through a questionnaire ("Q"), video/audio coding of the activity ("C"), testing participants before/after the experiment ("T"), qualitative interviews ("I"), an objective measure during the activity ("O") or complex measures produced by machine learning algorithms ("A"). For the first three types, bold font marks a validated measure. Measures showing significant effects are green, mixed or partial positive effects are cyan.}
    \label{tab:allpapers_experiments9}
    \centering
    \begin{tabular} {|p{0.1\columnwidth}|p{0.11\columnwidth}|p{0.1\columnwidth}|p{0.08\columnwidth}|p{0.11\columnwidth}|p{0.18\columnwidth}|p{0.08\columnwidth}|p{0.12\columnwidth}|}
        \toprule
        Reference     & Experiment Type & Robot Control & \# Groups \newline (Participants) & Experiment \newline Conditions & Measures & Participant \newline Specs & Relation/ \newline Reputation \\
        \hline \midrule
        Sebo et al \cite{Sebo2020TheBehavior, StrohkorbSebo2020} & User study (Lab) & Autonomous & 40 (120) & Ablated Robot \newline \textit{Inter-subject} & Perceived Group Inclusion Scale (\textbf{Q}, \cite{Jansen2014Inclusion:Measurement}) \newline Team Psychological Safety Scale (\textbf{Q}, \cite{Edmondson1999PsychologicalTeams}) \newline Back-channeling (\textbf{C}, based on \cite{Ward2000ProsodicJapanese}) \newline Participation (\textcolor{cyan}{O}) & Teenagers \newline Strangers & \\
        \hline
        Shamekhi et al \cite{Shamekhi2018} & User study (Lab) & Wizard & 20 (40) & No Embodiment \newline \textit{Inter-subject} & Relational Affect (\textcolor{cyan}{Q}) \newline Experience (Q, following \cite{Haller2010TheMeetings}) \newline Consensus \& Opinion Shift (O) \newline Solution Speed (O) \newline Participation Balance (\textcolor{lime}{O}) \newline Outcome Satisfaction (Q) & Adults \newline Strangers & None\\
        \hline
        Shamekhi \& Bickmore \cite{Shamekhi2019} & User study (Lab) & Autonomous & 20 (40) & No Mediator \newline \textit{Inter-subject} & Feelings towards Team (\textcolor{lime}{\textbf{Q}}) \newline Perceived Conflict (\textcolor{lime}{\textbf{Q}}, \cite{Jehn1995AConflict}) \newline Individual Task Confidence (\textcolor{lime}{Q}) \newline Experience (I) & Students \newline Strangers & None\\
        \hline
        Shen, Tennent et al \cite{Shen2017RobotConformity} & Design &  &  &  &  &  & \\
        \hline
        Shen, Slovak et al \cite{Shen2018} & User study (Lab) & Wizard & 32 (64) & Ablated Robot \newline \textit{Inter-subject} & Socialness of Play Behavior (\textbf{C}, based on \cite{Luckey2006UnderstandingChildhood}) \newline Constructiveness of Play Behavior (\textbf{C}, based on \cite{Luckey2006UnderstandingChildhood}) \newline Conflict Resolution (\textcolor{lime}{\textbf{C}}, based on \cite{Chen2001PeerPatterns}) & Young Children \newline Family/ Friends/ Strangers & Neutral/\newline Mediation Expert\\
        \hline
        Shim et al \cite{Shim2017AnEvaluation, Arkin2014MoralCaregiving, Pettinati2015TowardsRelationships, Shim2015AnRelationships} & Proof of Concept & Autonomous &  &  &  &  & None\\
        \bottomrule
\end{tabular}
\end{table}
\end{landscape}
\begin{landscape}
\begin{table}[!htbp]
    \caption{Overview of the experiment parameters of all concepts included in this review (Part X). The Measures column reports if results were produced through a questionnaire ("Q"), video/audio coding of the activity ("C"), testing participants before/after the experiment ("T"), qualitative interviews ("I"), an objective measure during the activity ("O") or complex measures produced by machine learning algorithms ("A"). For the first three types, bold font marks a validated measure. Measures showing significant effects are green, mixed or partial positive effects are cyan.}
    \label{tab:allpapers_experiments10}
    \centering
    \begin{tabular} {|p{0.1\columnwidth}|p{0.11\columnwidth}|p{0.1\columnwidth}|p{0.08\columnwidth}|p{0.11\columnwidth}|p{0.18\columnwidth}|p{0.08\columnwidth}|p{0.12\columnwidth}|}
        \toprule
        Reference     & Experiment Type & Robot Control & \# Groups \newline (Participants) & Experiment \newline Conditions & Measures & Participant \newline Specs & Relation/ \newline Reputation \\
        \hline \midrule
        Shimoda et al \cite{Shimoda2022ApplicationRehabilitation} & Proof of Concept & Wizard & 1 (7) & Human Mediator \newline \textit{Intra-subject} & Mediator Effort/Speaking Time (O) & Adults \newline Peers/\newline  Authority \newline Aphasia \newline Novice Moderator & None\\
        \hline
        Shin et al \cite{Shin2021} & User study (Lab) & Autonomous & 9 (18) & No Mediator \newline Ablated Robot \newline \textit{Inter-subject} & \# Conversations (O) \newline Perceived Conversation Quality/Closeness (Q, based on \cite{Burgoon1987ValidationCommunication}) & Adults \newline Strangers & None\\
        \hline
        Short \& Matari\'{c} \cite{Short2017} & User study (Lab) & Autonomous & 13 (38) & No Mediator \newline Intervention Type \newline \textit{Intra-subject} & Performance (O) \newline Participant Helpfulness (\textcolor{lime}{O}) \newline Group Cohesiveness Scale (\textcolor{lime}{\textbf{Q}}, based on \cite{Wongpakaran2013TheInpatients} & Students \newline Strangers & \\
        \hline
        Sinnema \& Alimardani \cite{Sinnema2019TheInteraction} & User study (Lab) & Autonomous & 9 (24) \newline 8 (28) & None & Perceived Interaction Quality (Q) \newline \# Social Interactions (C) \newline Mood (Q) & Older Adults/ Students \newline Peers/ Strangers & None\\
        \hline
        Stoican et al \cite{Stoican2022LearningCollaboration} & Design &  &  &  &  &  & \\
        \hline
        Stoll et al \cite{Stoll2018KeepingFormat} & Proof of Concept &  &  &  &  &  & \\
        \hline
        Strohkorb et al \cite{Strohkorb2016ImprovingRobot} & User study (Lab) & Autonomous & 43 (88) & Intervention Type \newline \textit{Inter-subject} & Performance (\textcolor{lime}{O}) \newline Perceived Performance \& Interpersonal Cohesiveness (I/Q, based on Subjective Value Inventory \cite{Curhan2006WhatNegotiation}) & Children \newline Peers & \\
        \hline
        Tahir et al \cite{Tahir2020AConversations, Tahir2014PerceptionDialogs} & User study (Lab) & Autonomous & 20 (20) & None & Mediator Usefulness (Q) & Students \& Confederate \newline Strangers & None\\
        \bottomrule
\end{tabular}
\end{table}
\end{landscape}
\begin{landscape}
\begin{table}[!htbp]
    \caption{Overview of the experiment parameters of all concepts included in this review (Part XI). The Measures column reports if results were produced through a questionnaire ("Q"), video/audio coding of the activity ("C"), testing participants before/after the experiment ("T"), qualitative interviews ("I"), an objective measure during the activity ("O") or complex measures produced by machine learning algorithms ("A"). For the first three types, bold font marks a validated measure. Measures showing significant effects are green, mixed or partial positive effects are cyan.}
    \label{tab:allpapers_experiments11}
    \centering
    \begin{tabular} {|p{0.1\columnwidth}|p{0.11\columnwidth}|p{0.1\columnwidth}|p{0.08\columnwidth}|p{0.11\columnwidth}|p{0.18\columnwidth}|p{0.08\columnwidth}|p{0.12\columnwidth}|}
        \toprule
        Reference     & Experiment Type & Robot Control & \# Groups \newline (Participants) & Experiment \newline Conditions & Measures & Participant \newline Specs & Relation/ \newline Reputation \\
        \hline \midrule
        Takano et al \cite{Takano2009PsychologicalNeeds} & Field Study & Wizard & 108 (110) & Human Mediator \newline Ablated Robot \newline \textit{Inter-subject} & Mediator Usefulness (Q) \newline Satisfaction (\textcolor{cyan}{Q}) & Adults \newline Authority \newline Doctor \& Patient & \\
        \hline
        Takeuchi \& You \cite{Takeuchi2014Whirlstools:Affordance} & Design &  &  &  &  &  & \\
        \hline
        Tennent et al \cite{Tennent2019} & User study (Lab) & Autonomous & 36 (108) & No Mediator \newline Ablated Robot \newline \textit{Inter-subject} & Performance (O) \newline Task-relevant Communication (\textbf{C}) \newline Participation Balance (O) \newline Back-channeling Balance (\textcolor{lime}{O}) & Students \newline Strangers & \\
        \hline
        Traeger et al \cite{Traeger2020, StrohkorbSebo2018TheTeams} & User study (Lab) & Autonomous & 51 (153) & Ablated Robot \newline \textit{Inter-subject} & Talking Time (\textcolor{lime}{O}) \newline Participation Balance (O) \newline Pairwise Interaction Balance (C) \newline Experience (I) & Students \newline Strangers & \\
        \hline
        Uchida et al \cite{Uchida2020ImprovingExperience} & Proof of Concept & Autonomous & 1 (6) & None & Inclusion of Other in the Self (\textbf{Q}, \cite{Aron1992InclusionCloseness}) \newline Interaction Duration (O) & Adults \newline Peers & None\\
        \hline
        Utami \& Bickmore \cite{Utami2019CollaborativeRobot, Utami2017} & User study (Lab) & Wizard & 12 (24) & None &  Positive and Negative Affect Scale (\textcolor{lime}{\textbf{Q}}, \cite{Watson1988DevelopmentScales}) \newline Inclusion of Other in the Self (\textbf{Q}, \cite{Aron1992InclusionCloseness}) \newline Closeness and Intimacy (\textcolor{lime}{\textbf{Q}}, \cite{Alea2007IllMemory}) \newline Enjoyment (Q) \newline Perceived Partner’s Responsiveness (Q) \newline Intimate Behaviors (C) & Adults \newline Couples & \\
        \bottomrule
\end{tabular}
\end{table}
\end{landscape}
\begin{landscape}
\begin{table}[!htbp]
    \caption{Overview of the experiment parameters of all concepts included in this review (Part XII). The Measures column reports if results were produced through a questionnaire ("Q"), video/audio coding of the activity ("C"), testing participants before/after the experiment ("T"), qualitative interviews ("I"), an objective measure during the activity ("O") or complex measures produced by machine learning algorithms ("A"). For the first three types, bold font marks a validated measure. Measures showing significant effects are green, mixed or partial positive effects are cyan.}
    \label{tab:allpapers_experiments12}
    \centering
    \begin{tabular} {|p{0.1\columnwidth}|p{0.11\columnwidth}|p{0.1\columnwidth}|p{0.08\columnwidth}|p{0.11\columnwidth}|p{0.18\columnwidth}|p{0.08\columnwidth}|p{0.12\columnwidth}|}
        \toprule
        Reference     & Experiment Type & Robot Control & \# Groups \newline (Participants) & Experiment \newline Conditions & Measures & Participant \newline Specs & Relation/ \newline Reputation \\
        \hline \midrule
        Wang et al \cite{Wang2023ExploringConflict} & Design &  &  &  &  &  & \\
        \hline
        Xu et al \cite{Xu2014} & Design &  &  &  &  &  & None\\
        \hline
        Yamazaki et al \cite{Yamazaki2012AVisitor} & Field Study & Autonomous & 31 (71) & None & \# Interactions (C) \newline Laughing (C) & Adults \newline Friends/ Couples \& Confederate & None\\
        \hline
        Zhang et al \cite{Zhang2023} & User study (Lab) & Wizard & 49 (98) & No Embodiment \newline \textit{Inter-subject} & Ease of Starting Conversation (Q) \newline Experience (\textcolor{cyan}{Q}, based on \cite{Kardas2022OverlyConversation}) \newline Relational Affect (Q) \newline Mediator Usefulness (Q) \newline Interpersonal Closeness (\textcolor{cyan}{Q}, based on \cite{Aron1992InclusionCloseness}) \newline Engagement (\textcolor{cyan}{\textbf{C}}) \newline Conversation Depth (\textcolor{cyan}{\textbf{C}}) & Adults \newline Strangers & None\\
        \hline
        Zheng et al \cite{Zheng2005DesigningGuide} & Proof of Concept & Autonomous &  &  & Experience (Q) &  & None/\newline Task Expert\\
        \bottomrule
\end{tabular}
\end{table}
\end{landscape}

\end{document}